\documentclass[10pt,journal,compsoc,final]{IEEEtran}
\ifCLASSOPTIONcompsoc
  \usepackage[nocompress]{cite}
\else
  \usepackage{cite}figs
\fi
\ifCLASSINFOpdf
   \usepackage[pdftex]{graphicx}
\else
   \usepackage[dvips]{graphicx}
\fi
\usepackage{amsmath}
\usepackage{epsfig}
\usepackage{graphicx}
\usepackage{amsmath}
\usepackage{amssymb}
\usepackage{color}
\usepackage{xifthen}
\usepackage{rotating}
\usepackage{mathtools}
\usepackage{nicefrac}
\usepackage{array}
\usepackage{mathtools}
\usepackage{flushend}
\usepackage{algorithmic}
\usepackage{array}
\usepackage{subfiles}
\ifCLASSOPTIONcompsoc
  \usepackage[caption=false,font=footnotesize,labelfont=sf,textfont=sf]{subfig}
\else
  \usepackage[caption=false,font=footnotesize]{subfig}
\fi
\usepackage{url}

\newcommand{\todo}[1][]{
\ifthenelse{\isempty{#1}}
	 {\textcolor{red}{(TODO)} \marginpar{\textcolor{red}{$\star$}}}
   {\textcolor{red}{(TODO: \marginpar{\textcolor{red}{$\star$}} #1)}}
}

\newcommand{\vcenteredinclude}[2]{\begingroup
\setbox0=\hbox{\includegraphics[#1]{#2}}
\parbox{\wd0}{\box0}\endgroup}

\begin{document}
\pdfminorversion=7

\title{Reframing Neural Networks: Deep Structure\\ in Overcomplete Representations}

\author{
Calvin~Murdock,~\IEEEmembership{Member,~IEEE,}
~George~Cazenavette,~\IEEEmembership{Student Member,~IEEE,}
and~Simon~Lucey,~\IEEEmembership{Member,~IEEE}
\IEEEcompsocitemizethanks{\IEEEcompsocthanksitem C. Murdock was with 
Carnegie Mellon University.\
\IEEEcompsocthanksitem G. Cazenavette is with Carnegie Mellon University.
\IEEEcompsocthanksitem S. Lucey is with Carnegie Mellon University and The University of Adelaide.}
}

\markboth{}
{Murdock \MakeLowercase{\textit{et al.}}: Reframing Neural Networks: Deep Structure 
in Overcomplete Representations}

\IEEEtitleabstractindextext{
\begin{abstract}

In comparison to classical shallow representation learning techniques, deep neural networks have achieved superior performance in nearly every application benchmark. But despite their clear empirical advantages, it is still not well understood what makes them so effective. To approach this question, we introduce deep frame approximation: a unifying framework for constrained representation learning with structured overcomplete frames. While exact inference requires iterative optimization, it may be approximated by the operations of a feed-forward deep neural network. We indirectly analyze how model capacity relates to frame structures induced by architectural hyperparameters such as depth, width, and skip connections. We quantify these structural differences with the deep frame potential, a data-independent measure of coherence linked to representation uniqueness and stability. As a criterion for model selection, we show correlation with generalization error on a variety of common deep network architectures and datasets. We also demonstrate how recurrent networks implementing iterative optimization algorithms can achieve performance comparable to their feed-forward approximations while improving adversarial robustness. This connection to the established theory of overcomplete representations suggests promising new directions for principled deep network architecture design with less reliance on ad-hoc engineering.
\end{abstract}

}

\maketitle

\IEEEdisplaynontitleabstractindextext

\IEEEpeerreviewmaketitle

\IEEEraisesectionheading{\section{Introduction}}

\IEEEPARstart{R}{epresentation learning} 
has become a key component 
of computer vision and machine learning. 
In place of manual feature engineering, deep neural networks have enabled more 
effective representations to be learned from data
for state-of-the-art performance in nearly every application benchmark. 
While this modern influx of deep learning originally began with the task of 
large-scale image recognition~\cite{krizhevsky2012imagenet}, new datasets, loss functions, and 
network configurations have expanded its scope to include a much wider range of applications. 
Despite this, the underlying architectures used to learn effective image representations 
have generally remained
consistent across all settings. This can be seen through the 
quick adoption of the newest state-of-the-art deep networks 
from AlexNet~\cite{krizhevsky2012imagenet} to VGGNet~\cite{simonyan2014very}, 
ResNets~\cite{he2016deep}, DenseNets~\cite{huang2017densely}, and so on. But this begs the question: 
why do some deep network architectures work better than others? Despite years of groundbreaking
empirical results, an answer to this question still remains elusive. 

Fundamentally, the difficulty in comparing network architectures arises from the lack of a 
theoretical foundation for characterizing their generalization capacities. 
Shallow machine learning techniques like support vector machines~\cite{cortes1995support} were 
aided by theoretical tools like the VC-dimension~\cite{vapnik1971uniform} for determining when 
their predictions could be trusted to avoid overfitting. 
The complexity of deep neural networks, on the other hand, has made similar analyses challenging.
Theoretical explorations of deep generalization
are often 
disconnected from practical applications and rarely provide actionable insights into how architectural 
hyper-parameters contribute to performance. 
Without a clear theoretical understanding, progress is largely 
driven by ad-hoc engineering and trial-and-error experimentation.

\begin{figure}
\centering
\captionsetup[subfigure]{justification=centering}
\subfloat[Deep Frame Approximation]{
\includegraphics[height=3.6cm]{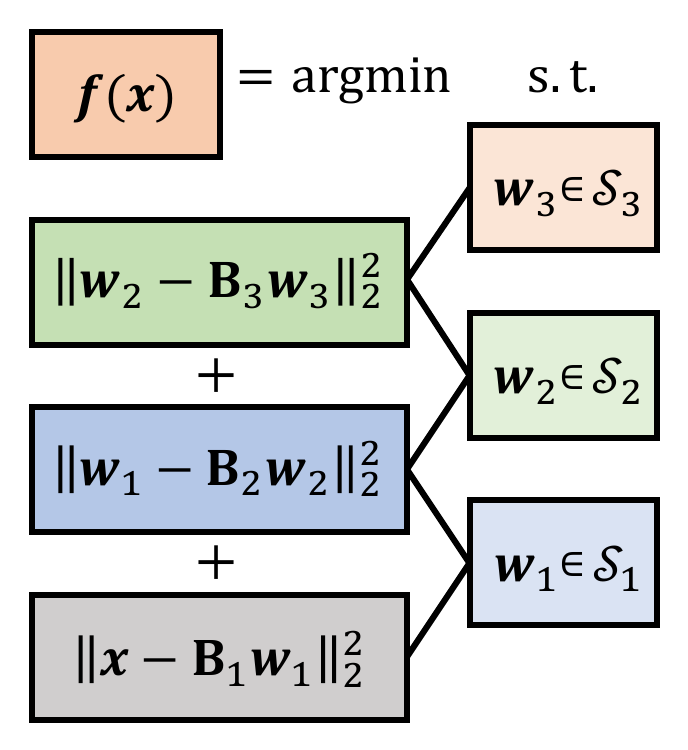}
}
\hspace{\fill}
\subfloat[Neural Network]{
\includegraphics[height=3.6cm]{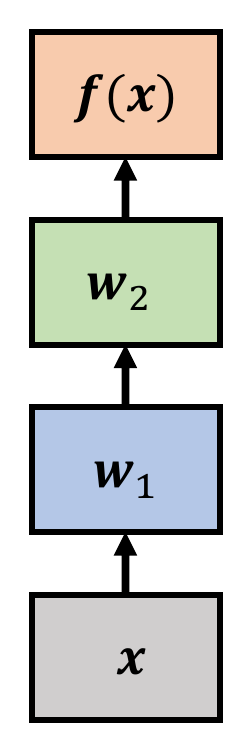}
}
\hspace{\fill}
\subfloat[Iterative Optimization Algorithm]{
\includegraphics[height=3.6cm]{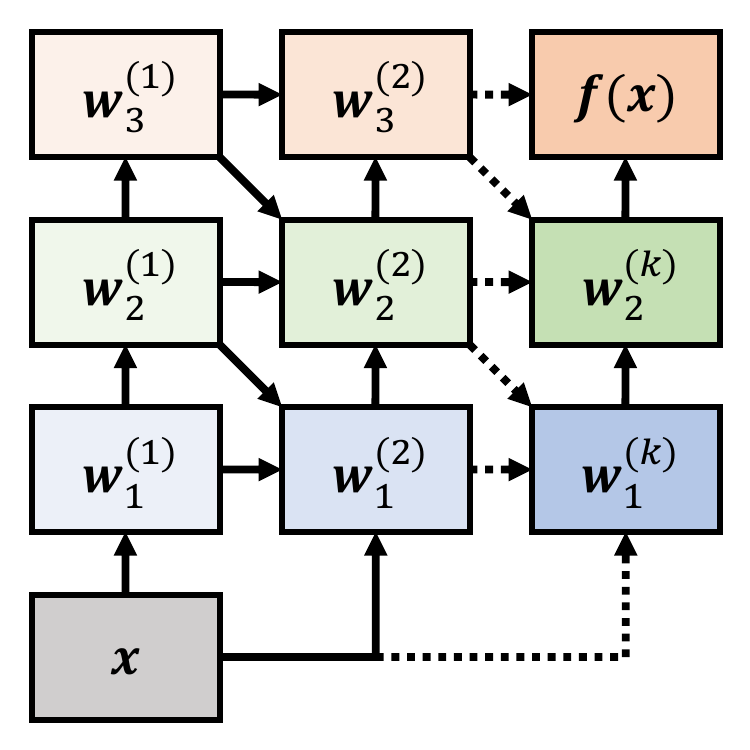}
}
\caption{
(a) Deep frame approximation is a unifying framework for multilayer representation learning where inference is posed as the constrained optimization of a multi-layer reconstruction objective. 
(b) The problem structure allows for effective feed-forward approximation with the activations of a standard deep neural network. 
(c) More accurate approximations can be found using an iterative optimization algorithm with updates implemented as recurrent feedback connections. 
}
\label{fig:deepca}
\end{figure}

Building upon recent connections between deep learning and sparse 
approximation~\cite{papyan2017convolutional, murdock2018deep, murdock2020frame}, we
introduce deep frame approximation: a unifying framework for representation learning
with structured overcomplete frames. 
These problems aim to 
optimally reconstruct 
input data through layers of 
constrained linear combinations of components from architecture-dependent overcomplete frames. 
As shown in Fig.~\ref{fig:deepca}, exact inference in our 
model amounts to finding representations
that minimize reconstruction error subject to constraints, a process that requires 
iterative optimization. However, the problem structure allows for efficient approximate inference
using standard feed-forward neural networks. This connection
between the complicated nonlinear operations of deep neural networks and convex optimization 
provides new insights about the analysis and design of different real-world network architectures.

\begin{figure}
\centering
\captionsetup[subfigure]{labelformat=empty, justification=centering}
\subfloat[(a) Chain Network]{
\includegraphics[width=0.3\columnwidth]{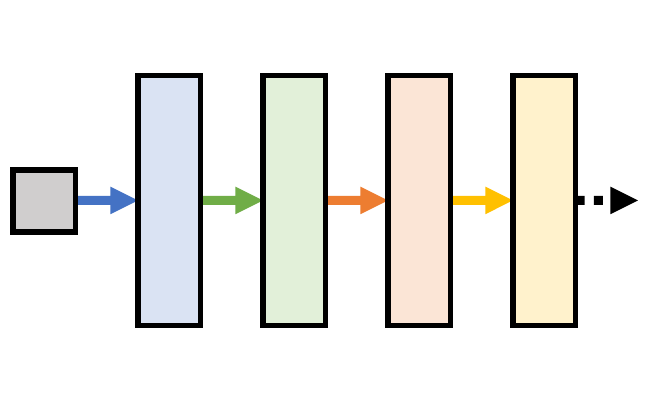}
}\hspace*{\fill}
\subfloat[(b) ResNet]{
\includegraphics[width=0.3\columnwidth]{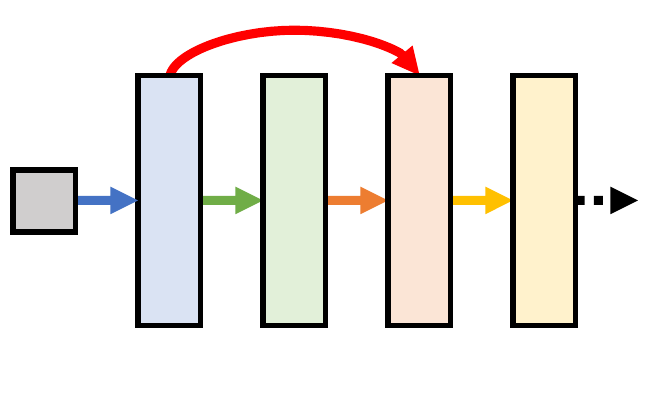}
}\hspace*{\fill}
\subfloat[(c) DenseNet]{
\includegraphics[width=0.3\columnwidth]{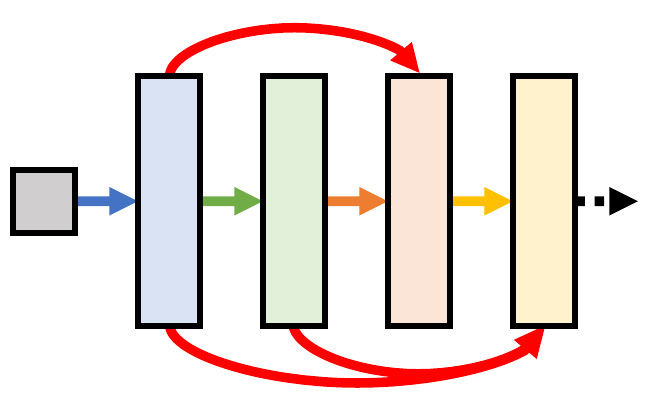}
}

\subfloat[(d) Induced Deep Frame Approximation Structures]{
\includegraphics[width=0.3\columnwidth]{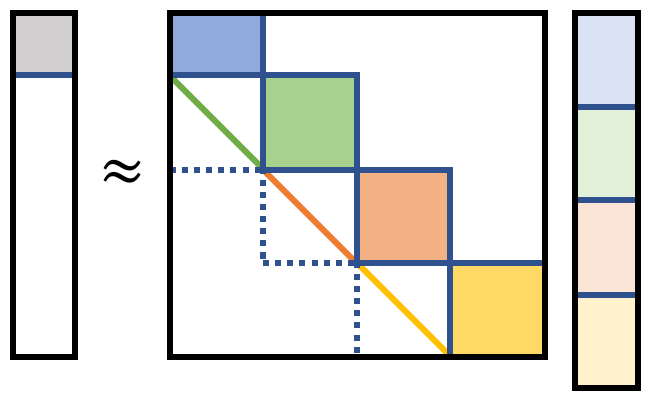}
\hspace*{0.4em}
\includegraphics[width=0.3\columnwidth]{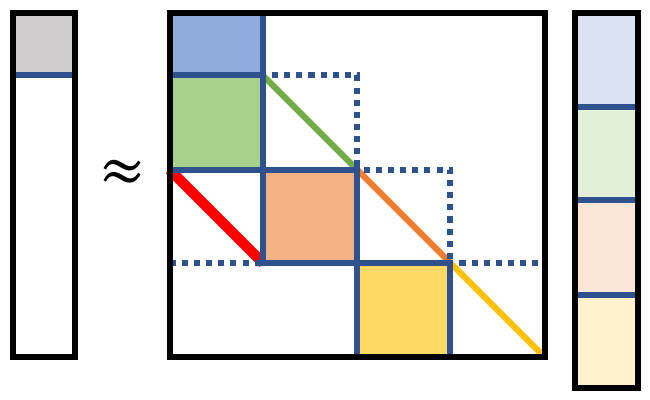}
\hspace*{0.4em}
\includegraphics[width=0.3\columnwidth]{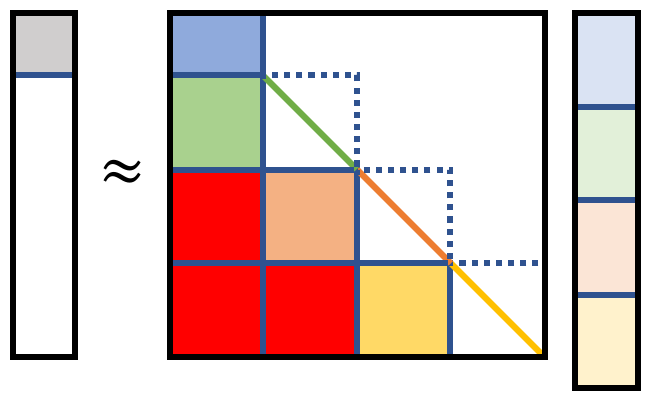}
}
\caption{
In comparison to (a) standard chain connections,
skip connections like those in (b) ResNets and 
(c) DenseNets have demonstrated significant 
improvements in parameter efficiency and 
generalization performance. We provide one possible explanation for this
phenomenon by approximating network activations as (d) solutions
to deep frame approximation problems with different induced
frame structures. 
}
\label{fig:front}
\end{figure}

Specifically, we indirectly analyze practical deep network architectures 
like residual networks (ResNets)~\cite{he2016deep} and 
densely connected convolutional networks (DenseNets)~\cite{huang2017densely} which
have achieved state-of-the-art performance in many computer vision applications.
Often very deep and with skip connections across layers, these 
complicated network architectures typically lack convincing explanations for 
their specific design choices. Without clear theoretical justifications, they are instead driven by 
performance improvements on standard benchmark datasets like ImageNet~\cite{russakovsky2015imagenet}.
As an alternative, we provide a novel perspective for evaluating and comparing these network architectures via the 
global structure of 
their corresponding deep frame approximation problems, as
shown in Fig.~\ref{fig:front}. Here, the additive optimization objective from 
Fig~\ref{fig:deepca}a is equivalently expressed within a single system where all intermediate
latent activation vectors are concatenated and all network parameters are combined 
within a global frame matrix with block-sparse structure determined by layer connectivity. More 
details may be found later in Sec.~\ref{sec:deep_frame}.

Our approach is motivated by sparse approximation 
theory~\cite{elad2010sparse}, 
a field that 
studies
shallow representations using overcomplete frames~\cite{kovavcevic2008introduction}.
In contrast to feed-forward representations constructed through linear
transformations and nonlinear activation functions, techniques like sparse coding
seek parsimonious representations that efficiently reconstruct input data.
Capacity is controlled 
by the number of 
additive components
used in sparse data reconstructions. While
adding more parameters  
allows for more accurate representations that better represent training data,
it may also increase input sensitivity;
as the number of 
components
increases, the distance between them decreases, 
so they are more likely to be confused with
one another. 
This may cause
representations of similar data points to become very far apart, leading to poor 
generalization performance.
This fundamental tradeoff between the capacity and robustness of shallow representations
can be formalized using 
similarity measures like mutual coherence--the maximum magnitude of the normalized inner products 
between all pairs of components--which are theoretically
tied to representation sensitivity and generalization~\cite{donoho2003optimally}.

\begin{figure}
\centering

\subfloat[Chain Network Gram Matrix]{
\includegraphics[height=1.5cm]{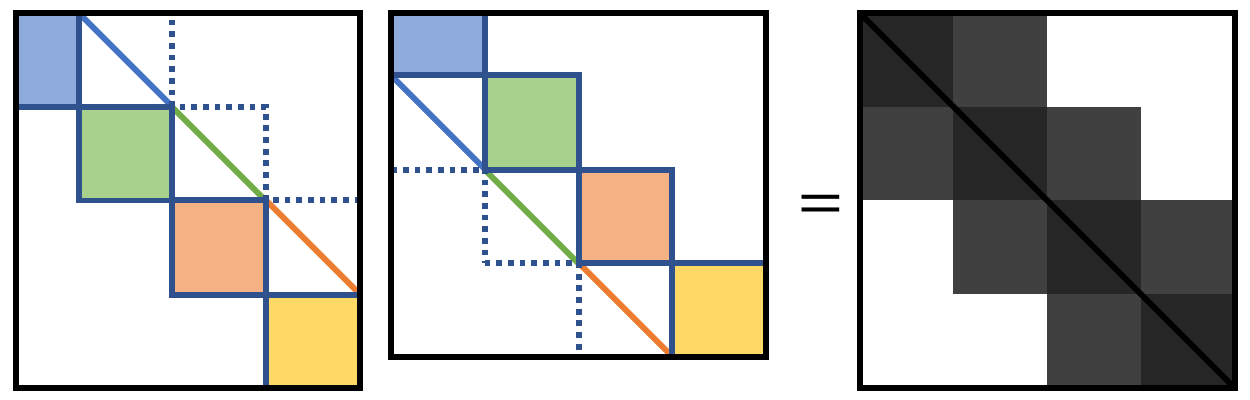}
}
\hspace*{\fill}
\subfloat[ResNet]{
\includegraphics[height=1.5cm]{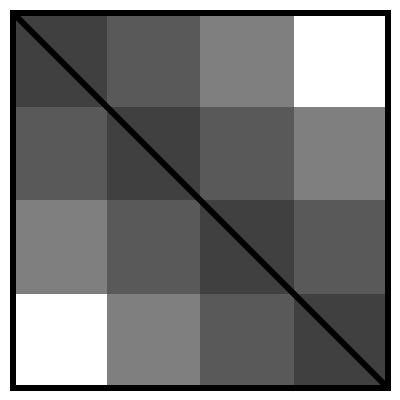}
}
\hspace*{\fill}
\subfloat[DenseNet]{
\includegraphics[height=1.5cm]{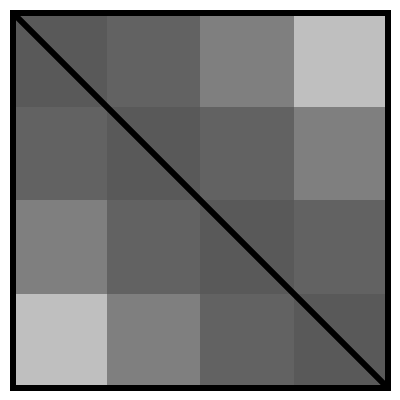}
}

\subfloat[Minimum Deep Frame Potential]{
\includegraphics[width=0.48\columnwidth]{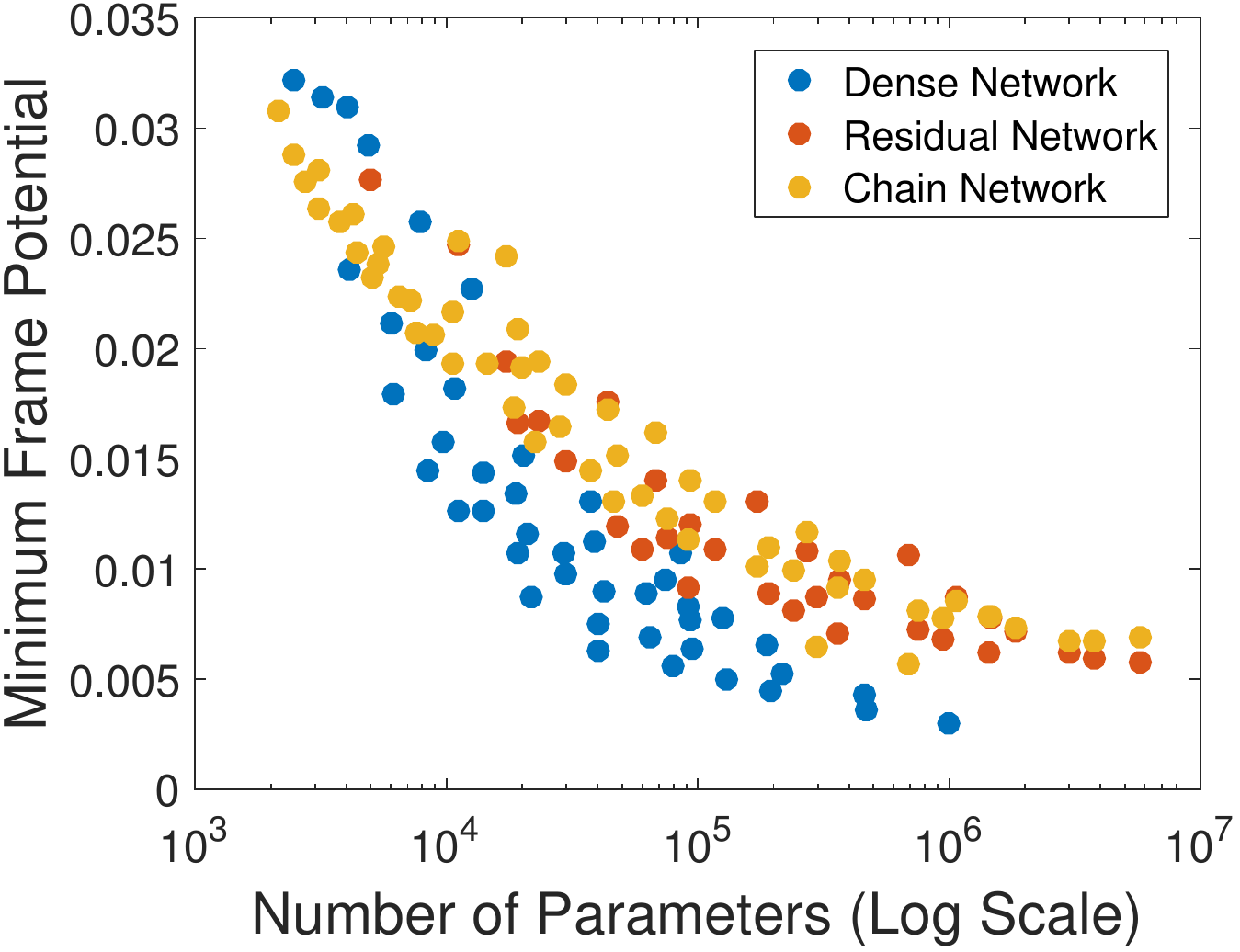}
}
\subfloat[Validation Error]{
\includegraphics[width=0.48\columnwidth]{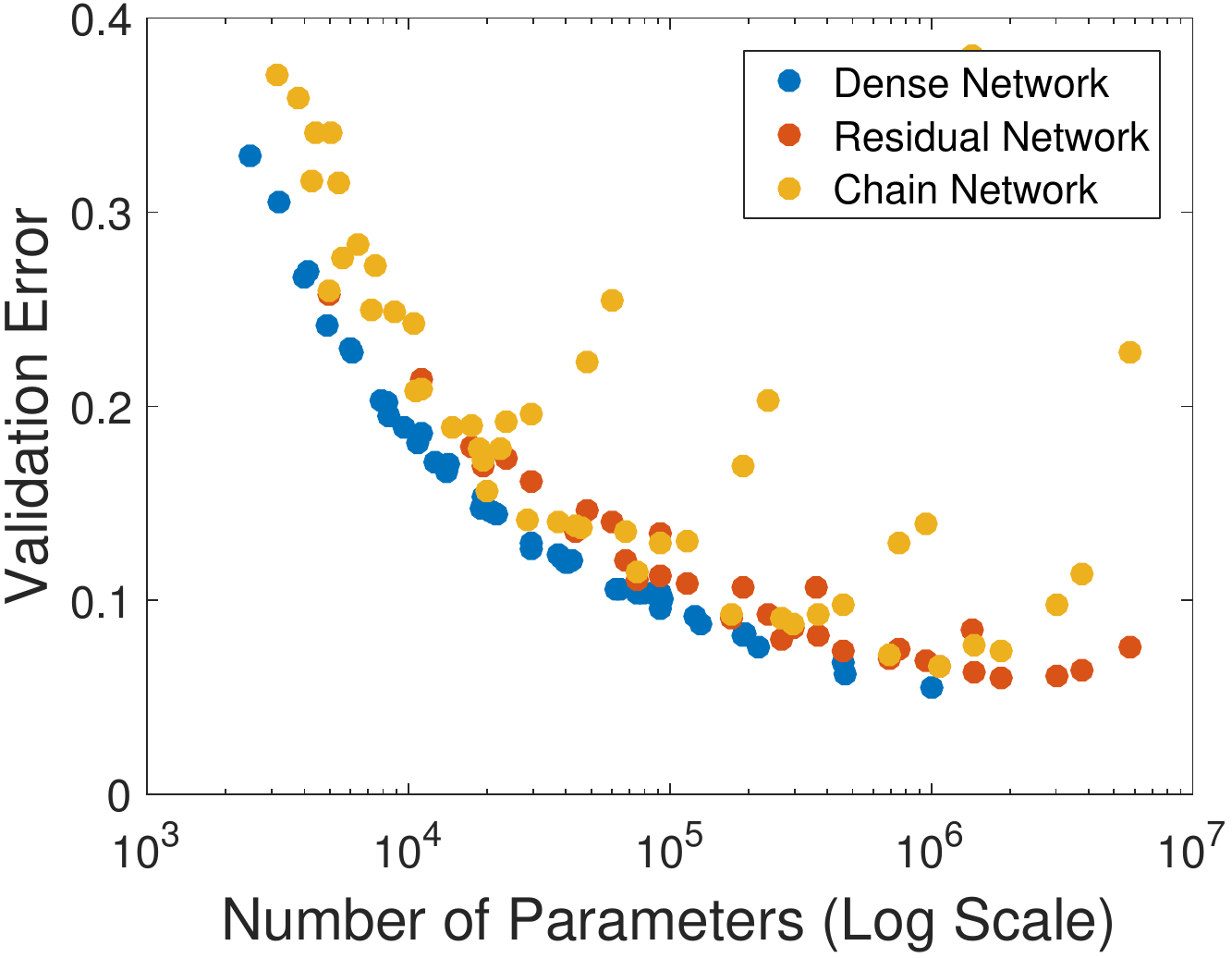}
}
\caption{
Parameter count is not a good indicator of generalization performance for deep networks. Instead, 
we compare different 
network architectures via the minimum deep frame potential, a lower bound on the mutual coherence
of their corresponding structured frames. In comparison to (a) chain networks, 
the skip connections in (b) ResNets and (c) DenseNets induce Gram matrix structures 
with more nonzero elements
allowing for (d) lower deep frame potentials across network sizes. (e) This correlates with improved parameter 
efficiency giving lower validation error with fewer parameters.
}
\label{fig:gram}
\end{figure}

Deep representations, on the other hand, have not shown the same correlation between model size and 
sensitivity~\cite{zhang2017understanding}. 
While adding more layers to a deep neural network increases its capacity, it also simultaneously
introduces implicit regularization to reduce overfitting.
Attempts to quantify this effect are theoretically challenging and often lack
intuition towards the design of architectures that more effectively balance 
memorization and generalization.
From the perspective of deep frame approximation, however, we can 
interpret this observed phenomenon simply by 
applying theoretical results from shallow representation learning.
Additional layers induce overcomplete frames with structures that 
increase both capacity and effective input dimensionality, allowing more 
components to be spaced further apart for more robust representations.
Furthermore, architectures with denser skip 
connections induce structures with more nonzero elements, providing additional 
freedom to further reduce mutual coherence with fewer parameters as shown in Fig.~\ref{fig:gram}.
From this perspective, we interpret deep learning through the lens of 
shallow learning to gain new insights towards understanding its unparalleled performance.

\subsection{Contributions}

In order to unify
the intuitive and theoretical insights of 
shallow representation learning with the 
practical advances made possible through deep learning, 
we introduce the deep frame potential as a cue for model selection that summarizes the 
interactions between parameters in 
deep neural networks.
As a lower bound on mutual coherence, it is 
tied to the generalization properties of the 
related deep frame approximation inference problems.
However, for networks with fixed connectivity, its minimizers depend only on the frame structures induced by   
the corresponding architectures. This enables 
\emph{a priori} model comparison across disparate families of 
deep networks by jointly quantifying the
contributions of depth, width, and layer connectivity.
Instead of requiring expensive validation 
on a specific dataset to approximate generalization performance, architectures can
then
be  
chosen based on how efficiently they can reduce 
mutual coherence with respect to the 
number of model parameters. While these general principles are widely applicable to many architecture 
components, they cannot be directly applied to networks with data-dependent connectivity such as
the self-attention mechanisms of Transformers~\cite{dosovitskiy2020image} where the minimum frame 
potential would depend on the underlying data distribution. 

In the case of fully-connected chain networks, we derive an analytic expression for the minimum achievable deep 
frame potential.
We also provide an efficient method for frame potential minimization applicable to a 
general class of convolutional networks with skip connections, of which ResNets and DenseNets are shown to be 
special cases.  Experimentally, we 
demonstrate correlation with validation error across a variety of 
network architectures.

This paper expands upon our original work in~\cite{murdock2020frame}, which introduced the deep frame potential
as a criterion for data-independent model selection for feed-forward deep network architectures.
Here, we provide a more thorough background of shallow representation learning with overcomplete tight 
frames to better motivate the proposed framework of deep frame approximation. While~\cite{murdock2020frame} only 
considered feed-forward approximations, we propose an iterative algorithm for exact inference
extending our work in~\cite{murdock2018deep}.
This allows us to construct recurrent versions
of networks like ResNets and DenseNets that achieve similar generalization performance. 
We also include more extensive examples building on our work in~\cite{cazenavette2021adversarial} to demonstrate the improved adversarial robustness of these recurrent optimization networks.
Experimentally, 
we compare minimum deep frame potential with generalization error on a variety of networks and datasets, including CIFAR-100 and ImageNet,
further demonstrating its ability to predict generalization capacity 
without requiring training data.

\subsection{Related Work}

In this section, we present a brief overview of shallow representation learning techniques, 
deep neural network architectures, 
some theoretical and experimental explorations of their 
generalization properties, and the relationships between deep learning and sparse approximation theory.

\subsubsection{Component Analysis and Sparse Coding}

Prior to the advent of modern deep learning, 
shallow representation learning techniques such as principal
component analysis~\cite{wold1987principal} and sparse coding~\cite{bao2016dictionary} 
were widely used due to their superior performance in
applications like facial recognition~\cite{turk1991eigenfaces}
and image classification~\cite{yang2009linear}. 
Towards the common goal of extracting meaningful information from high-dimensional
images, these data-driven approaches are  often guided by intuitive goals of incorporating 
prior knowledge into learned representations. For example, statistical 
independence allows for the separation of signals into distinct generative 
sources~\cite{jutten1991blind}, non-negativity leads to  parts-based decompositions of
objects~\cite{lee1999learning}, and sparsity gives rise to locality and frequency 
selectivity~\cite{olshausen1996emergence}. 

Even non-convex
formulations of matrix factorization are often associated with guarantees of 
convergence~\cite{bao2016dictionary}, generalization~\cite{liu2016dimensionality}, 
uniqueness~\cite{gillis2012sparse}, and even global optimality~\cite{haeffele2014structured}.
Linear autoencoders with tied weights
are guaranteed to find globally optimal solutions spanning 
principal eigenspace~\cite{baldi1989neural}. Similarly, other component analysis techniques
can be equivalently expressed in a least squares regression framework for efficient
learning with differentiable 
loss functions~\cite{delatorre2012least}. 
This unified view of undercomplete 
shallow representation learning has provided insights for characterizing certain
neural network architectures and designing more efficient optimization algorithms 
for nonconvex subspace learning problems. 

\subsubsection{Deep Neural Network Architectures}

Deep neural networks have since emerged as the preferred technique for representation learning
in nearly every application.
Their ability to jointly learn multiple 
layers of abstraction has been shown to allow for encoding increasingly complex features such as 
textures and object parts~\cite{lee2009convolutional}. 

Due to the vast space of possible deep network architectures and the computational difficulty
in training them, deep model selection has largely been guided by ad-hoc engineering and 
human ingenuity. 
Despite slowing progress in the years following early breakthroughs~\cite{lecun1998gradient},
recent interest in deep learning architectures began anew because of empirical 
successes largely attributed to computational advances like efficient training using GPUs and 
rectified linear unit (ReLU) activation 
functions~\cite{krizhevsky2012imagenet}. Since then, numerous architectural modifications have been
proposed. For example, much deeper networks with residual connections were shown 
to achieve consistently better performance with fewer parameters~\cite{he2016deep}.
Building upon this, densely connected convolutional networks with skip connections between 
more layers yielded
even better performance~\cite{huang2017densely}.

\subsubsection{Deep Model Selection}

Because theoretical explanations for effective deep network architectures are lacking, 
consistent experimentation 
on standardized benchmark datasets has been the primary driver of empirical success.
However, due to slowing progress and the need for increased 
accessibility of deep learning techniques to a wider range of practitioners, more principled
approaches to architecture search have gained traction. 
Motivated by observations of extreme redundancy in the parameters of trained 
networks~\cite{denil2013predicting}, techniques have been proposed to systematically reduce
the number of parameters without adversely affecting performance. 
Examples include sparsity-inducing regularizers during training~\cite{alvarez2016learning}
or through post-processing to prune the parameters of trained networks~\cite{he2017channel}. 
Constructive approaches to model selection like neural architecture search~\cite{elsken2019neural}
instead attempt to compose 
architectures from basic building blocks through tools like reinforcement learning.  
Efficient model scaling has also been proposed to enable more effective 
grid search for selecting architectures subject to resource 
constraints~\cite{tan2019efficientnet}.
Automated techniques can match or even 
surpass manually engineered alternatives, but they require a validation dataset 
and rarely provide insights transferable to other settings. 

\subsubsection{Deep Learning Generalization Theory}

To better understand the implicit benefits of different network architectures, there have 
also been theoretical explorations of deep network generalization. These works are often 
motivated by the surprising observation that good performance can still be achieved using 
highly over-parametrized models 
with degrees of freedom that surpass the number of training 
data. This contradicts many commonly accepted ideas about 
generalization, spurning new experimental explorations
that have demonstrated properties unique to deep learning. Examples include the ability of 
deep networks to express random 
data labels~\cite{zhang2017understanding} with a tendency towards learning 
simple patterns first~\cite{arpit2017closer}. 

Some promising directions towards explaining the generalization properties of deep 
networks are founded upon the natural stability of learning with stochastic 
gradient descent~\cite{hardt2016train}. While this has led to explicit generalization
bounds that do not rely on parameter count, they are typically only applicable for
unrealistic special cases such as very wide networks~\cite{cao2019generalization}.
Similar explanations have also arisen from the information bottleneck theory, which
provides tools for analyzing architectures with respect to the optimal tradeoff between
data fitting and compression due to random diffusion~\cite{shwartz2017opening}. However,
they are not causally connected to generalization performance and similar behaviors
can arise even without the implicit randomness of mini-batch stochastic gradients~\cite{saxe2018information}.

While exact theoretical explanations are lacking,
empirical measurements of network sensitivity such as the Jacobian norm have been shown to correlate
with generalization performance across different datasets~\cite{novak2018sensitivity}. Similarly, Parseval 
regularization~\cite{moustapha2017parseval} encourages robustness by constraining the Lipschitz
constants of individual layers. This also promotes
parameter diversity within layers, but unlike the deep frame potential, it does not take into account
global interactions between layers.

\subsubsection{Sparse Approximation}

Because of the difficulty in analyzing deep networks directly, other approaches have 
instead drawn connections to the rich field of sparse approximation theory.
The relationship between feed-forward neural networks and principal component analysis has long been 
known for the case of linear activations~\cite{baldi1989neural}.
More recently, nonlinear deep networks with ReLU activations 
have been linked to multilayer sparse coding 
to prove theoretical properties of deep 
representations~\cite{papyan2017convolutional}. This connection has been used to motivate
new recurrent architecture designs that resist adversarial noise attacks~\cite{romano2018adversarial},
improve classification performance~\cite{sulam2019multi},
or enforce prior knowledge through output constraints~\cite{murdock2018deep}. 
We further build upon these relationships to instead provide novel explanations for 
networks that have already proven to be effective  
while providing a novel approach to selecting efficient deep network architectures.

\section{Shallow Representation Learning}

Shallow representations are typically defined in terms of what they are not: deep. In contrast to deep neural networks 
composed of many layers, shallow learning techniques construct latent representations of data using only a 
single hidden layer. In neural networks, this layer typically consists of a learned linear 
transformation followed by an activation function. Specifically, given a data vector 
$\boldsymbol{x}\in\mathbb{R}^d$, a matrix of learned parameters $\mathbf{B}\in\mathbb{R}^{d\times k}$, 
and a fixed nonlinear activation function $\phi:\mathbb{R}^d\rightarrow\mathbb{R}^d$, a $k-$dimensional 
shallow representation can be found as $\boldsymbol{f}(\boldsymbol{x})=\phi(\mathbf{B}^\mathsf{T}\boldsymbol{x})$.
However, shallow representation learning also encompasses a wide range of other methods 
such as principal component analysis and sparse coding that 
are often better suited to certain tasks. Unlike neural networks, these techniques are often equipped with
simple intuition and theoretical guarantees. In this section, we survey some of these methods to
elucidate the fundamental limitations of shallow representations
and motivate our unifying framework for deep learning.

\subsection{Undercomplete Bases}

Due to the inefficiency of learning in high dimensions, 
alternative data representations are often employed for dimensionality reduction.
While high-dimensional data such as images consist of a large number of individual features
(e.g. pixels), they are often highly correlated with one another. 
More efficient encoding schemes can disentangle this structure while 
preserving the information contained in data. For example, image compression algorithms 
often take advantage of the band-limited nature of images by representing them not as individual pixels, but as  
compact linear combinations of different low-frequency components from a truncated 
basis.
In shallow representation learning, these components are instead tuned to optimize performance
on a specific dataset.

\subsubsection{Representation Inference by Best Approximation}

Dimensionality reduction techniques approximate data points $\boldsymbol{x}\in\mathbb{R}^d$ 
as linear combinations of $k\leq d$ fixed component vectors $\boldsymbol{b}_j$ for $j=1,\dotsc,k$ 
which form the columns of the undercomplete parameter matrix $\mathbf{B}\in\mathbb{R}^{d\times k}$:
\begin{equation}
\boldsymbol{x}\approx\sum_{j=1}^{k}w_{j}\boldsymbol{b}_{j}=\mathbf{B}\boldsymbol{w}
\label{eq:approximation}
\end{equation}
The vectors of reconstruction coefficients
$\boldsymbol{w}\in\mathbb{R}^k$ serve as lower-dimensional representations found by
minimizing approximation error. 
In the typical case of squared Euclidean distance, representations $\boldsymbol{f}(\boldsymbol{x})$
are inferred by solving the following optimization problem as:
\begin{equation}
\boldsymbol{f}(\boldsymbol{x})=\underset{\boldsymbol{w}}{\arg\min}\,\tfrac{1}{2}\left\lVert \boldsymbol{x}-\mathbf{B}\boldsymbol{w}\right\rVert _{2}^{2}
\label{eq:pca_inference}
\end{equation}
Because the number of component vectors is lower than their dimensionality, this problem is strongly convex
and admits a unique global solution given in closed form by simple matrix operations
as $\mathbf{B}^+\boldsymbol{x}$ where $\mathbf{B}^+=(\mathbf{B}^{\mathsf{T}}\mathbf{B})^{-1}\mathbf{B}^{\mathsf{T}}$ 
is the pseudoinverse of the rectangular matrix $\mathbf{B}$. 
In the special case of orthogonal components where $\mathbf{B}^\mathsf{T}\mathbf{B}=\mathbf{I}$, the
parameters $\mathbf{B}$ form a basis that spans a $k-$dimensional subspace of $\mathbb{R}^d$.
Furthermore, when $k=d$, $\mathbf{B}$ is complete and acts as a rotation that simply aligns data to different coordinate
vectors. From Parseval's identity, this preserves the magnitudes of representations so that 
$\lVert\mathbf{B}^{\mathsf{T}}\boldsymbol{x}\rVert_{2}=\left\lVert \boldsymbol{x}\right\rVert _{2}$.
While the representation $\boldsymbol{f}(\boldsymbol{x})=\mathbf{B}^\mathsf{T}\boldsymbol{x}$ now
takes the form of a neural network layer with a linear activation function, it has a clear geometric interpretation as 
the optimal projection onto a subspace.

\subsection{Overcomplete Frames}

For many supervised applications, dimensionality reduction is not an effective 
goal for representation learning. On the contrary, higher-dimensional representations may be necessary 
to accentuate discriminative details by encouraging linear separability. In place of undercomplete
bases, data may be represented using overcomplete frames with more components than dimensions.
While subspace representations necessitate the loss of information, frames can span the entire 
space allowing for the exact reconstruction of any data point. 

A frame is defined as a sequence $\{\boldsymbol{b}_j\}_{j\in I}$ of components from a Hibert space
that satisfy the following property for any $\boldsymbol{x}$ and some frame bounds 
$0<A\leq B<\infty$~\cite{kovavcevic2008introduction}:
\begin{equation}
A \lVert \boldsymbol{x}\rVert ^{2}\leq \sum_{j\in I}|\langle \boldsymbol{x},\boldsymbol{b}_{j}\rangle |^{2} \leq B\lVert \boldsymbol{x}\rVert ^{2}
\label{eq:frame_bound}
\end{equation}
In other words, the sum of the squared inner products between $\boldsymbol{x}$ and every $\boldsymbol{b}_j$ 
does not deviate far from the squared norm of $\boldsymbol{x}$ itself. While frequently employed 
in the analysis of infinite dimensional function spaces, we consider finite frames 
in $\mathbb{R}^d$ with the index set $I=\{1,\dotsc,k\}$. In this case, we can concatenate the frame
components $\boldsymbol{b}_j$ as the columns of $\mathbf{B}\in\mathbb{R}^{d\times k}$ where $k>d$. 

\subsubsection{Tight Frames}

If $\mathbf{B}$ is a complete orthogonal basis with $k=d$, then 
it is also a frame with frame bounds $A=B=1$~\cite{waldron2018introduction} so that:
\begin{equation}
A\lVert\boldsymbol{x}\rVert _{2}^{2}=\sum_{j=1}^{k}|\langle \boldsymbol{x},\boldsymbol{b}_{j}\rangle |^{2}=\lVert \mathbf{B}^{\mathsf{T}}\boldsymbol{x}\rVert _{2}^{2}
\label{eq:parseval_identity}
\end{equation}
More generally, a rich family of overcomplete frames with $k>d$ also satisfy this property for other $A=B$ and 
are denoted as tight frames. 
These frames are of particular interest because they behave similarly to  
complete orthogonal bases in that the frame operator matrix $\mathbf{F}=\mathbf{B}\mathbf{B}^\mathsf{T}$
is the rescaled identity matrix $A^{-1}\mathbf{I}$.
Thus, a higher-dimensional representation 
$\boldsymbol{f}(\boldsymbol{x})=\mathbf{B}^\mathsf{T}\boldsymbol{x}$ completely preserves information 
and allows for efficient exact reconstructions 
$\boldsymbol{x}=A\mathbf{B}\boldsymbol{f}(\boldsymbol{x})$.
The redundancy provided by these overcomplete frame representations also enables theoretical robustness guarantees 
that are beneficial for certain applications such as data transmission~\cite{casazza2003equal}.

\subsubsection{The Frame Potential}

Tight frames are useful in practice and can be constructed efficiently
by considering the
Gram matrix $\mathbf{G}=\mathbf{B}^\mathsf{T}\mathbf{B}$ with rank $r\leq d$, 
which contains the inner products between all combinations of frame elements. Note that for
normalized frames with magnitudes constrained to have unit norm, the diagonal contains all ones
and so its trace, the sum of its $r$ nonzero eigenvalues $\lambda_i$, is fixed to be $k$.
For tight frames, the nonzero eigenvalues are uniform with the value $A^{-1}$, the same as for the frame operator matrix 
$\mathbf{F}=\mathbf{B}\mathbf{B}^\mathsf{T}$. Thus, in order to have a fixed trace with 
uniform eigenvalues, the Gram matrix of normalized tight frames must have the minimum possible 
Frobenius norm. 
This quantity, which can be equivalently expressed as the sum of squared eigenvalues, is 
referred to as the frame potential~\cite{benedetto2003finite} and is given as:
\begin{equation}
\mathrm{FP}(\mathbf{B})=\sum_{i=1}^{r}\lambda_{i}^{2}=\sum_{j=1}^{k}\sum_{j^{\prime}=1}^{k}|\langle\boldsymbol{b}_{j},\boldsymbol{b}_{j^{\prime}}\rangle|^{2}=\left\lVert \mathbf{G}\right\rVert _{F}^{2}
\label{eq:frame_potential}
\end{equation}

Minimizers of the frame potential completely characterize the set of all
normalized tight frames. Furthermore, despite its nonconvexity, 
the frame potential can be effectively optimized using gradient descent~\cite{casazza2009gradient}.
For our purposes, this allows it to be naturally integrated into backpropagation training pipelines
as a criterion for model selection or as a regularizer.

\subsection{Nonlinear Constrained Approximation}

While tight frames admit feed-forward representations that can be efficiently decoded to reconstruct
input data, inference for general overcomplete frames again requires solving the approximation error minimization 
problem in Eq.~\ref{eq:pca_inference}. However, because the number of components is greater than the 
dimensionality, there may be an infinite subspace of coefficients $\boldsymbol{w}$ that all exactly
reconstruct any data point as $\boldsymbol{x}=\mathbf{B}\boldsymbol{w}$.
In order to guarantee uniqueness and facilitate the effective representation of data in applications 
like classification, additional constraints or penalties must be included in the optimization
problem. This yields nonlinear representations $\boldsymbol{f}(\boldsymbol{x})$ that can not be 
computed solely using linear transformations.

\subsubsection{Constraints, Penalties, and Proximal Operators} \label{sec:constraints}

\begin{figure}
\centering
\hspace{\fill}
\subfloat[Simplex Projection]{
\includegraphics[width=0.4\columnwidth]{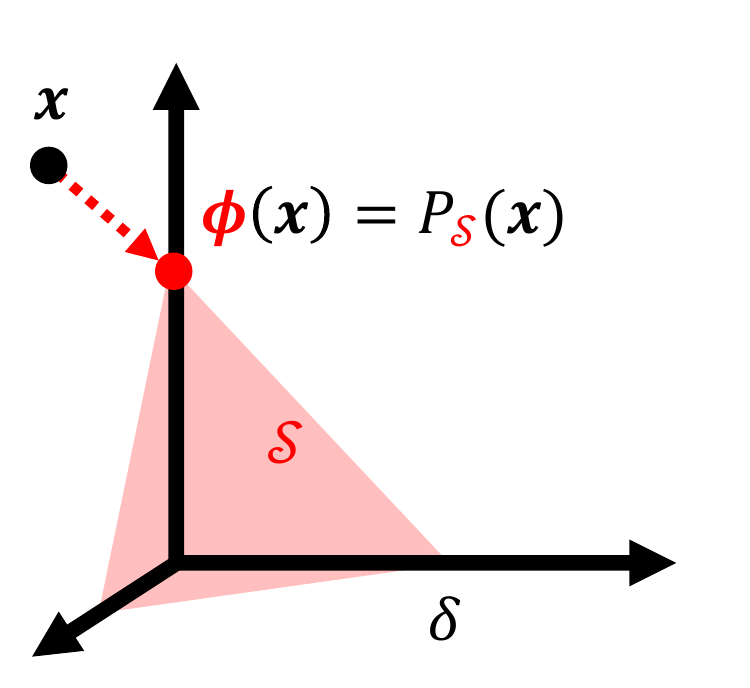}
\label{fig:relu_project}
}
\hspace{\fill}
\subfloat[Rectified Linear Unit]{
\includegraphics[width=0.4\columnwidth]{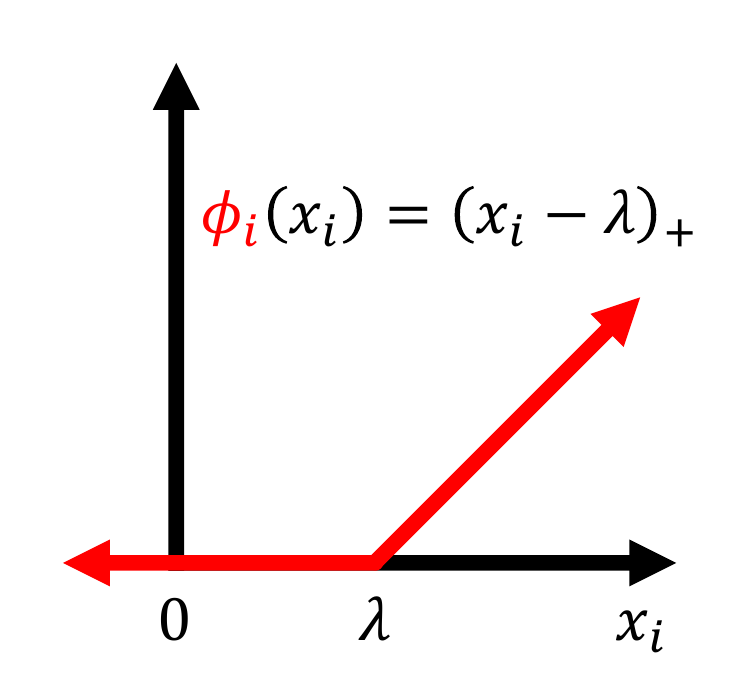}
\label{fig:relu_act}
}
\hspace{\fill}
\caption{A comparison between (a) projection onto a simplex, which corresponds to nonnegative and $\ell_1$ norm
constraints, and (b) the rectified linear unit (ReLU) nonlinear activation function, which is equivalent to a nonnegative 
softthresholding proximal operator. 
}
\label{fig:relu}
\end{figure}

\begin{figure*}
\centering
\valign{#\cr
  \hsize=0.24\textwidth
  \subfloat[Feed-Forward Inference]{\includegraphics[width=\hsize]{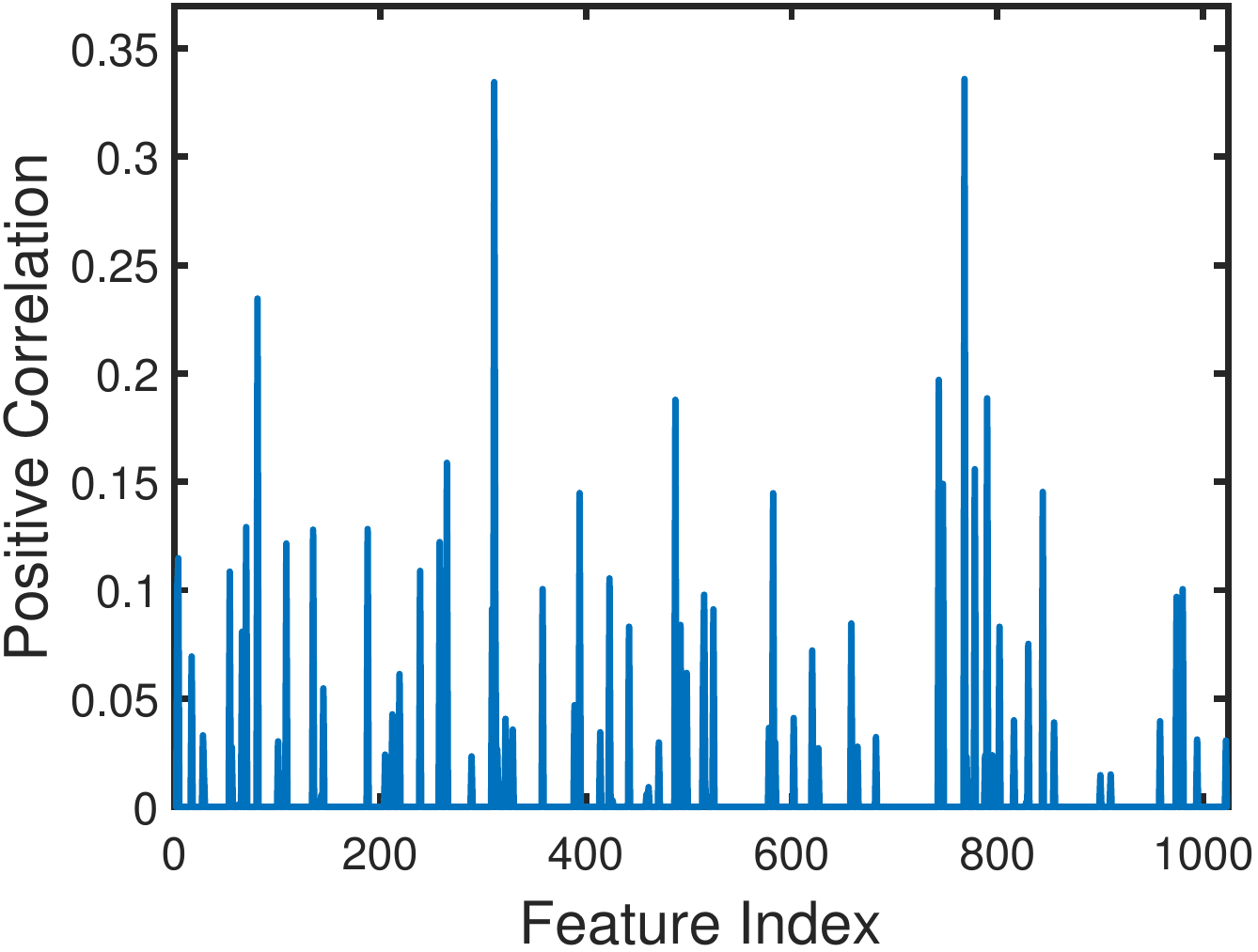}}\cr
  \noalign{\hfill}
  \hsize=0.24\textwidth
  \subfloat[Optimization Inference]{\includegraphics[width=\hsize]{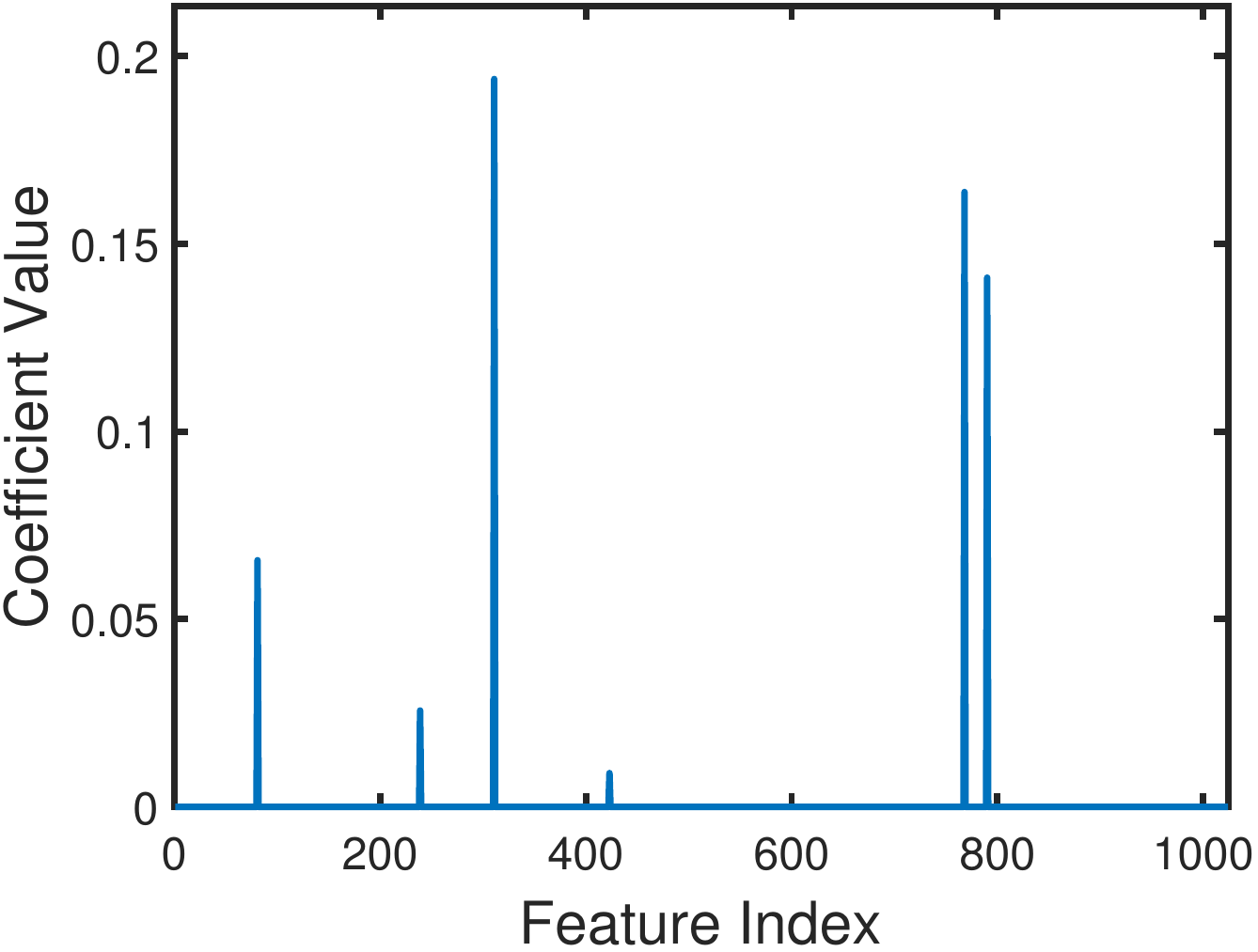}}\cr
  \noalign{\hfill}
  \hsize=0.49\textwidth
  \subfloat{
		\vcenteredinclude{width=0.1\hsize}{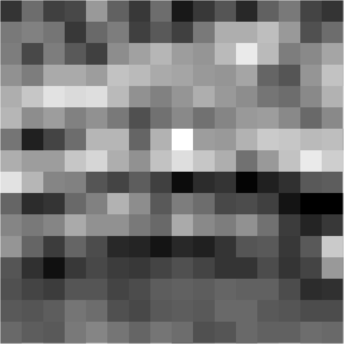}
		\makebox[0.7em]{}
		\vcenteredinclude{width=0.1\hsize}{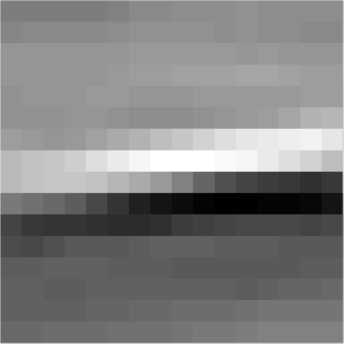}
		\makebox[0.7em]{\small $...$}
		\vcenteredinclude{width=0.1\hsize}{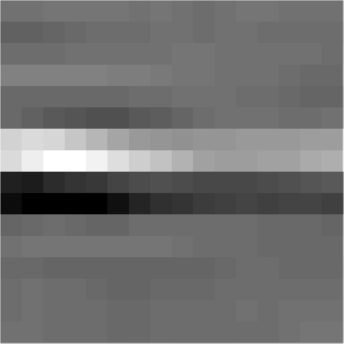}
		\makebox[0.7em]{\small $...$}
		\vcenteredinclude{width=0.1\hsize}{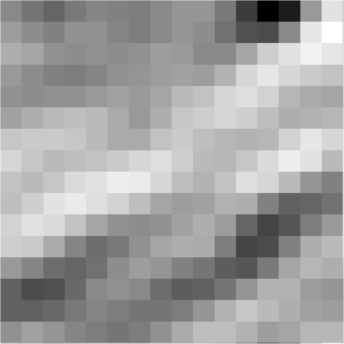}
		\makebox[0.7em]{\small $...$}
		\vcenteredinclude{width=0.1\hsize}{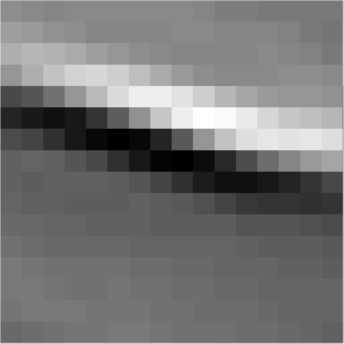}
		\makebox[0.7em]{\small $...$}
		\vcenteredinclude{width=0.1\hsize}{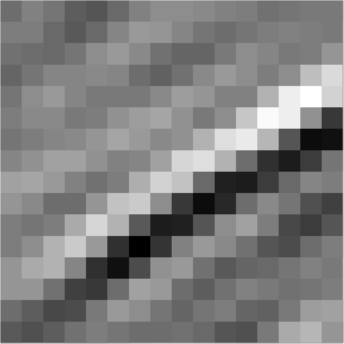}
		\makebox[0.7em]{\small $...$}
		\vcenteredinclude{width=0.1\hsize}{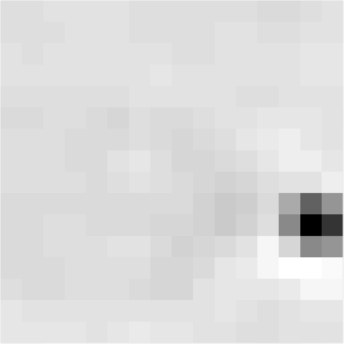}
	}
	\vspace{0.3em}
	\addtocounter{subfigure}{-1}
  \subfloat[Feed-Forward Filter Responses]{
		\makebox[0.1\hsize]{\small $\boldsymbol{x}$}
		\makebox[0.7em]{}
		\makebox[0.1\hsize]{\small 0.336}
		\makebox[0.7em]{}
		\makebox[0.1\hsize]{\small 0.122}
		\makebox[0.7em]{}
		\makebox[0.1\hsize]{\small 0.083}
		\makebox[0.7em]{}
		\makebox[0.1\hsize]{\small 0.039}
		\makebox[0.7em]{}
		\makebox[0.1\hsize]{\small 0.024}
		\makebox[0.7em]{}
		\makebox[0.1\hsize]{\small 0.001}
	}
	\vfill
  \subfloat{
		\vcenteredinclude{width=0.1\hsize}{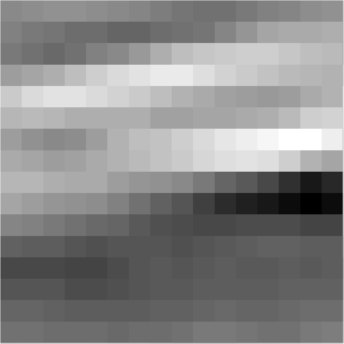}
		\makebox[0.7em]{\small =}
		\vcenteredinclude{width=0.1\hsize}{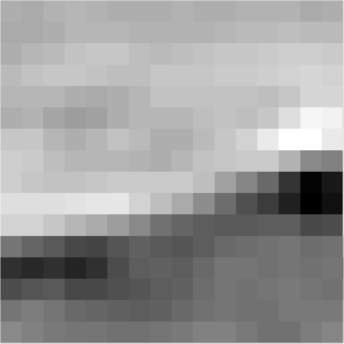}
		\makebox[0.7em]{\small +}
		\vcenteredinclude{width=0.1\hsize}{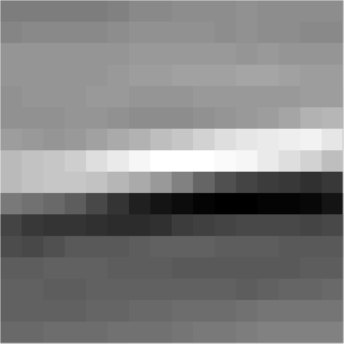}
		\makebox[0.7em]{\small +}
		\vcenteredinclude{width=0.1\hsize}{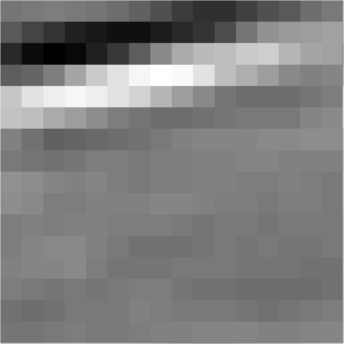}
		\makebox[0.7em]{\small +}
		\vcenteredinclude{width=0.1\hsize}{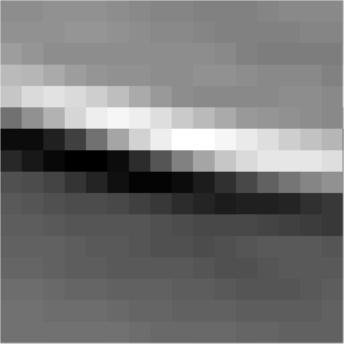}
		\makebox[0.7em]{\small +}
		\vcenteredinclude{width=0.1\hsize}{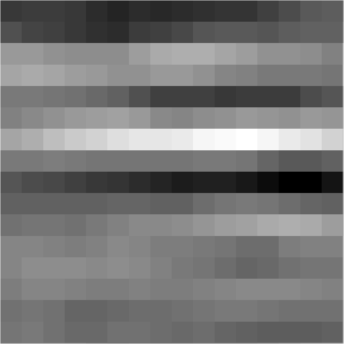}
		\makebox[0.7em]{\small +}
		\vcenteredinclude{width=0.1\hsize}{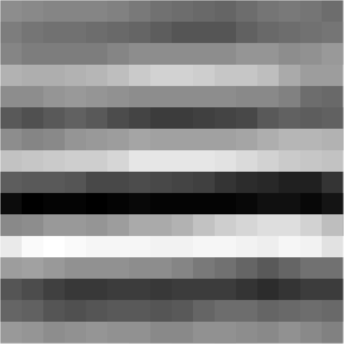}
	}
	\vspace{0.3em} 
	\addtocounter{subfigure}{-1}
	\subfloat[Optimal Reconstruction Coefficients]{
		\makebox[0.1\hsize]{\small $\hat{\boldsymbol{x}}$}
		\makebox[0.7em]{}
		\makebox[0.1\hsize]{\small 0.194}
		\makebox[0.7em]{}
		\makebox[0.1\hsize]{\small 0.164}
		\makebox[0.7em]{}
		\makebox[0.1\hsize]{\small 0.141}
		\makebox[0.7em]{}
		\makebox[0.1\hsize]{\small 0.066}
		\makebox[0.7em]{}
		\makebox[0.1\hsize]{\small 0.026}
		\makebox[0.7em]{}
		\makebox[0.1\hsize]{\small 0.009}
	}
	\cr
}

\caption{An example of the ``explaining away'' conditional dependence
provided by optimization-based inference. Sparse 
representations constructed by feed-forward nonnegative soft thresholding 
(a) have many more non-zero elements due to redundancy and spurious 
activations (c). On the other hand, sparse representations found by $\ell_1$-penalized, 
nonnegative least-squares optimization (b) yield a more parsimonious set of components (d) that 
optimally reconstruct approximations of the data. This figure was adapted from~\cite{murdock2018deep}.
}
\label{fig:explaining_away}
\end{figure*}

Constraints and penalties restrict the space of possible solutions and introduce nonlinearity that 
increases representation capacity. In addition to practical considerations, 
they can be used to encode prior knowledge for encouraging representation interpretability. For example,
consider the simplex constraint set $\mathcal{S}$ that consists of coefficient vectors that are nonnegative
with low $\ell_1$ norms: 
\begin{equation}
\mathcal{S}=\left\{ \boldsymbol{w}\,:\,\boldsymbol{w}\geq\mathbf{0},\left\lVert \boldsymbol{w}\right\rVert _{1}\leq\delta\right\} 
\label{eq:sparse_constraint}
\end{equation}
Nonnegativity ensures that learned components can have only additive effects on the data reconstruction. This 
is commonly used as a matrix factorization constraint to encourage learning more interpretable parts of objects
without supervision~\cite{lee1999learning}. The $\ell_1$ norm constraint, a convex surrogate to the $\ell_0$ ``norm'' 
that counts the number of nonzero coefficients, encourages sparsity. This gives data 
approximations that consist 
of relatively few components, resulting in the emergence of patterns resembling simple-cell receptive fields when trained 
on natural images~\cite{olshausen1996emergence}. 
The orthogonal projection onto this constraint, as visualized in Fig.~\ref{fig:relu_project}, can be found by solving the following optimization problem:
\begin{equation}
P_\mathcal{S}(\boldsymbol{x})=\underset{\boldsymbol{w}}{\arg\min}\,\tfrac{1}{2}\left\lVert 
\boldsymbol{x}
-
\boldsymbol{w}
\right\rVert _{2}^{2}\,\mathrm{s.t.}\,\boldsymbol{w}\in\mathcal{S}
\label{eq:project}
\end{equation}

We can also equivalently describe this constraint set as the penalty function 
$\Phi:\mathbb{R}^d\rightarrow\mathbb{R}$ defined as:
\begin{equation}
\Phi(\boldsymbol{w}) =\mathbb{I}_{\geq 0}(\boldsymbol{w})+\lambda\left\lVert \boldsymbol{w}\right\rVert _{1}
\label{eq:penalty}
\end{equation}
Here, the convex characteristic function $\mathbb{I}_{\geq 0}$ is defined to be 
zero when each component of its argument is nonnegative and infinity 
otherwise. The $\ell_1$ constraint radius $\delta$ is also replaced with a corresponding penalty weight $\lambda$. 

Analogous to the projection operator of a constraint set, the proximal operator of a penalty function is given by the
following optimization problem:
\begin{equation}
\boldsymbol{\phi}(\boldsymbol{x})=\underset{\boldsymbol{w}}{\arg\min}\,\tfrac{1}{2}\left\lVert 
\boldsymbol{x}
-
\boldsymbol{w}
\right\rVert _{2}^{2}+\Phi(\boldsymbol{w})
\label{eq:prox}
\end{equation}
Within the field of convex optimization, these operators are used in
proximal algorithms for solving nonsmooth optimization problems~\cite{parikh2014proximal}. 
Essentially, these techniques work by breaking a problem down 
into a sequence of smaller problems that can often be solved in closed-form.

For example, the proximal operator corresponding to the penalty function in Eq.~\ref{eq:penalty} is the 
projection onto a simplex and is given by the nonnegative soft thresholding operator:
\begin{equation}
\boldsymbol{\phi}(\boldsymbol{x})=P_{\mathcal{S}}(\boldsymbol{x})=(\boldsymbol{x}-\lambda\mathbf{1}){}_{+}
\label{eq:prox3}
\end{equation}
Visualized in Fig.~\ref{fig:relu_act}, this nonlinear function uniformly shrinks the input and clips nonnegative 
values to zero, resulting in a sparse output. Note that this is equivalent to the rectified linear unit (ReLU), 
a nonlinear activation function commonly used in deep learning, 
with a negative bias of $\lambda$. 
Many other nonlinearities can also be interpreted as proximal operators, including 
parametric rectified linear units, inverse square root units, arctangent, hyperbolic tangent, 
sigmoid, and softmax activation functions~\cite{combettes2020deep}.
This connection forms the basis of the close relationship between 
approximation-based shallow representation learning and feed-forward neural networks. 

\subsubsection{Iterative Optimization} \label{sec:iterative}

The optimization problem from Eq.~\ref{eq:pca_inference} can be 
adapted to include the sparsity-inducing constraints from Eq.~\ref{eq:sparse_constraint} as: 
\begin{equation}
\boldsymbol{f}(\boldsymbol{x})=\underset{\boldsymbol{w}\geq\mathbf{0}}{\arg\min}\,\tfrac{1}{2}\left\lVert \boldsymbol{x}-\mathbf{B}\boldsymbol{w}\right\rVert _{2}^{2}+\lambda \left\lVert \boldsymbol{w}\right\rVert _{1}
\label{eq:sparse_inference}
\end{equation}
Unlike feed-forward alternatives that construct 
representations in closed-form via independent feature detectors, 
penalized optimization problems like this require iterative solutions. One common example 
is proximal gradient descent, which separates the smooth and nonsmooth components of an 
objective function for faster convergence. Specifically, the algorithm alternates
between gradient descent on the smooth reconstruction error term and application of the 
nonnegative soft-thresholding proximal operator from Eq.~\ref{eq:prox3}. Given an initialized
representation $\boldsymbol{w}^{[0]}$ and an appropriate step size $\gamma$, 
repeated application of the update equation in Eq.~\ref{eq:prox_update} below is guaranteed to 
converge to a globally optimal solution~\cite{daubechies2004iterative}.
\begin{equation}
\boldsymbol{w}^{[t]}=\phi\big(\boldsymbol{w}^{[t-1]}
+
\gamma\mathbf{B}^{\mathsf{T}}
\big(
\boldsymbol{x}
-
\mathbf{B}\boldsymbol{w}^{[t-1]}
\big)\big)
\label{eq:prox_update}
\end{equation}

When $\boldsymbol{w}^{[0]}=0$ and $\gamma=1$, truncating this algorithm to a
single iteration results in the standard shallow feed-forward neural network representation 
$\boldsymbol{f}(\boldsymbol{x})=\phi(\mathbf{B}^\mathsf{T}\boldsymbol{x})$. Here, the proximal
operator $\phi$ takes on the role of a nonlinear activation function. 
When the parameters $\mathbf{B}$ are orthogonal, this feed-forward 
computation, which is commonly referred to as thresholding 
pursuit~\cite{papyan2017convolutional}, gives the
exact solution of Eq.~\ref{eq:sparse_inference}. More generally, parameters that are close to 
orthogonal lead to faster convergence and more accurate feed-forward approximations.

In this case, additional iterations 
introduce conditional dependence between features to  
represent data using redundant frame components. This phenomenon is 
commonly referred to as 
``explaining away'' within the context of graphical models~\cite{bengio2013representation}.
An example of this effect is shown in Fig.~\ref{fig:explaining_away}, which compares sparse 
representations constructed using a feed-forward neural network layer 
with those given by iterative optimization. 
When components in an overcomplete 
set of features have high-correlation with an image, constrained   
optimization introduces competition between them resulting in 
more parsimonious representations.

If certain conditions are met, however, the result of these two seemingly different
approaches can yield very similar results. At the same time, they can also 
enable theoretical guarantees such as uniqueness and robustness,
which are beneficial for effective representation learning. 

\subsection{Sparse Approximation Theory}

\begin{figure}
\centering
\subfloat[High Coherence]{
\includegraphics[width=0.32\columnwidth]{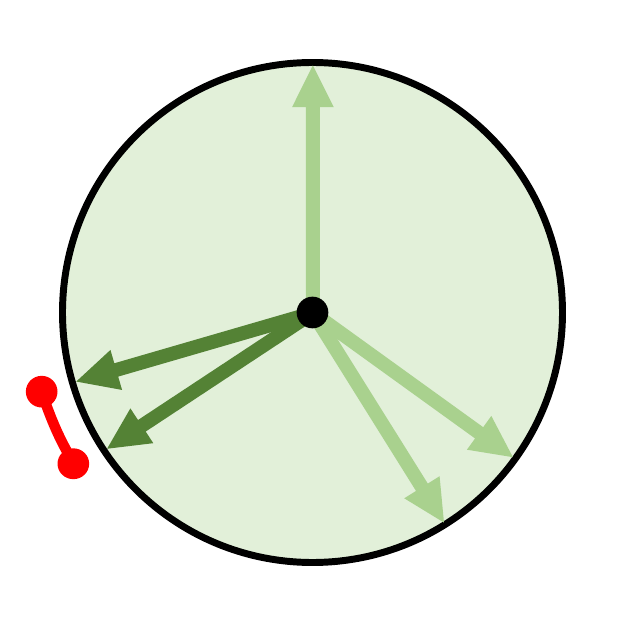}
}
\subfloat[Minimum Coherence]{
\includegraphics[width=0.32\columnwidth]{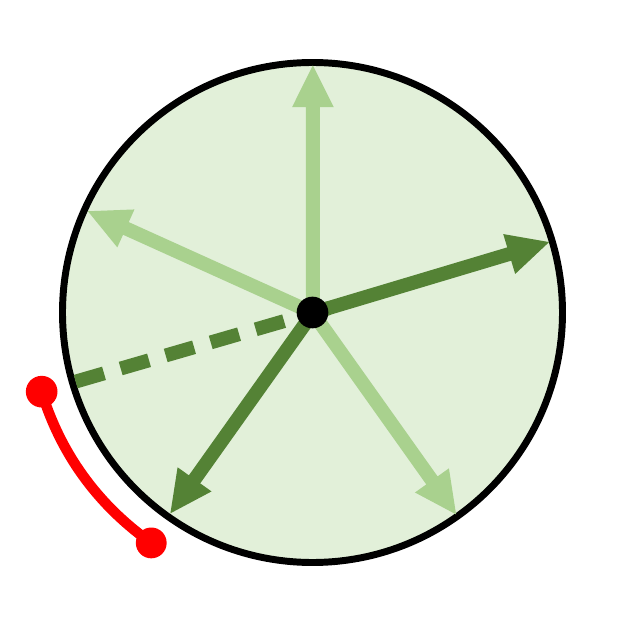}
}
\subfloat[Zero Coherence]{
\includegraphics[width=0.32\columnwidth]{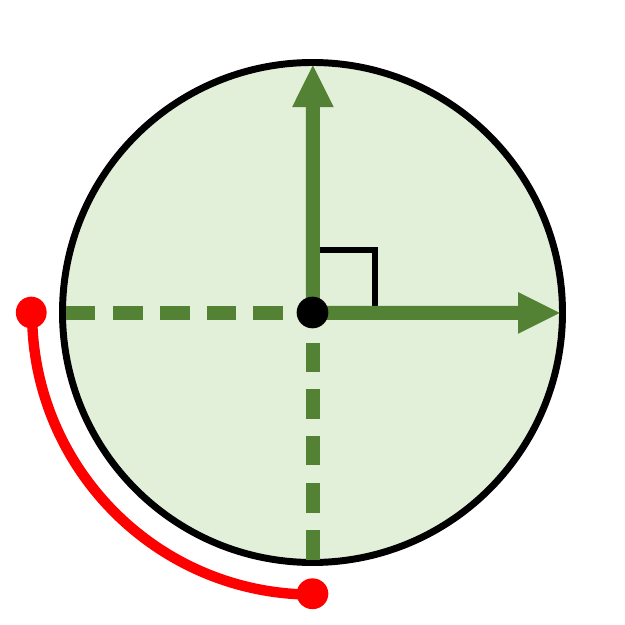}
}
\caption{Mutual coherence, visualized in red, of different frames
in $\mathbb{R}^2$. (a) When mutual coherence is high, some vectors may be very similar 
leading to representation instability and sensitivity to noise. (b) Mutual coherence is 
minimized at the Welch bound for equiangular tight frames where all vectors are equally distributed. 
(c) Mutual coherence is zero in the 
case of orthogonality, which can not occur for overcomplete frames. 
}
\label{fig:mutual_coherence}
\end{figure}

Sparse approximation theory concerns the properties of data representations  
given as sparse linear combinations
of components from overcomplete frames.
While the number of components $k$ is typically far greater than 
the dimensionality $d$, the $\ell_0$ ``norm'' $\left\Vert\boldsymbol{w}\right\Vert_0$ of the 
representation is restricted to be small. Because this quantity is
nonconvex and difficult to optimize directly, sparsity is often achieved in practice
using greedy selection algorithms like orthogonal matching pursuit~\cite{tropp2007signal}
or convex sparsity-inducing constraints like the $\ell_1$ norm~\cite{donoho2003optimally} or 
nonnegativity~\cite{bruckstein2008uniqueness}. 

Through applications like compressed sensing~\cite{donoho2006compressed}, sparsity
has been found to exhibit theoretical properties that enable data representation with 
efficiency far greater than what was previously thought possible. 
Central to these results is the 
requirement that the frame be ``well-behaved,'' essentially ensuring
that none of its columns are too similar. For undercomplete frames with $k\leq d$, this is satisfied 
by enforcing orthogonality, but overcomplete frames require other 
conditions. Specifically, we focus our attention on the mutual coherence $\mu$, 
the maximum magnitude normalized inner product of all pairs of frame components.
This is the maximum magnitude off-diagonal element in the Gram matrix 
$\mathbf{G}=\tilde{\mathbf{B}}^{\mathsf{T}}\tilde{\mathbf{B}}$ where the columns of $\tilde{\mathbf{B}}$
are normalized:
\begin{equation}
\mu(\mathbf{B})=\max_{i\neq j}\frac{|\boldsymbol{b}_{i}^{\mathsf{T}}\boldsymbol{b}_{j}|}
{\left\lVert \boldsymbol{b}_{i}\right\rVert \left\lVert \boldsymbol{b}_{j}\right\rVert }=\max_{i,j}|(\mathbf{G}-\mathbf{I})_{ij}|
\label{eq:mutual_coherence}
\end{equation}
Fig.~\ref{fig:mutual_coherence} visualizes the mutual coherence of different overcomplete frames compared
alongside an orthogonal basis. 

Because they both measure pairwise interactions between the components of an overcomplete frame,
mutual coherence and frame potential are closely related. The frame potential of a normalized 
frame can be used to compute the root mean square off-diagonal element of the Gram matrix, which 
provides a lower bound on the mutual coherence, the maximum magnitude off-diagonal element:
\begin{equation}
\mu(\mathbf{B}) \geq \sqrt{\frac{\mathrm{FP}(\tilde{\mathbf{B}}) - \mathrm{Tr}(\mathbf{G})}{\mathrm{N}(\mathbf{G})}}
\label{eq:average_fp}
\end{equation}
Here, $\mathrm{N}(\mathbf{G})$ is the number of nonzero elements in the Gram matrix and 
$\mathrm{Tr}(\mathbf{G})=k$ is constant since the columns of
$\tilde{\mathbf{B}}$ are normalized. Equality between mutual coherence and 
this averaged frame potential is met in the case of normalized equiangular tight frames, 
where all off-diagonal elements in the Gram matrix are equivalent. 

\subsubsection{Uniqueness and Robustness Guarantees} \label{sec:theory}

A model's 
capacity for low mutual coherence 
increases along with its capacity for both  
memorizing more training data through unique 
representations and generalizing to more validation data through robustness to input perturbations.
If representations from a mutually incoherent frame are sufficiently sparse, then 
they are guaranteed to be optimally sparse and unique~\cite{donoho2003optimally}. Specifically, if 
the number of nonzero coefficients 
$\left\lVert \boldsymbol{w}\right\rVert _{0}<\tfrac{1}{2}(1+\mu^{-1})$, then $\boldsymbol{w}$ is the unique,
sparsest representation for $\boldsymbol{x}$. Furthermore, if 
$\left\lVert \boldsymbol{w}\right\rVert _{0}<(\sqrt{2}-0.5)\mu^{-1}$, then it can be found efficiently 
by convex optimization with $\ell_1$ regularization. Thus, minimizing the 
mutual coherence of a frame increases 
its capacity for uniquely representing data points.

Sparse representations are also robust to input 
perturbations~\cite{donoho2005stable}. Specifically, given a noisy datapoint
$\boldsymbol{x}=\boldsymbol{x}_{0}+\boldsymbol{z}$ where $\boldsymbol{x}_{0}$ 
can be represented exactly as $\boldsymbol{x}_{0}=\mathbf{B}\boldsymbol{w}_{0}$ with 
$\left\lVert \boldsymbol{w}_{0}\right\rVert _{0}\leq\tfrac{1}{4}\left(1+\mu^{-1}\right)$
and the noise $\boldsymbol{z}$ 
has bounded magnitude $\left\lVert \boldsymbol{z}\right\rVert _{2}\leq\epsilon$, then $\boldsymbol{w}_0$
can be approximated by solving the $\ell_{1}$-penalized LASSO problem:
\begin{equation}
\underset{\boldsymbol{w}}{\arg\min}\,\tfrac{1}{2}\left\lVert \boldsymbol{x}-\mathbf{B}\boldsymbol{w}\right\rVert _{2}^{2}+\lambda\left\lVert \boldsymbol{w}\right\rVert _{1}
\label{eq:lasso}
\end{equation}
Its solution is stable and the approximation error is bounded from above in 
Eq.~\ref{eq:robustness}, where 
$\delta(\boldsymbol{x},\lambda)$ is a constant.
\begin{equation}
\left\lVert \boldsymbol{w}-\boldsymbol{w}_{0}\right\rVert _{2}^{2}\leq\frac{\left(\epsilon+\delta(\boldsymbol{x},\lambda)\right)^{2}}{1-\mu(4\left\Vert\boldsymbol{w}\right\Vert_0-1)}
\label{eq:robustness}
\end{equation}
Thus, minimizing the mutual coherence of a frame decreases the sensitivity of its sparse representations for improved robustness.
This is similar to evaluating input sensitivity using the Jacobian norm~\cite{novak2018sensitivity}. However, 
instead of estimating the average perturbation error over validation data, it bounds the 
worst-cast error over all possible data.

While most theoretical results rely on explicit sparse regularization, 
coherence has also been found to play a key role in 
other overcomplete representations. For example, nonnegativity can be sufficient
to guarantee unique solutions of incoherent underdetermined 
systems~\cite{bruckstein2008uniqueness}, formulations of overcomplete independent component analysis
often minimize coherence to promote parameter diversity~\cite{livezey2019learning}, and
similar regularizers have been employed to encourage orthogonality in deep neural network layers for reduced 
Lipschitz constants and improved generalization~\cite{moustapha2017parseval}. This suggests that coherence
control may be important even in cases when the sparsity levels required by known theoretical
guarantees are not met in practice. 

\subsubsection{The Welch Bound} \label{sec:welch}

The uniqueness and robustness of overcomplete representations can both be improved with
lower mutual coherence. Thus, one way of approximating a model's capacity for representations with 
effective memorization and generalization properties is through a lower bound on its minimum achievable 
mutual coherence. While this minimum value may not be necessary for effective representation learning 
with a particular dataset, it does provide a worst-case, data-agnostic means for comparison. 

Recall that mutual coherence and frame potential both attain minimum values of zero in the
case of orthogonal frames. For normalized overcomplete frames with more components than dimensions, 
the minimum frame potential is a positive constant that provides a lower bound on the minimum mutual 
coherence where equality is met in the case of equiangular frames. This value, typically denoted as
the Welch bound, was originally introduced in the context of bounding the effectiveness of
error-correcting codes~\cite{welch1974lower}. 

To derive this tight lower bound on the mutual coherence of a normalized frame 
$\tilde{\mathbf{B}}\in\mathbb{R}^{d\times k}$, we first apply the Cauchy-Schwarz inequality 
to find a lower bound on the frame potential, the sum of the the squared eigenvalues $\lambda_i$ 
of the positive semidefinite Gram matrix
$\mathbf{G}=\tilde{\mathbf{B}}^{\mathsf{T}}\tilde{\mathbf{B}}$:
\begin{equation}
\mathrm{FP}(\tilde{\mathbf{B}}) = \sum_{i=1}^{d}\lambda_{i}^2  \geq \frac{1}{d}\Big(\sum_{i=1}^{d}\lambda_{i}\Big)^2 = \frac{\mathrm{Tr}(\mathbf{G})^2}{d} = \frac{k^2}{d}
\label{eq:cauchy_schwarz}
\end{equation}
For normalized frames, $\mathrm{Tr}(\mathbf{G})=k$. Furthermore, 
the number of nonzero off-diagonal elements is $\mathrm{N}(\mathbf{G})=k(k-1)$. 
From Eq.~\ref{eq:average_fp}, the Welch bound is then given as:
\begin{equation}
\mu(\mathbf{B}) \geq \sqrt{\frac{kd^{-1} - 1}{k-1}}
\label{eq:welch_bound}
\end{equation}
Note that for a fixed input dimensionality $d$, as the number of parameters increases
with additional components $k$, the minimum achievable mutual coherence also increases.

\subsubsection{Structured Convolutional Frames}

\begin{figure}
\centering
\subfloat[Convolutional Frame]{
\includegraphics[width=0.4\columnwidth]{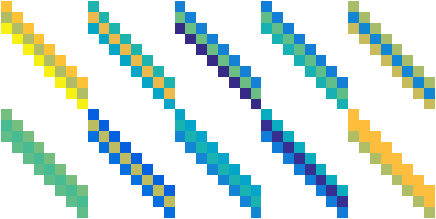} 
}
\hspace{0.25em}
\subfloat[Permuted Convolutional Frame]{
\hspace*{1.55em}
\includegraphics[width=0.4\columnwidth]{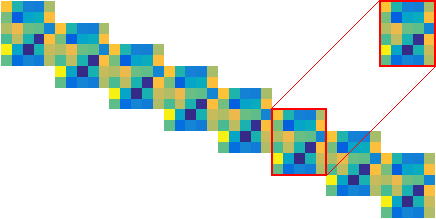}
\hspace*{1.55em}
}

\subfloat[Convolutional Gram Matrix]{
\includegraphics[width=0.4\columnwidth]{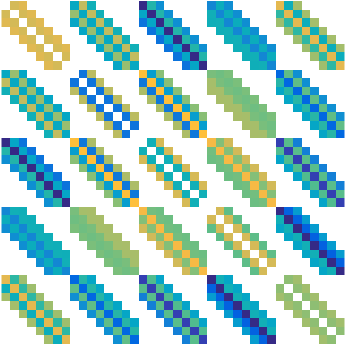}
}
\hspace{0.25em}
\subfloat[Permuted Convolutional Gram Matrix]{
\hspace*{1.55em}
\includegraphics[width=0.4\columnwidth]{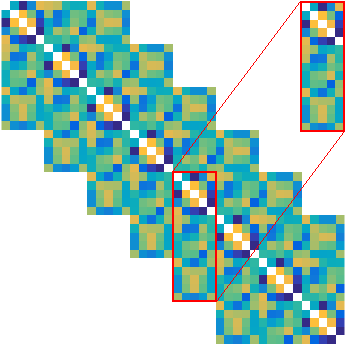}
\hspace*{1.55em}
}

\caption{
A visualization of a one-dimensional convolutional frame with two input 
channels, five output channels, and a filter size of three. (a) The filters are repeated over
eight spatial dimensions resulting in (b) a block-Toeplitz structure that is
revealed through row and column permutations. (c) The corresponding gram matrix can be 
efficiently computed by (d) repeating local filter interactions. 
}
\label{fig:conv}
\end{figure}

For complicated, high-dimensional images, sparse representation with 
overcomplete frames is generally not possible. Instead, data can be broken into smaller
overlapping patches that are represented using shared parameters with localized spatial support. 
These local representations
may then be concatenated together to form global image representations. 
This is accomplished in convolutional sparse coding by replacing the frame matrix 
multiplication in Eq.~\ref{eq:sparse_inference} with 
a linear operator that convolves a set of filters over 
coefficient feature maps to approximately reconstruct images~\cite{zeiler2010deconvolutional}. 
This operator can 
equivalently be represented as multiplication by a matrix with repeating implicit 
structures of nonzero parameters as visualized in Fig.~\ref{fig:conv}. 

Due to the repeated local structure of convolutional frames, more efficient
inference algorithms are possible~\cite{bristow2013fast} and theoretical guarantees similar to
those in Sec.~\ref{sec:theory} can be derived requiring only local sparsity independent
of input dimensionality~\cite{papyan2017working}. The frame potential, mutual coherence, 
and Welch bound may also be computed efficiently by considering only local interactions
as shown in Fig.~\ref{fig:conv}. 

Consider the 2-dimensional convolution of a $p\times p$
input with $d$ channels using $k$ filters of size $f\times f$ with $q<p$. Padding the
input with a border of zeros is used to ensure size consistency while a convolution stride of $s\geq 1$
gives filter overlaps of $o=\left\lceil\nicefrac{f}{s}\right\rceil$ for
an output of size $q\times q$ where $q=\left\lceil \nicefrac{p}{s} \right\rceil$. The normalized
convolutional frame $\tilde{\mathbf{B}}\in\mathbb{R}^{dp^2 \times kq^2}$ has a Gram matrix
with trace $\mathrm{Tr}(\mathbf{G})=kq^2$. Due its sparse repeating structure, the total number
of nonzero off-diagonal elements is:
\begin{equation}
\mathrm{N}(\mathbf{G})=k\left(k\left(o\left(2q-o+1\right)-q\right)^{2}-q^{2}\right)
\label{eq:conv_n}
\end{equation}
Using the lower bound from Eq.~\ref{eq:cauchy_schwarz}, the frame potential can be bounded as
$\mathrm{FP}(\mathbf{B})\geq q^4 k^2 p^{-2} d^{-1}$. By plugging these values into 
Eq.~\ref{eq:average_fp}, a lower bound on the mutual coherence can be found as:
\begin{equation}
\mu(\mathbf{B}) \geq \sqrt{\frac{k  d^{-1} s^{-2} - 1
}{k\left(\left(2-\left(o-1\right)sp^{-1}\right)o-1\right)^{2}-1}}
\label{eq:conv_welch_bound}
\end{equation}
While this bound depends on the size of the input $p$ and the convolution stride $s$,
the limit with increasing spatial dimensionality $p\rightarrow\infty$ for $s=1$ is
the standard multichannel aperiodic bound from~\cite{welch1974lower}:
\begin{equation}
\mu(\mathbf{B}) \geq \sqrt{\frac{kd^{-1}  - 1}{k\left(2f-1\right)^{2}-1}}
\label{eq:conv_welch_bound_limit}
\end{equation}
While the minimum mutual coherence of a convolutional frame is greater than that of a dense 
frame with the same size due to parameter sharing and its fixed structure of zeros, 
far fewer parameters are required.
This allows overcomplete representations of high-dimensional images to be more effectively 
learned from data. However, these representations are still fundamentally limited since
mutual coherence must increase as the number of filters increases.

\section{Deep Representation Learning} \label{sec:deep}

Deep representations are constructed simply as the composition of multiple feed-forward
shallow representations. For a simple chain-structured deep network with $l$ layers, 
an image $\boldsymbol{x}\in\mathbb{R}^d$ is passed through a composition of alternating linear 
transformations with parameters $\mathbf{B}_{j}\in\mathbb{R}^{k_{j-1}\times k_j}$ with $k_0=d$ and 
fixed nonlinear activation functions $\phi_j$ for layers $j=1,\dotsc,l$ as follows: 
\begin{equation}
\boldsymbol{f}(\boldsymbol{x})=\phi_l\big(\mathbf{B}_{l}^{\mathsf{T}}\cdots\phi_2(\mathbf{B}_{2}^{\mathsf{T}}(\phi_1(\mathbf{B}_{1}^{\mathsf{T}}\boldsymbol{x}))\cdots\big)
\label{eq:feed_forward}
\end{equation}
By allocating parameters across multiple nonlinear layers, representational capacity can be increased
without sacrificing generalization performance. These parameters may then be learned from data
with arbitrary loss functions using stochastic gradient descent and backpropagation. 

More complicated architectures such as ResNets and DenseNets can also be constructed by adding or 
concatenating previous outputs together, which introduces skip connections between layers. 
Despite their straightforward construction, analyzing the implicit regularization provided by 
different deep network architectures to prevent overfitting has proven to be challenging. 

\subsection{Multilayer Convolutional Sparse Coding}

Building upon the connection between feed-forward shallow representations and thresholding pursuit
for approximate sparse coding, deep neural network analysis may be simplified by accumulating the behavior of 
individual layers as discussed in further detail by Papyan et al.~\cite{papyan2017convolutional}. 
Inference is then considered as the solution of a sequence of shallow sparse 
coding problems instead of the feed-forward inference function $\boldsymbol{f}(\boldsymbol{x})$ 
from Eq.~\ref{eq:feed_forward}. Specifically, the analogous multilayer sparse coding problem is:
\begin{equation}
\begin{gathered}
\mathrm{Find}\,\{\boldsymbol{w}_{j}\}_{j=1}^{l}\quad\mathrm{s.t.}\quad\left\lVert \boldsymbol{w}_{j-1}-\mathbf{B}_{j}\boldsymbol{w}_{j}\right\rVert _{2}^{2}\leq\mathcal{E}_{j-1},\,\,
\\ \hspace{\fill}\Psi(\boldsymbol{w}_{j})\leq\delta_{j},\quad\forall j=1,\dotsc,l
\end{gathered}
\label{eq:dcp}
\end{equation}
Here, $\boldsymbol{w}_0=\boldsymbol{x}$ and $\boldsymbol{w}_j$ for $j=1,\dotsc,l$ are inferred latent 
variables corresponding to layers parameterized by $\mathbf{B}_j$. The inference 
problem amounts to jointly finding the outputs of each layer $\boldsymbol{w}_j$ such that the 
error in approximating the previous layer is at most $\mathcal{E}_{j-1}$ and they are
sufficiently sparse as measured by the function $\Psi$ with a threshold $\delta_j$. 
For convolutional sparse coding, the sparsity function may 
be locally relaxed to allow for invariance to input size~\cite{papyan2017working}. 

Assuming that the parameters have been learned such that solutions exist for a particular
dataset, the final layer output representations $\boldsymbol{f}(\boldsymbol{x})=\boldsymbol{w}_l$ satisfy certain
uniqueness and robustness guarantees given sufficiently 
low mutual coherence at each layer~\cite{papyan2017convolutional}. 

\subsubsection{Approximate Inference}

The problem in Eq.~\ref{eq:dcp} may be approximated by 
the composition of layered thresholding pursuit algorithms with 
feed-forward deep neural network operations:
\begin{equation}
\boldsymbol{w}_{j}\coloneqq\boldsymbol{\phi}_{\lambda_{j}}(\mathbf{B}_{j}^{\mathsf{T}}\boldsymbol{w}_{j-1}),\quad\forall j=1,\dotsc,l
\label{eq:layered_thresh}
\end{equation}
Here, nonlinear activation functions $\phi$ are interpreted as proximal operators 
corresponding to sparsity-inducing penalty functions $\Phi$. 
Because the sparsity constraint
functions $\Psi$ from Eq.~\ref{eq:dcp} used in theoretical analyses are  
difficult to optimize directly, convex surrogates like those from Eq.~\ref{sec:constraints} may be
used instead as long as the penalty weights $\lambda_j$ are chosen to satisfy the required
sparsity level. This allows standard convolutional networks with sparsity-inducing
activation functions like ReLU to be analyzed under the framework of multilayer sparse coding. 

Other optimization algorithms may also be used to better approximate the solution to
Eq.~\ref{eq:dcp}. In layered basis pursuit, inference is posed as a sequence of shallow sparse coding problems.
Specifically, a deep representation is given by composing the solutions for each layer $j=1,\dotsc,l$:
\begin{equation}
\boldsymbol{w}_{j}\coloneqq\underset{\boldsymbol{w}_{j}}{\arg\min}\,\tfrac{1}{2}\left\lVert \boldsymbol{w}_{j-1}-\mathbf{B}_{j}\boldsymbol{w}_{j}\right\rVert _{2}^{2}+\lambda_{j}\Phi(\boldsymbol{w}_{j})
\label{eq:layered_bp}
\end{equation}
As discussed in Sec.~\ref{sec:iterative}, these shallow problems may be solved by iterative optimization
algorithms like proximal gradient descent
implemented 
using neural network operations. In comparison to standard feed-forward networks, 
these recurrent architectures provide improved theoretical robustness 
guarantees and reduced sensitivity to adversarial input perturbations~\cite{romano2018adversarial}.

Interpreting deep networks as compositions of approximate sparse coding algorithms provides 
some insights into their robustness and generalization properties, but the theory does not 
apply to models typically used in practice. 
The error accumulation used in the analysis applies only to chain-structured compositions of layers, and it  
does not effectively explain the empirical successes of increasingly deeper networks. Furthermore, discriminative 
training often leads to networks with good generalization performance despite relatively high mutual coherence and 
low sparsity. 

\subsection{Deep Frame Approximation} \label{sec:deep_frame}

Deep frame approximation is an extension of multilayer sparse coding where inference is posed 
not as a sequence of layer-wise problems, but as a single global problem with structure induced by
the corresponding network architecture.
This formulation was first introduced in Deep Component Analysis with the motivation of iteratively enforcing
prior knowledge with optimization constraints~\cite{murdock2018deep}. 
From the perspective of deep neural networks, it 
can be derived by relaxing the implicit assumption that layer inputs are constrained 
to be the outputs of previous layers. Instead of being computed in sequence, 
the latent variables $\boldsymbol{w}_j\in\mathbb{R}^{k_j}$ 
representing the outputs of all layers $j=1,\dotsc,l$
are inferred jointly as shown in Eq.~\ref{eq:chain} below,
where $\boldsymbol{w}_0=\boldsymbol{x}$.
\begin{equation}
\underset{\{\boldsymbol{w}_{j}\}}{\arg\min}\sum_{j=1}^{l}\tfrac{1}{2}\left\lVert \boldsymbol{w}_{j-1} - \mathbf{B}_{j}\boldsymbol{w}_{j}\right\rVert _{2}^{2}+\Phi_{j}(\boldsymbol{w}_{j})
\label{eq:chain}
\end{equation}
While we only consider the non-negative sparsity-inducing penalties $\Phi_j$ from Sec.~\ref{sec:constraints} 
corresponding to ReLU activation functions, other convex regularizers may also be used. 

The compositional constraints between adjacent layers from Eq.~\ref{eq:layered_bp} have been relaxed and 
the reconstruction error penalties are combined into a single convex, nonnegative sparse coding problem. 
By combining the terms in the summation of Eq.~\ref{eq:chain} together, this problem can be 
equivalently represented as the structured problem in Eq.~\ref{eq:chain_opt} below. 
The latent  variables $\boldsymbol{w}_j$ for each layer are stacked in the vector 
$\boldsymbol{w}$, the regularizer $\Phi(\boldsymbol{w})=\sum_j \Phi_j(\boldsymbol{w}_j)$, 
and the input $\boldsymbol{x}$ is augmented by padding it with zeros.
The layer parameters 
$\mathbf{B}_j\in\mathbb{R}^{k_{j-1}\times k_j}$ are blocks in the induced global frame $\mathbf{B}$, 
which has $\sum_j k_{j-1}$ rows and $\sum_j k_j$ columns. 
\begin{equation}
\arraycolsep=1.3pt\def\arraystretch{1.0}
\underset{\boldsymbol{w}}{\arg\min}\,\tfrac{1}{2}\left\Vert 
\begin{array}{c}
\\
\\
\\
\\
\\
\end{array}\hspace{-0.3em}
\right.\overbrace{\left[\begin{array}{cccc}
\mathbf{B}_{1} & \mathbf{0} &  & \\
-\mathbf{I} & \mathbf{B}_{2} & \ddots\\
 & \ddots & \ddots & \mathbf{0}\\
 &  & -\mathbf{I} & \mathbf{B}_{l}
\end{array}\right]}^{\mathbf{B}}\overbrace{\left[\begin{array}{c}
\boldsymbol{w}_{1}\\
\vphantom{\ddots}\boldsymbol{w}_{2}\\
\vdots\\
\boldsymbol{w}_{l}
\end{array}\right]}^{\boldsymbol{w}}-\left[\begin{array}{c}
\boldsymbol{x}\\
\vphantom{\ddots}\mathbf{0}\\
\vdots\\
\mathbf{0}
\end{array}\right]\left.\begin{array}{c}
\\
\\
\\
\\
\\
\end{array}\hspace{-0.3em}\right\Vert _{2}^{2}\hspace{-0.4em}+\Phi(\boldsymbol{w})
\hspace{-0.3em}
\label{eq:chain_opt}
\end{equation}
The frame $\mathbf{B}$ has a
structure of nonzero elements that summarizes the corresponding feed-forward deep network architecture 
wherein off-diagonal identity matrices represent connections between adjacent layers.
Because the data is augmented with zeros increasing its dimensionality, the feed-forward shallow thresholding 
pursuit approximation is ineffective, always giving $\boldsymbol{w}_j=0$ for $j>1$.
However, due to its lower-block-triangular structure, Eq.~\ref{eq:chain_opt} can 
be effectively approximated by considering each layer in sequence using the same 
feed-forward operations from Eq.~\ref{eq:layered_thresh}.

From the perspective of deep frame approximation, deep neural networks approximate 
shallow structured sparse coding problems in higher dimensions.
Model capacity can be increased both by adding 
parameters to a layer or by adding 
layers, which implicitly pads the input data $\boldsymbol{x}$ with more zeros. 
By increasing dimensionality, this can actually reduce mutual coherence by 
expanding the space between frame components. 
Thus, depth allows model complexity to scale jointly alongside effective input 
dimensionality so that the induced frame structures have the capacity 
for low mutual coherence
and representations with improved memorization and generalization.

\subsubsection{Architecture-Induced Frame Structure}

This model formulation may be extended to accommodate
more complicated network architectures with skip connections. This allows for 
the direct comparison of problem structures induced by different families of
architectures such as ResNets and DenseNets via their capacity for achieving
incoherent global frames. 

Mutual coherence is computed from normalized frame components, so it
can be lowered simply by increasing the number of nonzero elements, which reduces
their pair-wise inner products. This may be interpreted as 
providing  more degrees of freedom to spread out by  
removing implicit lower-dimensional subspace constraints.
In Eq.~\ref{eq:dense_structure} below, we replace the identity connections of Eq.~\ref{eq:chain_opt} 
with blocks of nonzero parameters.
\begin{equation}
\arraycolsep=1pt\def\arraystretch{1.0}
\mathbf{B}=\left[\begin{array}{cccc}
\hphantom{-}\mathbf{B}_{11} & \vphantom{\ddots}\mathbf{0} &  \cdots & \mathbf{0}\\
-\mathbf{B}_{21}^{\mathsf{T}} & \hphantom{-}\mathbf{B}_{22} & \ddots &  \vdots \\
\vdots & \ddots & \ddots & \mathbf{0}\\
-\mathbf{B}_{l1}^{\mathsf{T}} & \cdots & -\mathbf{B}_{l(l-1)}^{\mathsf{T}} & \vphantom{\ddots}\hphantom{-}\mathbf{B}_{ll}
\end{array}\right]
\label{eq:dense_structure}
\end{equation}
This frame structure corresponds to the global deep frame approximation inference problem:
\begin{equation}
\begin{gathered}
\underset{\{\boldsymbol{w}_{j}\}}{\arg\min}\,\tfrac{1}{2}\big\lVert \boldsymbol{x}-\mathbf{B}_{11}\boldsymbol{w}_{1}\big\rVert _{2}^{2} + \sum_{j=1}^{l}\Phi_{j}(\boldsymbol{w}_{j}) 
\\[-0.5em]
+\sum_{j=2}^{l}\tfrac{1}{2}\,\Big\Vert \mathbf{B}_{jj}\boldsymbol{w}_{j}-\sum_{k=1}^{j-1}\mathbf{B}_{jk}^{\mathsf{T}}\boldsymbol{w}_{k}\Big\Vert _{2}^{2}
\end{gathered}
\label{eq:dense_opt} 
\end{equation}
Because of the block-lower-triangular structure, following the first layer activation
$\boldsymbol{w}_{1}\coloneqq\phi_{1}(\mathbf{B}_{11}^{\mathsf{T}}\boldsymbol{x})$,
inference can again be approximated 
by the composition of feed-forward thresholding pursuit activations:  
\begin{equation}
\boldsymbol{w}_{j} \coloneqq\phi_{j}\Big(\mathbf{B}_{j}^{\mathsf{T}}\sum_{k=1}^{j-1}\mathbf{B}_{jk}^{\mathsf{T}}\boldsymbol{w}_{k}\Big)\quad \forall j=2,\dotsc,l
\label{eq:dense}
\end{equation}
In comparison to Eq.~\ref{eq:chain}, additional parameters introduce skip connections 
between layers so that the activations $\boldsymbol{w}_j$ of layer $j$ now depend on those of 
all previous layers $k<j$.

\subsubsection{Residual Networks}

These connections are similar to the identity mappings in residual networks~\cite{he2016deep}, which introduce 
dependence between the activations of alternating pairs of layers. Specifically, after an initial layer 
$\boldsymbol{w}_{1} \coloneqq\phi_{1}(\mathbf{B}_{1}^{\mathsf{T}}\boldsymbol{x})$,
the residual layers are defined as follows for even $j=2,4,\dotsc,l-1$:
\begin{equation}
\boldsymbol{w}_{j} \coloneqq\phi_{j}(\mathbf{B}_{j}^{\mathsf{T}}\boldsymbol{w}_{j-1}),\,
\boldsymbol{w}_{j+1} \coloneqq\phi_{j+1}(\boldsymbol{w}_{j-1}+\mathbf{B}_{j+1}^{\mathsf{T}}\boldsymbol{w}_{j})
\label{eq:residual_act}
\end{equation}
In comparison to chain networks, no additional parameters are
required; the only difference is the 
addition of $\boldsymbol{w}_{j-1}$ in the argument of $\phi_{j+1}$.
As a special case of Eq.~\ref{eq:dense}, we interpret the activations in Eq.~\ref{eq:residual_act} as approximate  
solutions to the deep frame approximation problem:
\begin{gather}
\underset{\{\boldsymbol{w}_{j}\}}{\arg\min}\,\tfrac{1}{2}\big\lVert \boldsymbol{x}-\mathbf{B}_{1}\boldsymbol{w}_{1}\big\rVert _{2}^{2} + \sum_{j=1}^{l}\Phi_{j}(\boldsymbol{w}_{j}) \label{eq:residual_opt} 
\\[-0.5em]+\,\tfrac{1}{2}\hspace{-0.3em}\sum_{j=1}^{\lfloor l/2 \rfloor}\hspace{-0.2em}\big\lVert \boldsymbol{w}_{2j}{-}\mathbf{B}_{2j}^\mathsf{T}\boldsymbol{w}_{2j\text{-}1}\big\rVert_{2}^{2} 
+\big\lVert \boldsymbol{w}_{2j\text{+}1}{-}\boldsymbol{w}_{2j\text{-}1}{-}\mathbf{B}_{2j\text{+}1}^{\mathsf{T}}\boldsymbol{w}_{2j}\big\rVert _{2}^{2}
\nonumber
\end{gather}
This results in the induced frame structure 
of Eq.~\ref{eq:dense_structure} with $\mathbf{B}_{jj}=\mathbf{I}$
for $j>1$, $\mathbf{B}_{jk}=\mathbf{0}$ for $j>k+1$, $\mathbf{B}_{jk}=\mathbf{0}$ for $j>k$ with odd $k$, and 
$\mathbf{B}_{jk}=\mathbf{I}$ for $j>k$ with even $k$.

\subsubsection{Densely Connected Convolutional Networks}

Building upon the empirical successes of residual networks, densely connected convolution networks~\cite{huang2017densely} incorporate 
skip connections between earlier layers as well. This is shown in Eq.~\ref{eq:densenet_act} 
where the transformation $\mathbf{B}_j$ of concatenated variables $[\boldsymbol{w}_{k}]_k$ for $k=1,\dotsc,j-1$ 
is equivalently 
written as the summation of smaller transformations $\mathbf{B}_{jk}$.
\begin{equation}
\boldsymbol{w}_{j} \coloneqq\phi_{j}\Big(\mathbf{B}_{j}^{\mathsf{T}}\left[\boldsymbol{w}_{k}\right]_{k=1}^{j-1}\Big)=\phi_{j}\Big(\sum_{k=1}^{j-1}\mathbf{B}_{jk}^{\mathsf{T}}\boldsymbol{w}_{k}\Big)
\label{eq:densenet_act}
\end{equation}
These activations again provide approximate solutions to the problem in Eq.~\ref{eq:dense} with 
the induced frame structure of Eq.~\ref{eq:dense_structure} where 
$\mathbf{B}_{jj}=\mathbf{I}$ for $j>1$ and the lower blocks $\mathbf{B}_{jk}$ for $j>k$ are all 
filled with learned parameters. 

Skip connections enable effective learning 
in much deeper networks than chain-structured alternatives. While originally motivated from the perspective of 
improving optimization by introducing shortcuts~\cite{he2016deep}, adding more connections between layers 
was also shown to 
improve generalization performance~\cite{huang2017densely}.
As compared in Fig.~\ref{fig:gram}, denser skip connections induce frame structures with denser Gram 
matrices allowing for lower mutual coherence. This suggests that architectures' capacities for low 
validation error can be quantified and compared based on their capacities for 
inducing frames with low minimum mutual coherence. 

\subsubsection{Global Iterative Inference} \label{sec:iterative_deep}

As with multilayer sparse coding, inference approximation accuracy for deep frame approximation can be improved with 
recurrent connections implementing an optimization algorithm. However, instead of composing the 
solutions for each layer in succession, the latent variables of all layers must be jointly inferred 
by solving a single global optimization problem. In deep component analysis~\cite{murdock2018deep}, the 
alternating direction method of multipliers was used, but this required unrealistic assumptions for efficient
iterations. While standard proximal gradient descent could be applied to 
the concatenated variables with the global frame, this would be inefficient since it does not take advantage 
of the lower-block-triangular structure. 

Instead, we use prox-linear block coordinate descent~\cite{xu2013block}, which
cyclically updates subsets of variables corresponding to each layer. Viewed as a recurrent neural 
network, these operations implement feedback connections to ensure consistency between layers.
This algorithm was successfully
applied to a similar problem for chain-structured networks in Chodosh et al.~\cite{chodosh2018deep}.

A single iteration of this algorithm incrementally updates the latent variables for each layer as:
\begin{equation}
\boldsymbol{w}_{j}^{[t]}=\phi_{j}\big(\boldsymbol{w}_{j}^{[t-1]}-\gamma_{j}\boldsymbol{g}_{j}^{[t]}\big),\quad\forall j=1,\dotsc,l
\label{eq:bcd}
\end{equation}
These updates take the same form as proximal gradient descent with step sizes $\gamma_j$, where the 
$\boldsymbol{g}_{j}^{[t]}$ is defined for networks with skip connections using the general global 
structured frame from Eq.~\ref{eq:dense_structure} as:
\begin{align}
\boldsymbol{g}_{j}^{[t]} & =\frac{\partial}{\partial\boldsymbol{w}_{j}}\ell(\boldsymbol{w}_{1}^{[t]},\dotsc,\boldsymbol{w}_{j-1}^{[t]},\boldsymbol{w}_{j},\boldsymbol{w}_{j+1}^{[t-1]},\dotsc,\boldsymbol{w}_{l}^{[t-1]})\\
&
\begin{aligned}
=\mathbf{B}_{j}^{\mathsf{T}}\Big(\mathbf{B}_{j}\boldsymbol{w}_{j}-\sum_{k=1}^{j-1}\mathbf{B}_{jk}^{\mathsf{T}}\boldsymbol{w}_{k}^{[t]}\Big) +\sum_{j^{\prime}=j+1}^{l}\mathbf{B}_{j^{\prime}j}\Big(\mathbf{B}_{j^{\prime}j}^{\mathsf{T}}\boldsymbol{w}_{j}\\
+\sum_{k^{\prime}=1}^{j-1}\mathbf{B}_{j^{\prime}k^{\prime}}^{\mathsf{T}}\boldsymbol{w}_{k^{\prime}}^{[t]}+\sum_{k^{\prime}=j+1}^{j^{\prime}-1}\mathbf{B}_{j^{\prime}k^{\prime}}^{\mathsf{T}}\boldsymbol{w}_{k^{\prime}}^{[t-1]}-\mathbf{B}_{j^{\prime}}\boldsymbol{w}_{j^{\prime}}^{[t-1]}\Big)
\end{aligned} \nonumber
\label{eq:bcd_grad}
\end{align}
This is the partial gradient of the smooth reconstruction error term $\ell$ of the loss function 
from Eq.~\ref{eq:chain_opt}. Variables from previous layers are updated immediately in the current iteration 
facilitating faster convergence. The step sizes $\gamma_j$ may be 
selected automatically by bounding the 
Lipschitz constants of the corresponding gradients~\cite{chodosh2018deep}. 
Inference can be further accelerated by learning less conservative step sizes as
trainable parameters and using extrapolation~\cite{xu2013block}. 

This optimization
algorithm is unrolled to a fixed number of iterations for improved inference approximation accuracy. 
The resulting recurrent networks 
achieve similar performance to feed-forward approximations when trained on discriminative tasks 
like image classification. This suggests similar representational capacities and motivates the
indirect analysis of deep neural networks with the framework of deep frame approximation. 
In addition, while the computational overhead of recurrent inference may be impractical in these cases, 
it can more effectively enforce other constraints~\cite{murdock2018deep}
and improve adversarial robustness~\cite{cazenavette2021adversarial}.

\subsection{The Deep Frame Potential}

We propose to use lower bounds on the mutual coherence of induced 
structured frames for the data-independent comparison of feed-forward deep network architecture capacities. 
Note that while one-sided coherence is better suited to nonnegativity constraints, it has the 
same lower bound~\cite{bruckstein2008uniqueness}.
Directly optimizing mutual coherence from Eq.~\ref{eq:mutual_coherence}
is difficult due to its piecewise structure. Instead, we consider a  
tight lower by replacing the maximum off-diagonal element of the deep frame 
Gram matrix $\mathbf{G}$ with the mean.
This gives the lower bound from Eq.~\ref{eq:average_fp}, a normalized version of the frame potential
$FP(\mathbf{B})=\left\lVert \mathbf{G}\right\rVert_{F}^{2}$ from Eq.~\ref{eq:frame_potential}
that is strongly-convex and can be optimized more effectively~\cite{benedetto2003finite}.
Due to the block-sparse structure of the induced frames from Eq.~\ref{eq:dense_structure}, we  evaluate the frame potential in terms of local blocks $\mathbf{G}_{jj'}\in\mathbb{R}^{k_j\times k_{j'}}$ that are nonzero
only if layer $j$ is connected to layer $j'$.

To compute the Gram matrix, we first need to normalize the global induced frame $\mathbf{B}$
from Eq.~\ref{eq:dense_structure}.
By stacking the column magnitudes of layer $j$ as the elements in the diagonal matrix 
$\mathbf{C}_{j}=\mathrm{diag}(\boldsymbol{c}_{j})\in\mathbb{R}^{k_{j}\times k_{j}}$, the normalized 
parameters can be represented as $\tilde{\mathbf{B}}_{ij} = \mathbf{B}_{ij}\mathbf{C}_{j}^{-1}$. 
Similarly, the squared norms of the full set of columns in the global frame $\mathbf{B}$ are  
$\mathbf{N}_{j}^{2}=\sum_{i=j}^{l}\mathbf{C}_{ij}^{2}$.
The full normalized frame can then be found as  $\tilde{\mathbf{B}}=\mathbf{B}\mathbf{N}^{-1}$
where the matrix $\mathbf{N}$ is block diagonal with $\mathbf{N}_{j}$ as its blocks.
The blocks of the Gram matrix 
$\mathbf{G}=\tilde{\mathbf{B}}^\mathsf{T}\tilde{\mathbf{B}}$ are then given as:
\begin{equation}
\mathbf{G}_{jj'}=\sum_{i=j'}^{l}\mathbf{N}_{j}^{-1}\mathbf{B}_{ij}^{\mathsf{T}}\mathbf{B}_{ij'}\mathbf{N}_{j'}^{-1}
\label{eq:G_block}
\end{equation}
For chain networks, $\mathbf{G}_{jj'}\neq\mathbf{0}$ only when $j'=j+1$, which represents the 
connections between adjacent layers. In this case, the blocks can be simplified as:
\begin{align}
\mathbf{G}_{jj} & =(\mathbf{C}_{j}^{2}+\mathbf{I})^{-\frac{1}{2}}({\mathbf{B}}_{j}^{\mathsf{T}}{\mathbf{B}}_{j}+\mathbf{I})(\mathbf{C}_{j}^{2}+\mathbf{I})^{-\frac{1}{2}}\\
\mathbf{G}_{j(j+1)} & =-(\mathbf{C}_{j}^{2}+\mathbf{I})^{-\frac{1}{2}}{\mathbf{B}}_{j+1}(\mathbf{C}_{j+1}^{2}+\mathbf{I})^{-\frac{1}{2}}\label{eq:G_block_chain}\\
\mathbf{G}_{ll}&={\mathbf{B}}_{l}^{\mathsf{T}}{\mathbf{B}}_{l}
\end{align}

Because the diagonal is removed in the deep frame potential computation, the contribution of $\mathbf{G}_{jj}$ 
is simply a rescaled version of the local frame potential of layer $j$. The contribution of $\mathbf{G}_{j(j+1)}$,
on the other hand, can essentially be interpreted as rescaled $\ell_2$ weight decay where rows are
weighted more heavily if the corresponding columns of the previous layer's parameters have higher magnitudes.
Furthermore, since the global frame potential is averaged over the total number of nonzero elements in $\mathbf{G}$,
if a layer has more parameters, then it will be given more weight in this computation. 
For general networks with skip connections, the summation from Eq.~\ref{eq:G_block} 
has additional terms that introduce 
more complicated interactions. In these cases, it cannot be evaluated from local properties of  layers.

Essentially, the deep frame potential summarizes the structural properties of
the global frame $\mathbf{B}$
induced by a deep network architecture by balancing interactions within each individual layer through
local coherence properties and between connecting layers.

\subsubsection{Theoretical Lower Bound for Chain Networks}

While the deep frame potential is a function of parameter values, its minimum value is determined 
only by the architecture-induced frame structure. Furthermore, we know that
it must be bounded by a nonzero constant for overcomplete frames. 
In this section, we 
derive this lower bound for the special case of fully-connected chain networks
and provide intuition for why skip connections increase the capacity for low mutual coherence.

First, observe that a lower bound for the Frobenius norm of $\mathbf{G}_{j(j+1)}$ from 
Eq.~\ref{eq:G_block_chain} cannot be readily attained because the rows and columns are
rescaled independently. This means that a lower bound for the norm 
of $\mathbf{G}$ must be found by jointly considering the entire matrix structure, 
not simply through
the summation of its components. To accomplish this, we instead consider the matrix 
$\mathbf{F}=\tilde{\mathbf{B}}\tilde{\mathbf{B}}^{\mathsf{T}}$, which is full rank and has the same norm as 
$\mathbf{G}$:
\begin{equation}
\lVert \mathbf{G} \rVert _{F}^{2}=\lVert \mathbf{F} \rVert _{F}^{2}=\sum_{j=1}^{l}\left\lVert \mathbf{F}_{jj}\right\rVert _{F}^{2}+2\sum_{j=1}^{l-1}\left\lVert \mathbf{F}_{j(j+1)}\right\rVert _{F}^{2}
\label{eq:H_norm}
\end{equation}
We can then express the individual blocks of $\mathbf{F}$ as:
\begin{align}
\mathbf{F}_{11} & ={\mathbf{B}}_{1}(\mathbf{C}_{1}^{2}+\mathbf{I}_{k_{1}})^{-1}{\mathbf{B}}_{1}^{\mathsf{T}}\\
\mathbf{F}_{jj} & ={\mathbf{B}}_{j}(\mathbf{C}_{j}^{2}+\mathbf{I})^{-1}{\mathbf{B}}_{j}^{\mathsf{T}}+(\mathbf{C}_{j-1}^{2}+\mathbf{I})^{-1}\\
\mathbf{F}_{j(j+1)} & =-{\mathbf{B}}_{j}(\mathbf{C}_{j}^{2}+\mathbf{I})^{-1}
\label{eq:Hjj}
\end{align}
In contrast to $\mathbf{G}_{j(j+1)}$ in Eq.~\ref{eq:G_block_chain}, only the columns of 
$\mathbf{F}_{j(j+1)}$ in Eq.~\ref{eq:Hjj} are rescaled.
Since $\tilde{\mathbf{B}}_j$ has normalized columns, its norm can be exactly expressed
as:
\begin{equation}
\left\Vert \mathbf{F}_{j(j+1)}\right\rVert _{F}^{2} =\sum_{n=1}^{k_{j}}\Bigg(\frac{c_{jn}}{c_{jn}^{2}+1}\Bigg)^{2}
\label{eq:H_jk}
\end{equation}

For the other blocks, we find lower bounds for their norms through the same technique used in deriving
the Welch bound in Sec.~\ref{sec:welch}.
Specifically, we can lower bound the norms of the individual blocks as:
\begin{align}
\left\lVert \mathbf{F}_{11}\right\rVert _{F}^{2} & \geq\frac{1}{k_{0}}\Bigg(\sum_{n=1}^{k_{1}}\frac{c_{1n}^{2}}{c_{1n}^{2}+1}\Bigg)^{2}\\
\left\lVert \mathbf{F}_{jj}\right\rVert _{F}^{2} & \geq\frac{1}{k_{j-1}}\Bigg(\sum_{n=1}^{k_{j}}\frac{c_{jn}^{2}}{c_{jn}^{2}+1}+\sum_{p=1}^{k_{j-1}}\frac{1}{c_{(j-1)p}^{2}+1}\Bigg)^{2}\nonumber
\label{eq:Hjj_bound}
\end{align}

In this case of dense shallow frames, the Welch bound depends only on the data dimensionality
and the number of frame components. However, due to the structure
of the architecture-induced frames, the lower bound of the deep frame potential depends on the data
dimensionality, the number of layers, the number of units in each layer, the connectivity between layers,
and the relative magnitudes between layers. Skip connections increase the number of nonzero elements
in the Gram matrix over which to average and also enable off-diagonal blocks to have lower norms.

\subsubsection{Model Selection by Empirical Minimization}

For more general architectures that lack a simple theoretical lower bound, we instead  
bound the mutual coherence of the architecture-induced frame through 
empirical minimization of the deep frame potential in the lower bound from Eq.~\ref{eq:average_fp}. 
The lack of suboptimal local minima allows for effective optimization
using gradient descent. Because it is independent of data and
parameter instantiations, this provides a way to compare different architectures without training. Instead, model 
selection is performed by choosing the candidate architecture that achievest the lowest deep frame potential
subject to desired modeling constraints such as limiting the total number of parameters. 
 
\section{Experimental Results} \label{sec:experiments}

We experimentally validate the deep frame approximation framework
on the MNIST~\cite{lecun1998gradient}, CIFAR-10/100~\cite{krizhevsky2009learning}, and ImageNet-1k~\cite{russakovsky2015imagenet}
datasets across a wide variety of chain, residual, and densely connected network architectures.
For our ImageNet experiments, we adapt the standard ResNet-18 architecture~\cite{he2016deep} scaling depth and width 
by varying both the number of residual blocks in each group between 1 and 5 and the base number 
of filters between 16 and 96. 
For our smaller MNIST and CIFAR experiments, we adapt the 
simplified ResNet architecture from~\cite{wu2016tensorpack}, which consists of a single convolutional 
layer followed 
by three groups of residual blocks with activations given by Eq.~\ref{eq:residual_act}. 
We modify their depths by changing the number of residual blocks 
in each group between 2 and 10 and their widths by changing the base number of filters between 
4 and 32.
For our experiments with densely connected skip connections~\cite{huang2017densely}, we adapt the
simplified DenseNet architecture from~\cite{li2018densepack}. Like with residual networks, 
it consists of a convolutional layer followed by three groups of the activations 
from Eq.~\ref{eq:densenet_act}
with decreasing 
spatial resolutions and increasing numbers of filters. 
Network depth and width are modified by increasing 
the number of layers per group and the base growth rate both between 2 and 12. Batch normalization~\cite{ioffe2015batch} 
was also used in all convolutional networks.

\subsection{Iterative Deep Frame Approximation}

\begin{figure}
\centering
\subfloat[Chain Network]{
\includegraphics[width=0.32\columnwidth]{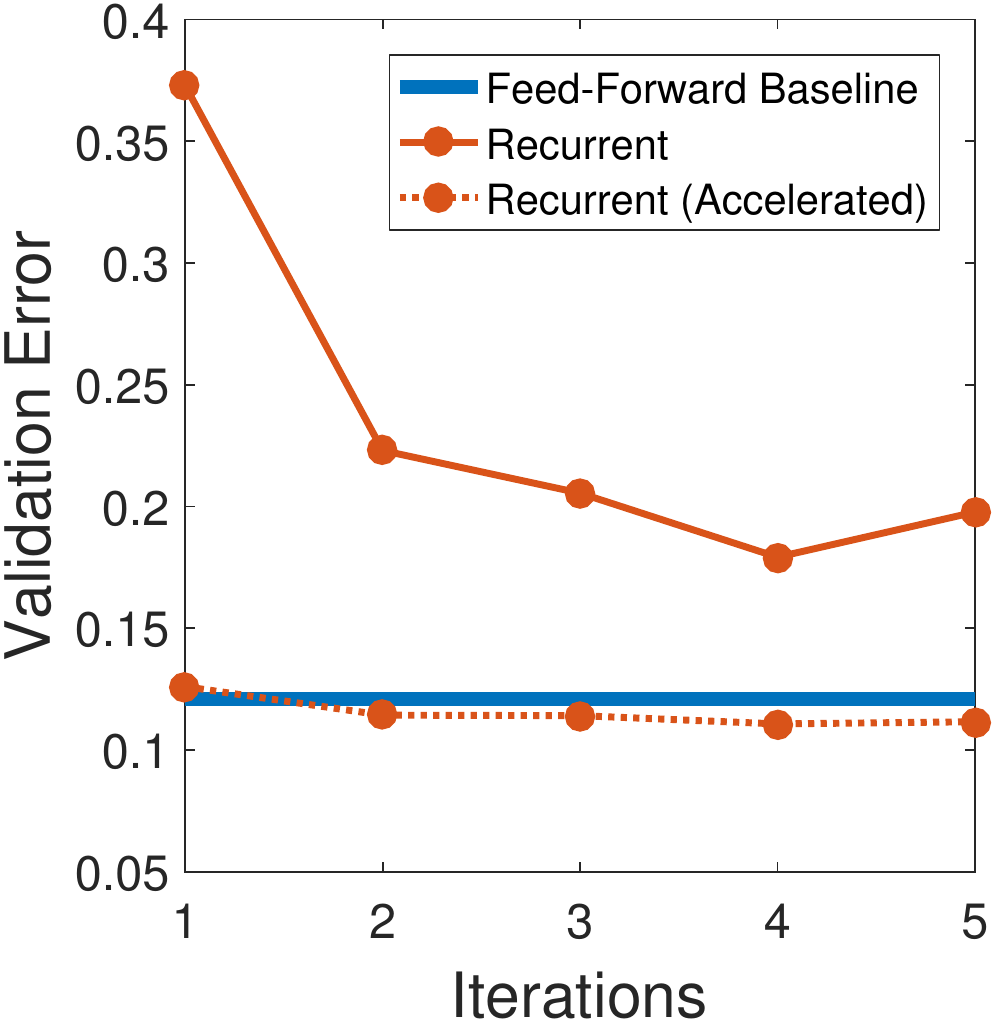}
}
\subfloat[ResNet]{
\includegraphics[width=0.32\columnwidth]{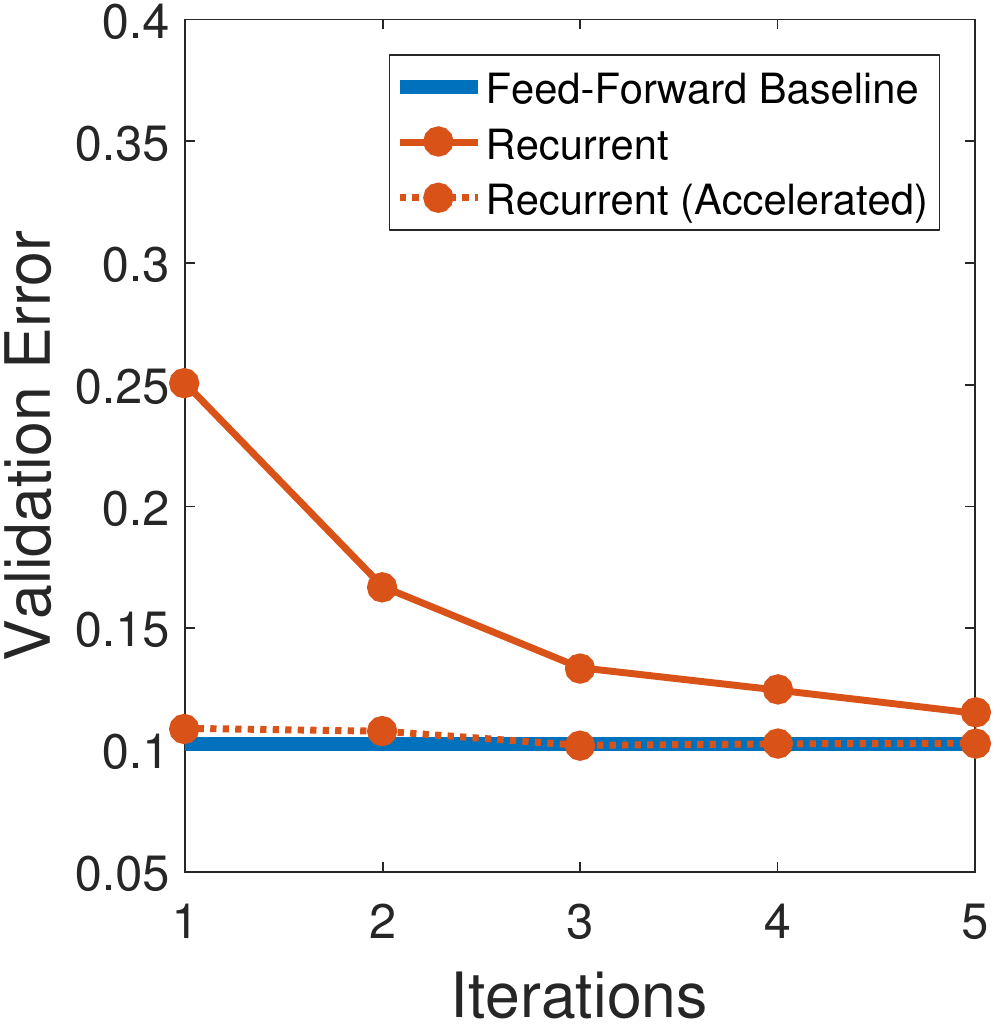}
}
\subfloat[DenseNet]{
\includegraphics[width=0.32\columnwidth]{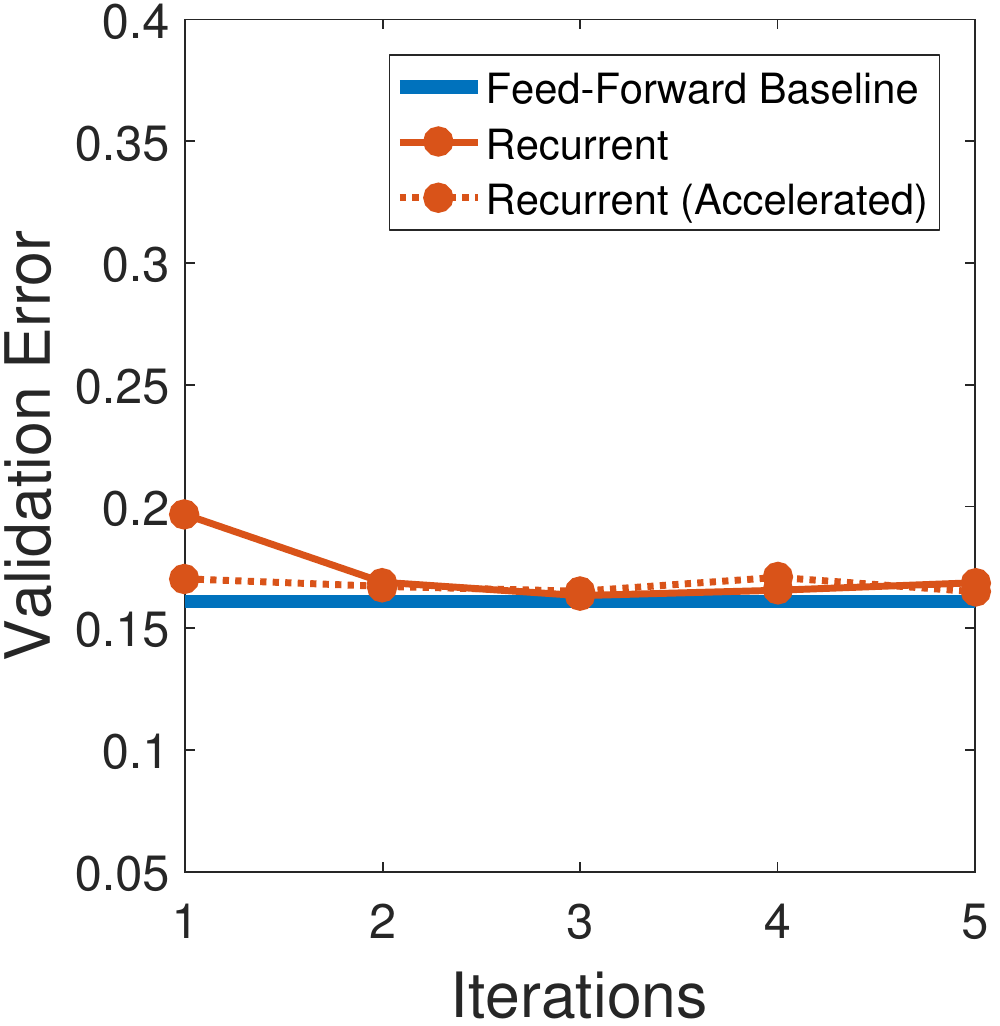}
}
\caption{
Deep frame approximation inference compared between feed-forward deep networks and recurrent optimization 
for (a) chain networks and (b) ResNets with a base width of 16 filters and 2 residual blocks per group, 
and (c) DenseNets with a growth rate of 4 and 4 layers per group. 
Unrolled to an increasing number of iterations, recurrent optimization
networks quickly converge to validation accuracies comparable to their feed-forward approximations. 
}
\label{fig:iter}
\end{figure}

We begin by experimentally validating our motivating assumption that feed-forward deep
networks may be 
analyzed indirectly as approximate techniques for deep frame approximation. 
We show in Fig.~\ref{fig:iter} that recurrent networks that explicitly optimize the inference objective function from
Eq.~\ref{eq:dense_opt} are able to achieve generalization 
capacities that are comparable to feed-forward alternatives.
The need for multiple iterations makes learning  
more challenging and inference less efficient, especially for deeper networks, but the acceleration techniques from 
Sec.~\ref{sec:iterative_deep} speed up convergence so that fewer iterations are required.

While the multiple iterations of recurrent inference have little effect on validation error,
they can substantially improve robustness to adversarial attacks. For the sake of brevity, we refer the reader to \cite{cazenavette2021adversarial} for details on the adversarial attacks themselves. 
Fig. \ref{fig:adversarial}c shows that adiditional iterations reduce adversarial error for both with and without residual connections. For chain networks, we see an iteration threshold after which adversarial robustness significantly improves.
For residual networks, however, we immediately see large performance gains with just a few iterations due to the added skip connections.

\subsection{Feed-Forward Analysis with Deep Frame Potentials}

Next, we show consistent 
correlations between validation error and minimum deep frame potential across different families of architectures
and datasets. This validates the use of deep frame approximation as a proxy for 
analyzing feed-forward deep neural networks.

\begin{figure}
\centering
\subfloat[Without Regularization]{
\includegraphics[width=0.48\columnwidth]{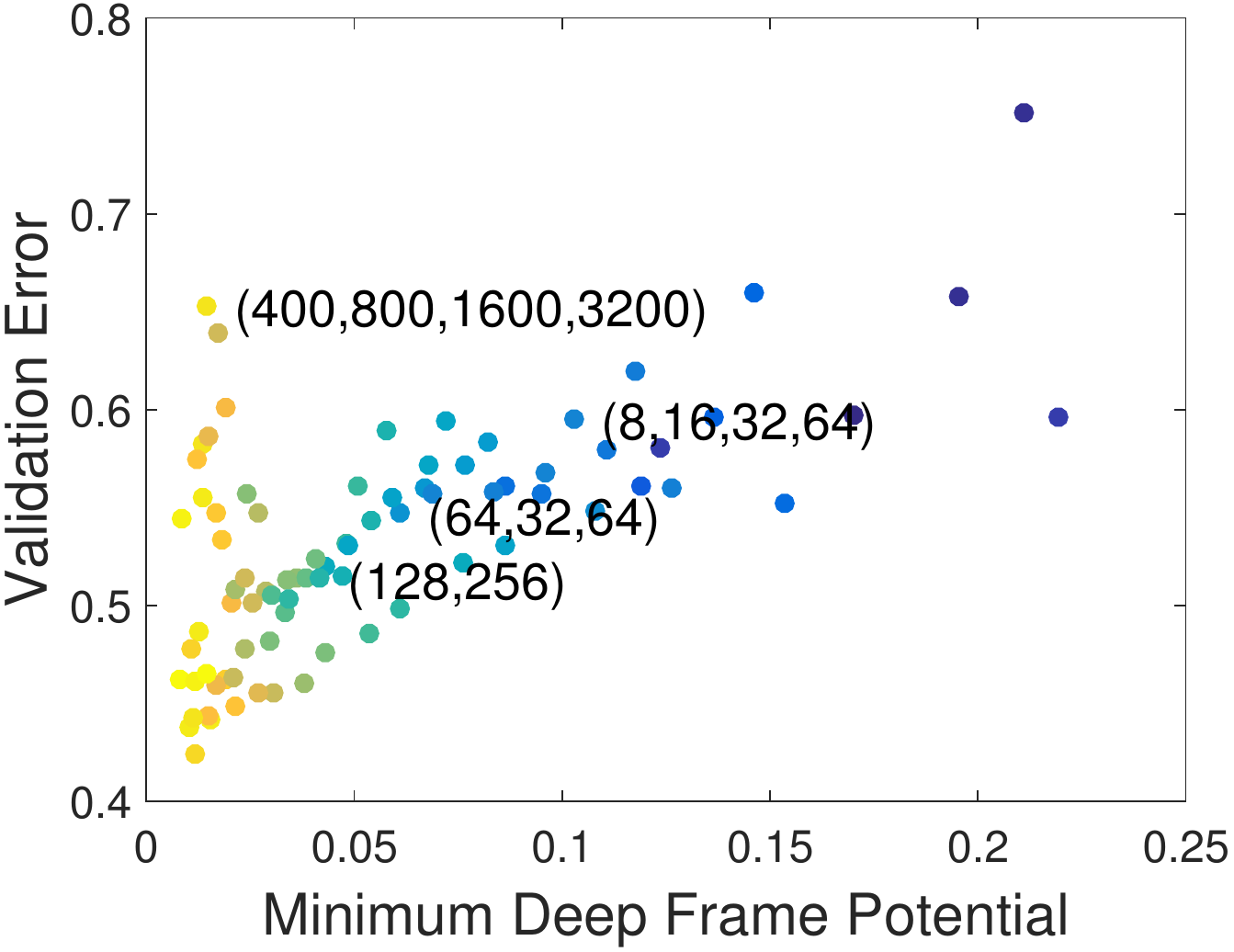}
\label{fig:regularization_without}
}
\subfloat[With Regularization]{
\includegraphics[width=0.48\columnwidth]{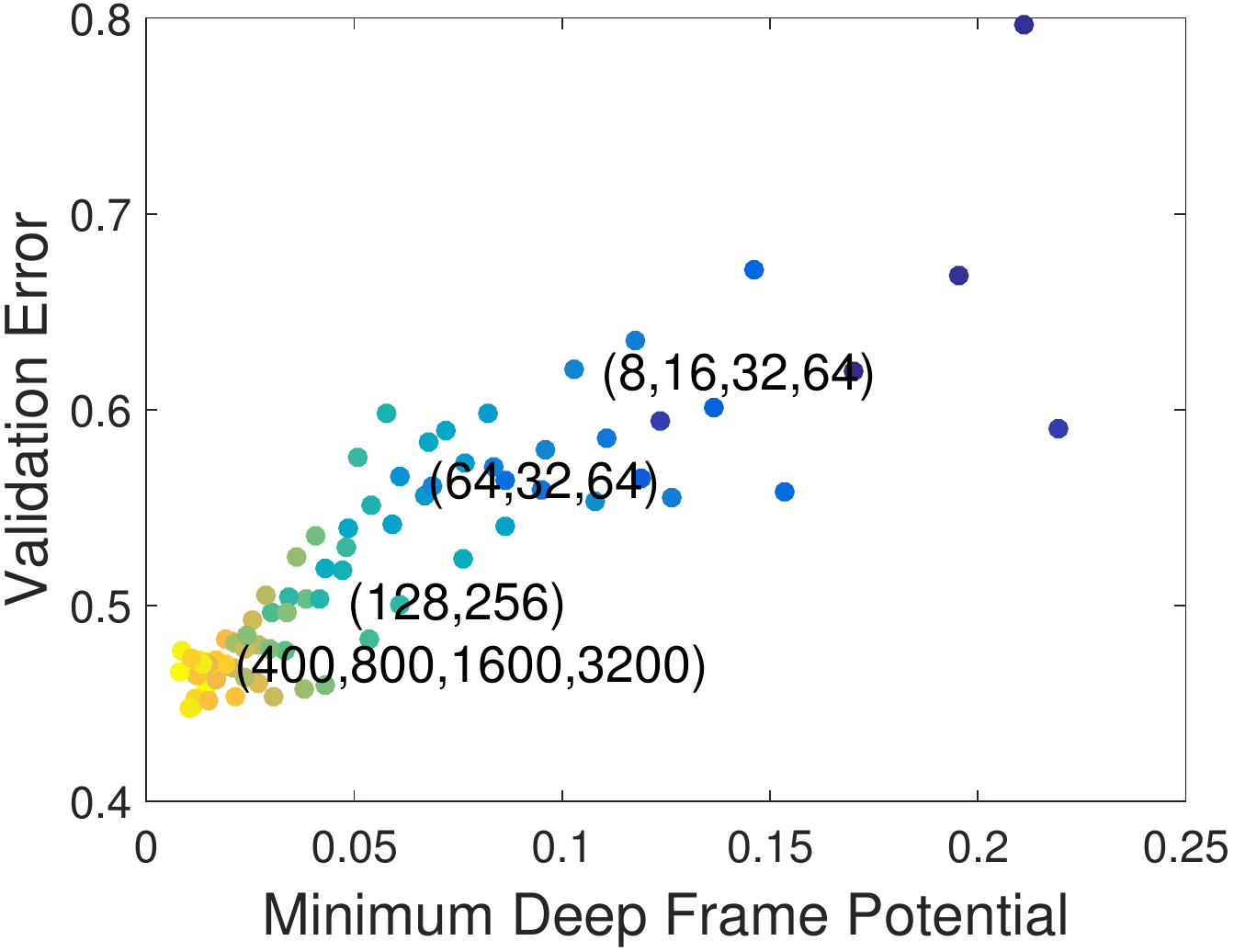}
\label{fig:regularization_with}
}
\caption{
A comparison of fully-connected 
deep network architectures with varying depths and widths. Warmer colors indicate models with more total parameters. 
(a) Some very large networks cannot be trained effectively resulting in unusually high validation errors. 
(b) This can be remedied with deep frame potential regularization, resulting in high correlation between 
minimum frame potential and validation error.
}
\label{fig:regularization}
\end{figure}

In Fig.~\ref{fig:regularization}, we visualize a scatter plot of trained  
fully-connected networks with three to five layers and 16 to 4096 units 
in each layer. The corresponding architectures are shown as a list of units per layer
for a few representative examples. 
The minimum deep frame potential of each architecture is compared against its validation 
error after training, and the total parameter count is indicated by color.
In Fig.~\ref{fig:regularization_without}, some networks with many parameters have
unusually high error due to the difficulty in training very large fully-connected networks. 
In Fig.~\ref{fig:regularization_with}, the addition of a deep frame
potential regularization term overcomes some of these optimization difficulties for improved 
parameter efficiency. This results in high
correlation between minimum deep frame potential and validation error. Furthermore, it emphasizes the  
diminishing returns of increasing the size of fully-connected chain networks; after a certain point, adding more 
parameters does little to reduce both validation error and minimum frame potential.

\begin{figure}
\centering

\subfloat[Validation Error]{
\includegraphics[width=0.48\columnwidth]{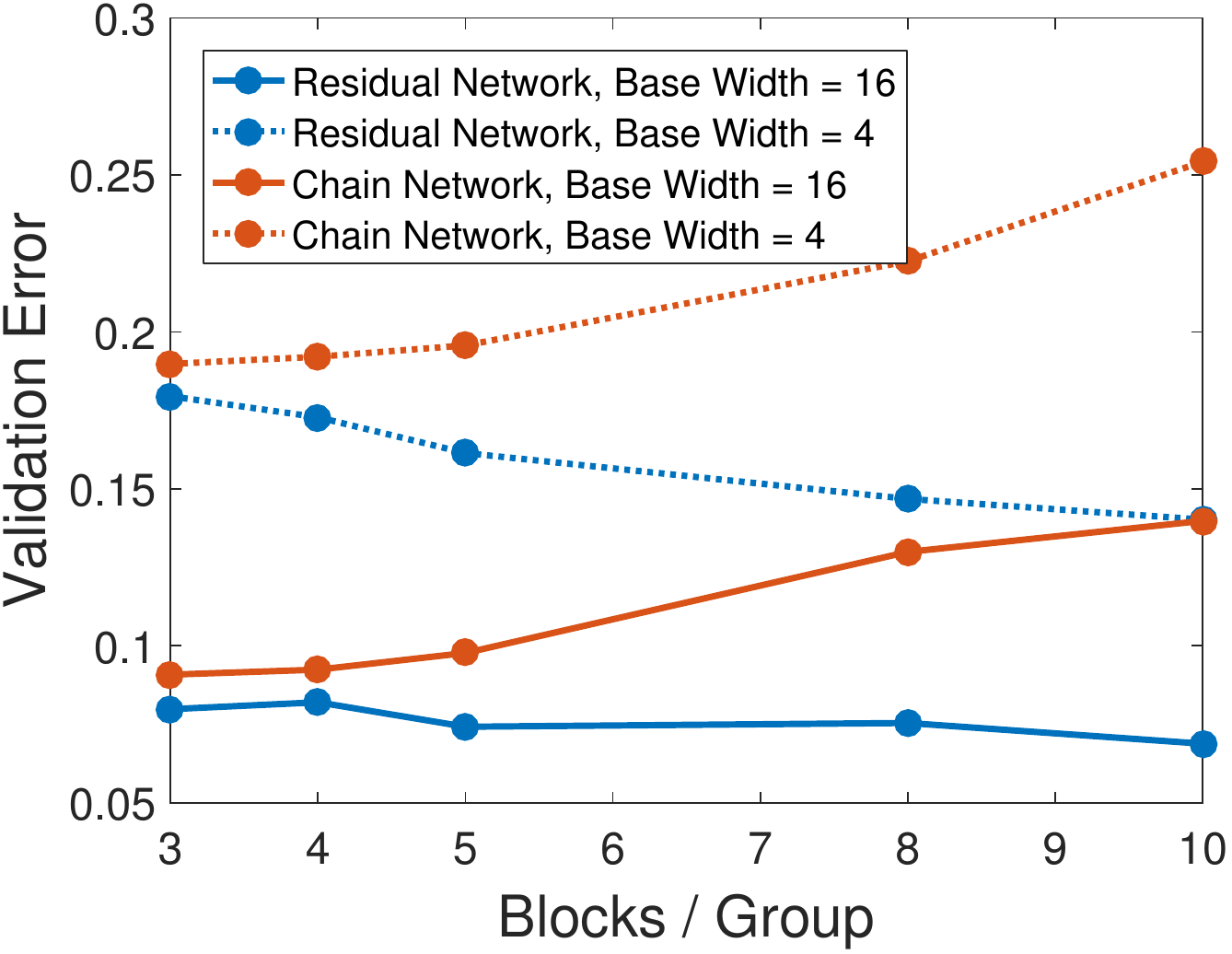}
\label{fig:resnet_layers_error}
}
\subfloat[Minimum Deep Frame Potential]{
\includegraphics[width=0.48\columnwidth]{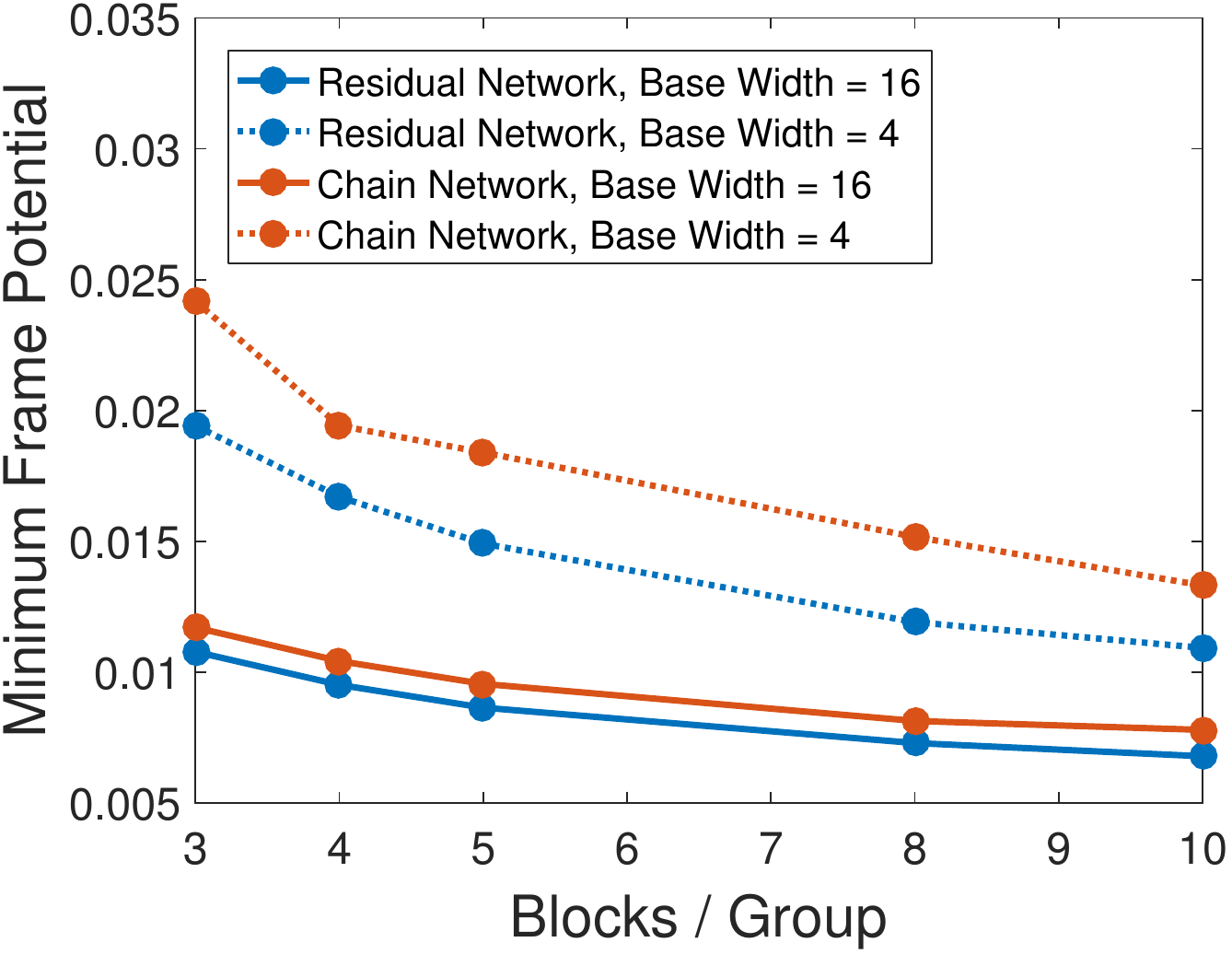}
\label{fig:resnet_layers_frame}
}
\caption{
The effect of increasing depth in chain and residual networks.
Validation error is compared against network depth for two different network widths. (a) Unlike  
chain networks, even very deep residual networks can be trained effectively for decreased
validation error. (b) Despite having the same number of total parameters, residual connections also 
induce frame structures with lower minimum deep frame potentials.
}
\label{fig:resnet_layers}
\end{figure}

\begin{figure}
\centering

\subfloat[Validation Error]{
\includegraphics[width=0.48\columnwidth]{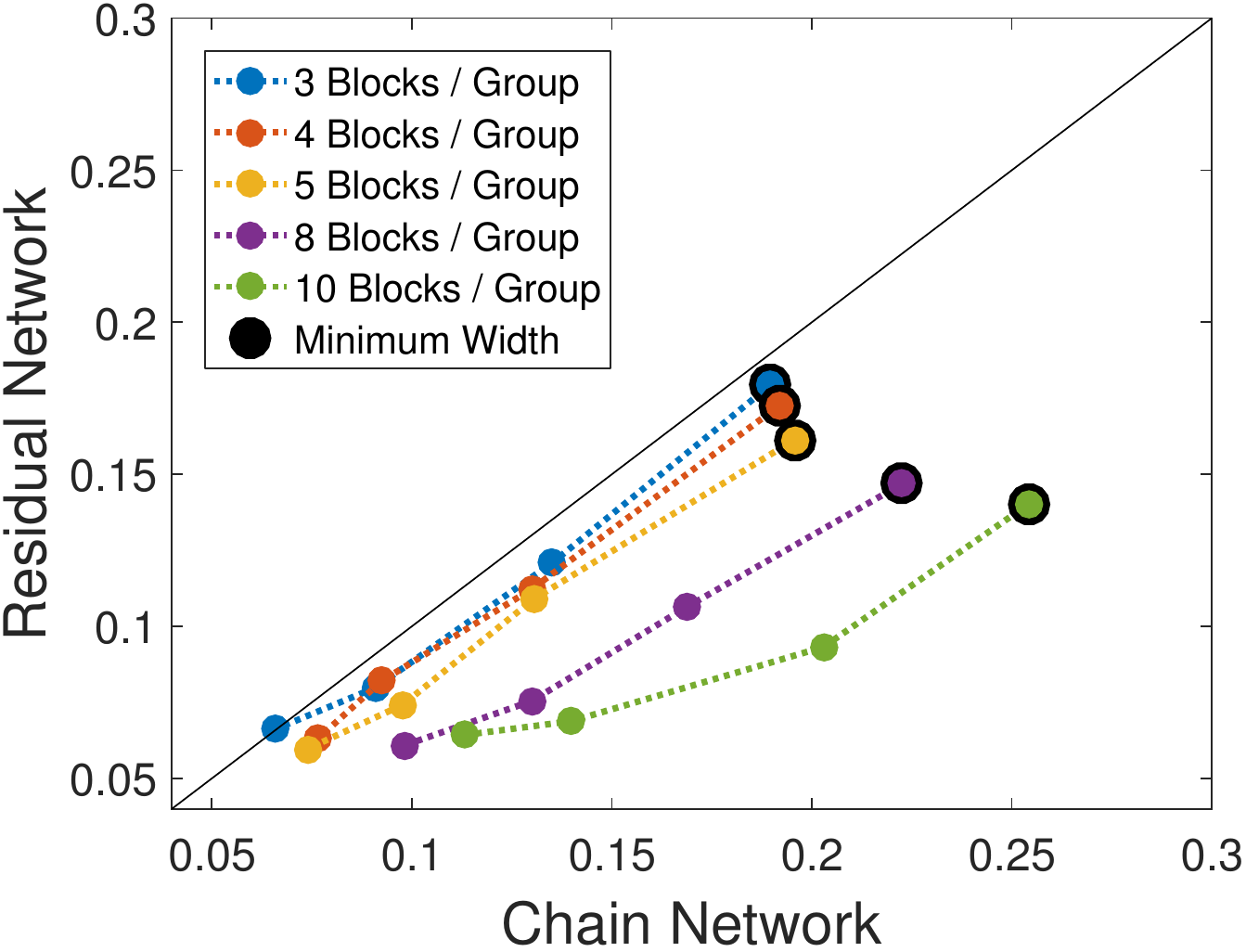}
}
\subfloat[Minimum Deep Frame Potential]{
\includegraphics[width=0.48\columnwidth]{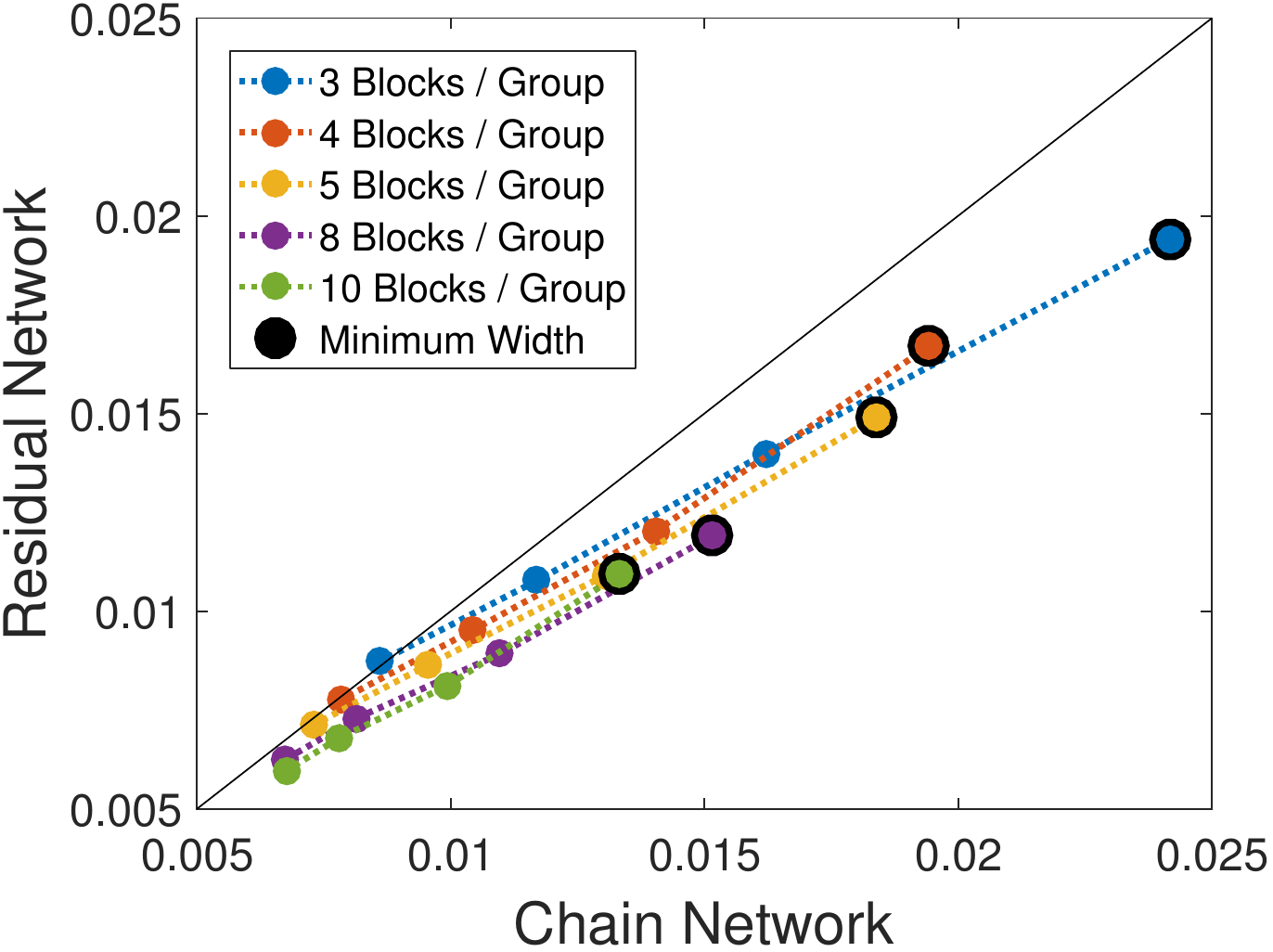}
}
\caption{
A comparison of (a) validation error and (b) minimum frame potential between  
residual networks and chain networks. Colors indicate different depths 
and datapoints are connected in order of increasing widths of 4, 8, 16, or 32 filters. 
Skip connections result in 
reduced error correlating with frame potential with dense networks showing 
superior efficiency with increasing depth.
}
\label{fig:scatter}
\end{figure}

Even without additional parameters, residual connections reduce both validation error and 
minimum deep frame potential. Fig.~\ref{fig:resnet_layers} 
compares validation errors and minimum deep frame potentials of
residual networks and chain networks with residual connections removed. 
In Fig.~\ref{fig:resnet_layers_error}, chain network validation error increases for  
deeper networks while that of residual networks is lower and consistently decreases. 
This emphasizes the difficulty in training very deep chain networks. 
In Fig.~\ref{fig:resnet_layers_frame}, residual connections are shown to enable lower minimum deep frame 
potentials following a similar trend with respect to model size.

In Fig.~\ref{fig:scatter}, we compare chain networks and residual 
networks with exactly the same number of parameters, where color 
indicates the number of residual blocks per group and connected data points 
have the same depths but different widths.
The addition
of skip connections reduces both validation error and minimum frame potential, as visualized 
by consistent placement below the diagonal line indicating lower validation errors for residual networks than 
comparable chain networks.
This effect becomes even more pronounced with increasing depths and widths. 

\begin{figure*}
\centering
\subfloat[MNIST]{
\includegraphics[width=0.24\linewidth]{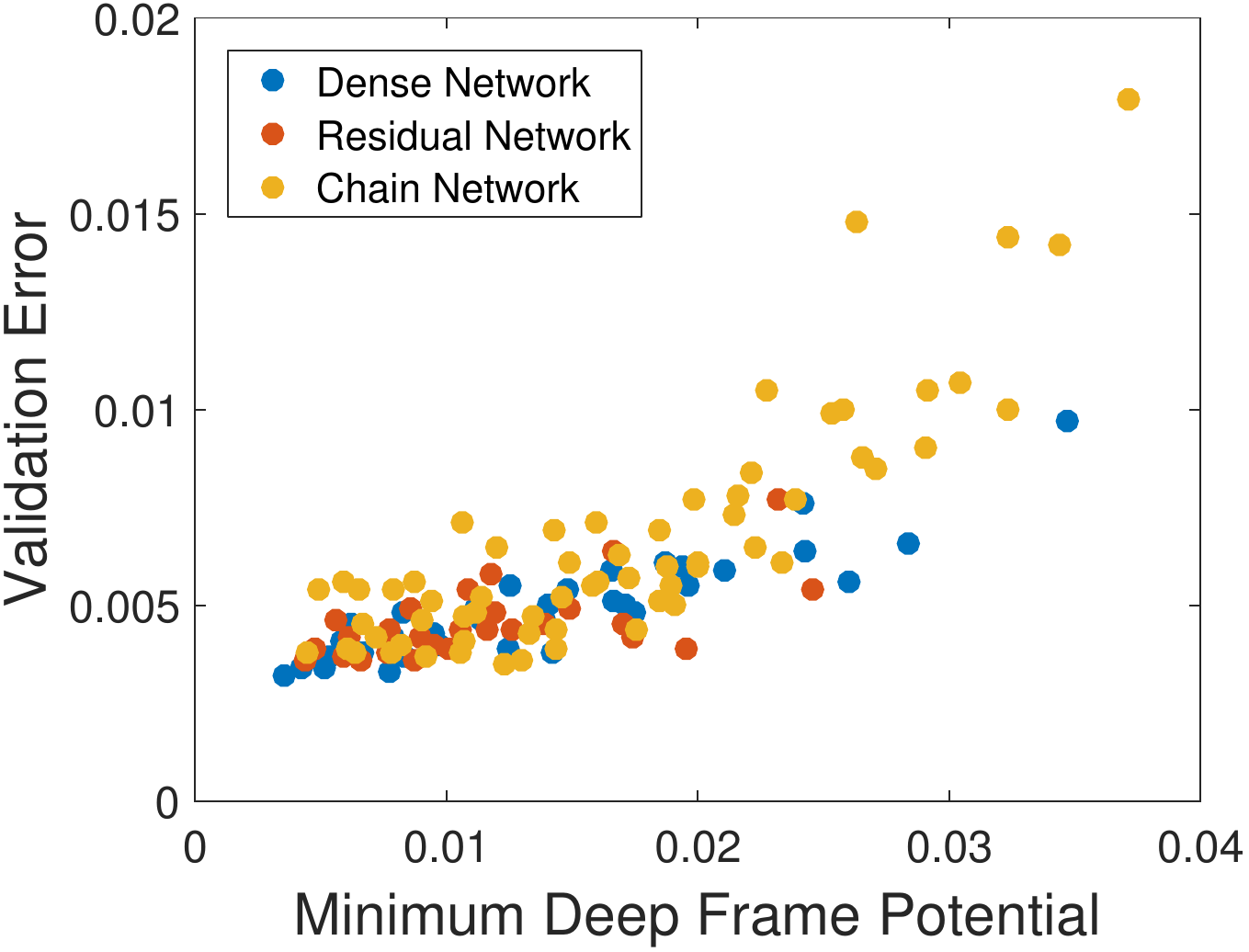}
}
\subfloat[CIFAR-10]{
\includegraphics[width=0.24\linewidth]{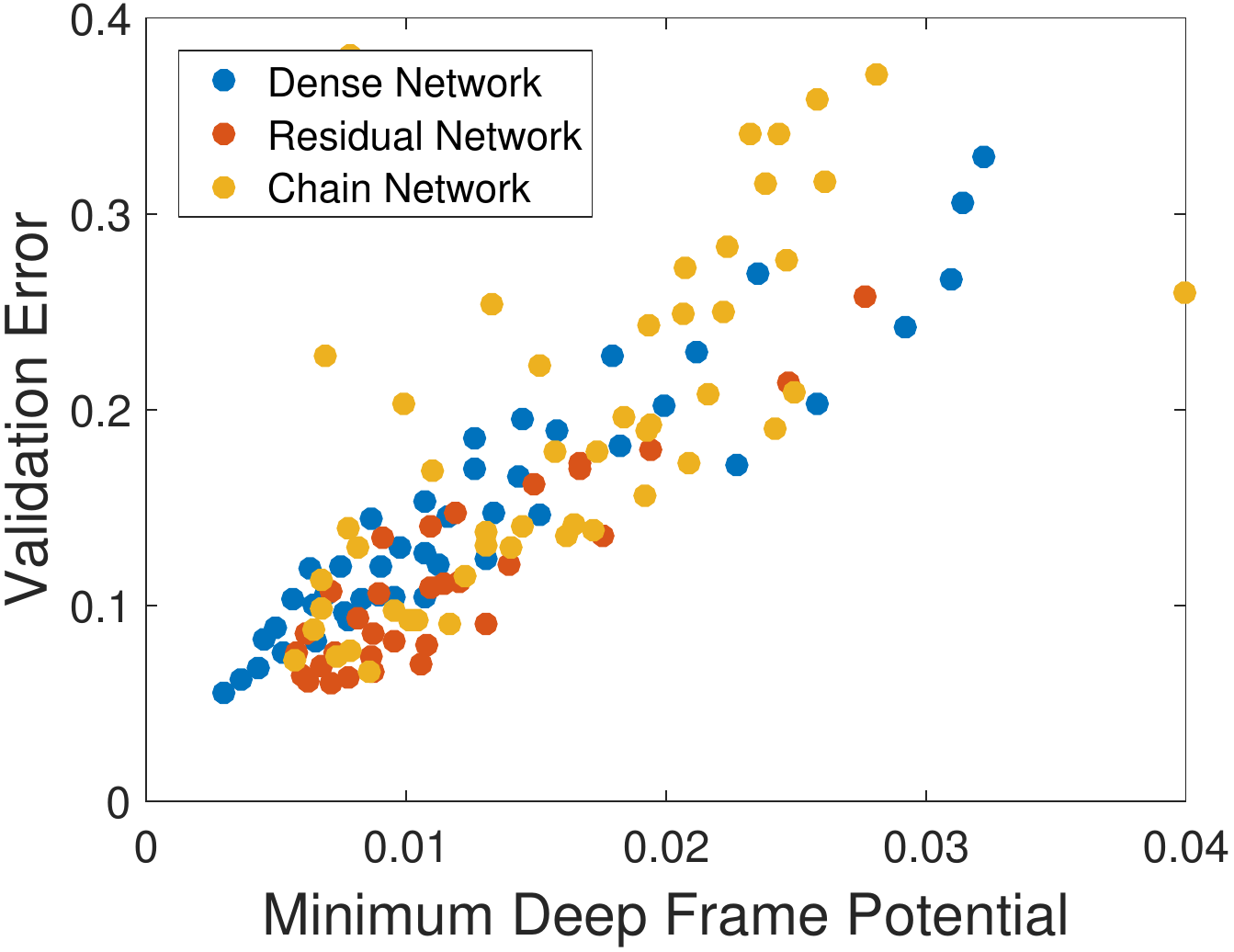}
}
\subfloat[CIFAR-100]{
\includegraphics[width=0.24\linewidth]{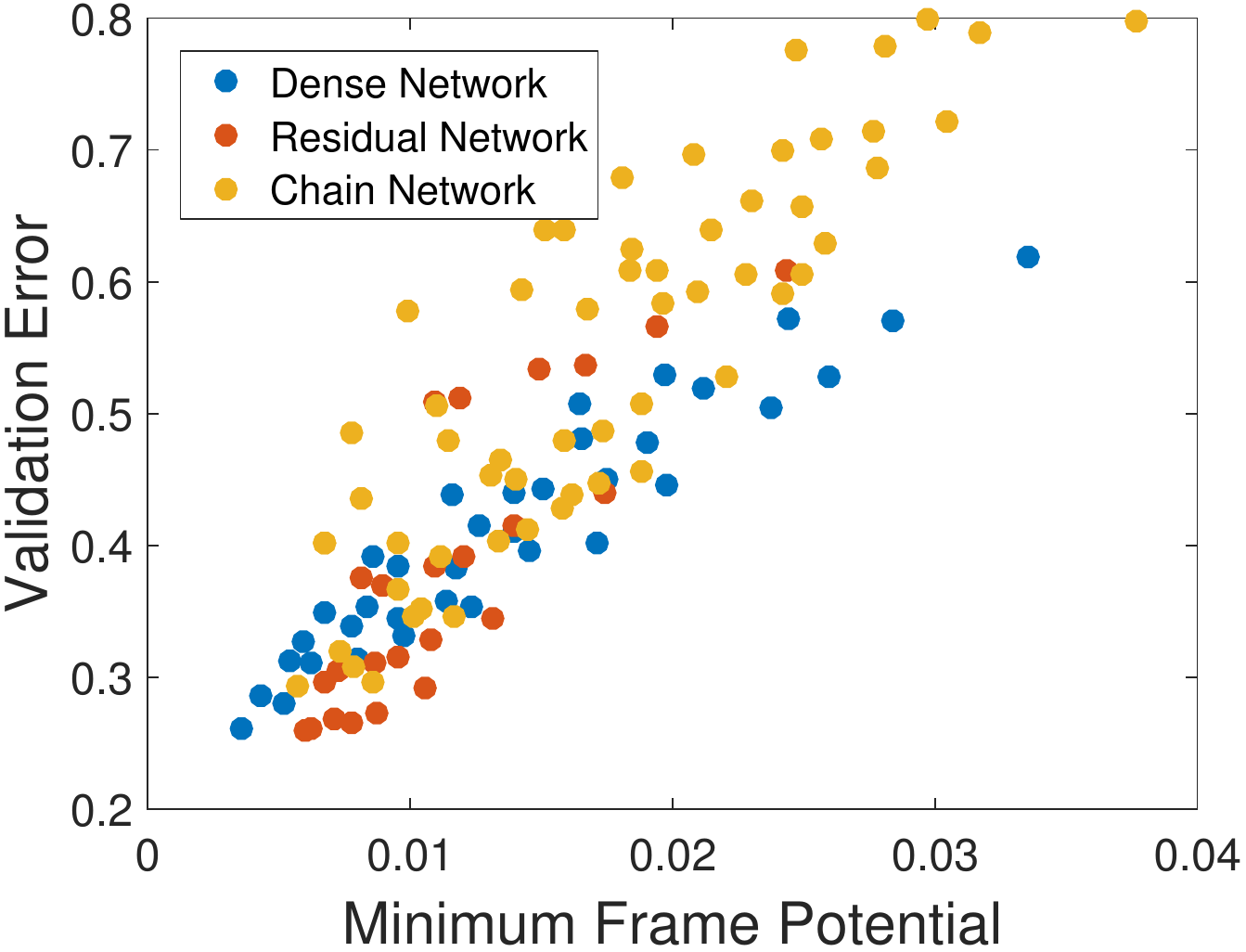}
}
\subfloat[ImageNet]{
\includegraphics[width=0.24\linewidth]{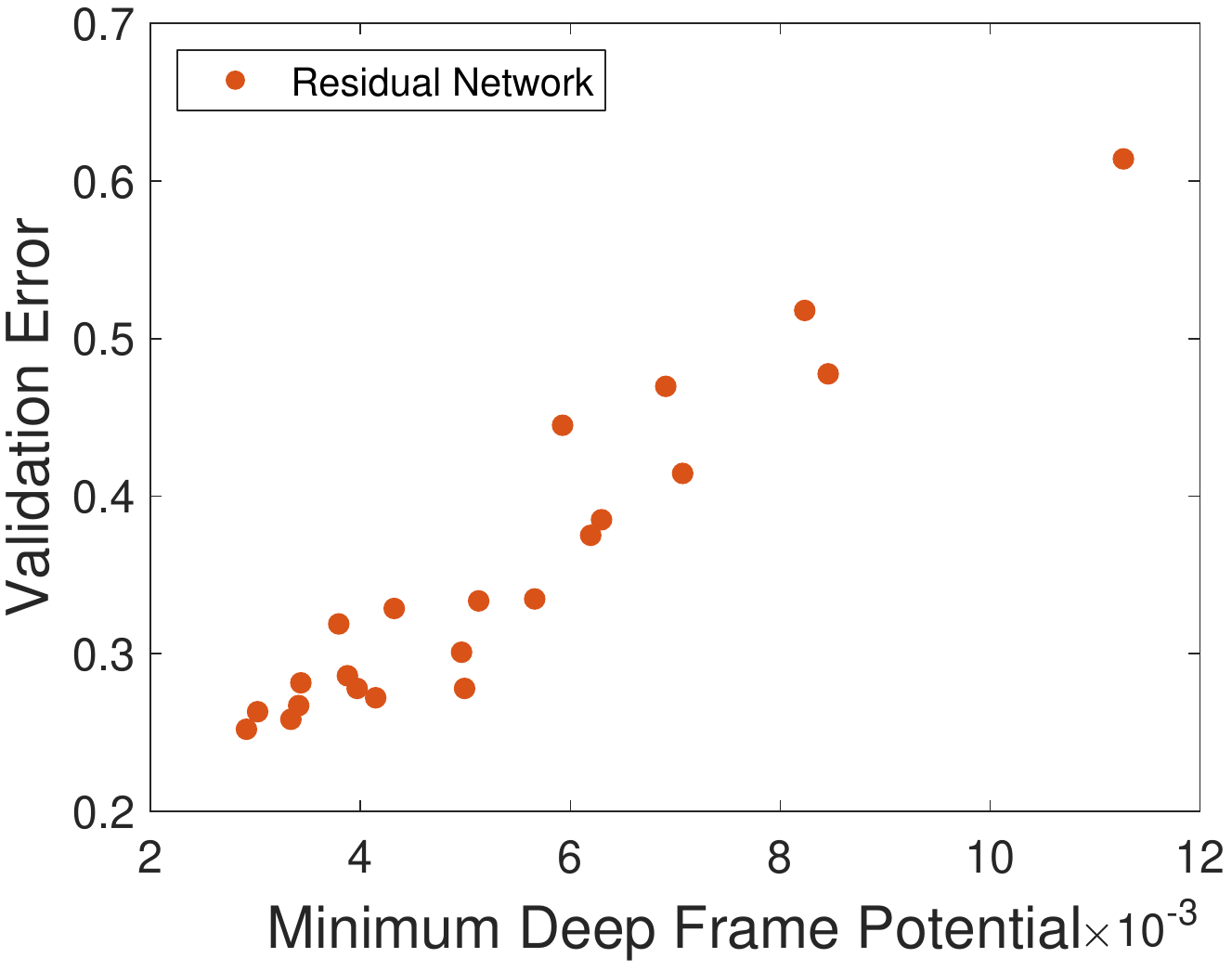}
}
\caption{The correlation between the minimum deep frame potential and validation error of varous network 
architectures on the (a) MNIST, (b) CIFAR-10, (c) CIFAR-100, and (d) ImageNet datasets. Each point represents a different architecture, and 
those in (b) are the same from Fig.~\ref{fig:gram}. Despite the simplicity of MNIST, minimum
deep frame potential is still a good predictor of validation error but on a much smaller scale. Furthermore, our insights 
extend to larger-scale architectures for CIFAR-100 and ImageNet classification.}
\label{fig:correlation}
\end{figure*}

Validation error correlates with minimum deep frame potential even across different families of architectures, as shown in Fig.~\ref{fig:correlation}.  
Because MNIST is a much simpler dataset, validation error saturates much quicker with respect to model size. 
Despite these dataset differences, minimum deep frame potential is still a good predictor of generalization error and 
DenseNets with many skip connections once again achieve superior performance. Furthermore, it remains a good predictor of performance when extended to larger-scale datasets like CIFAR-100 and  ImageNet.

\begin{figure}
\centering
\subfloat[Depth (CIFAR-10)]{
\includegraphics[width=0.32\columnwidth]{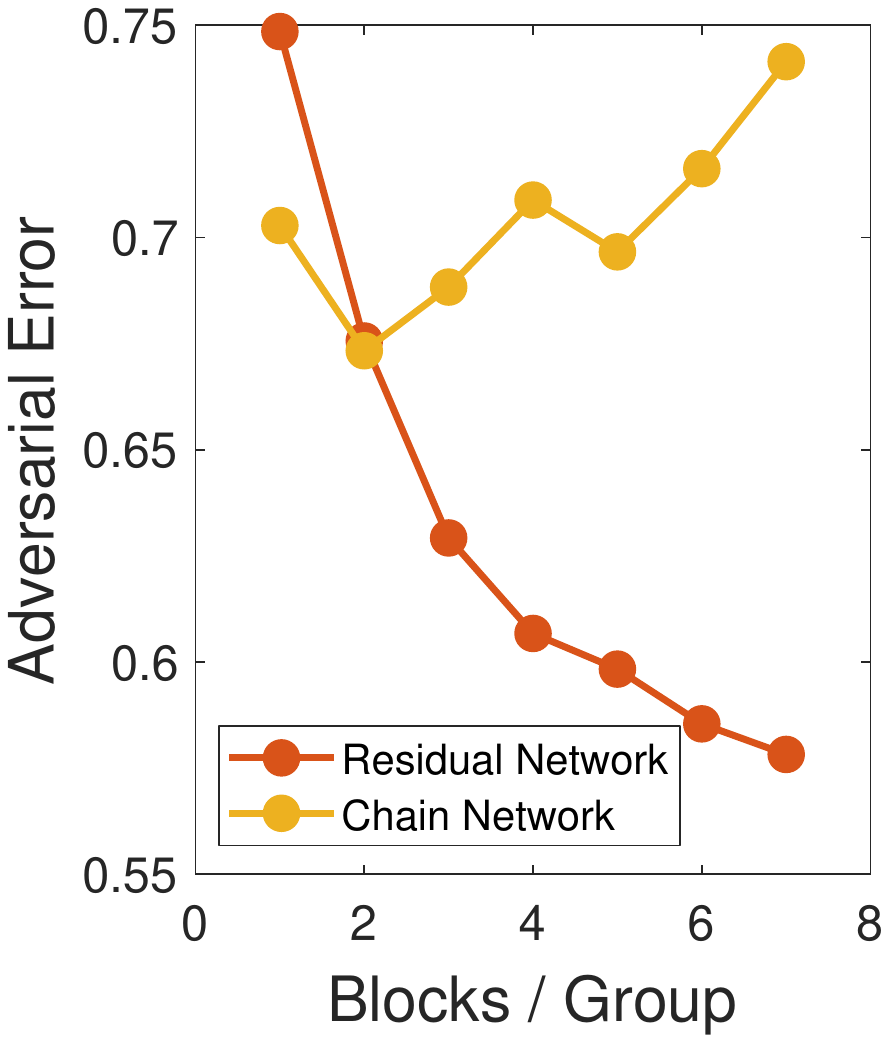}
}
\subfloat[Depth (ImageNet)]{
\includegraphics[width=0.32\columnwidth]{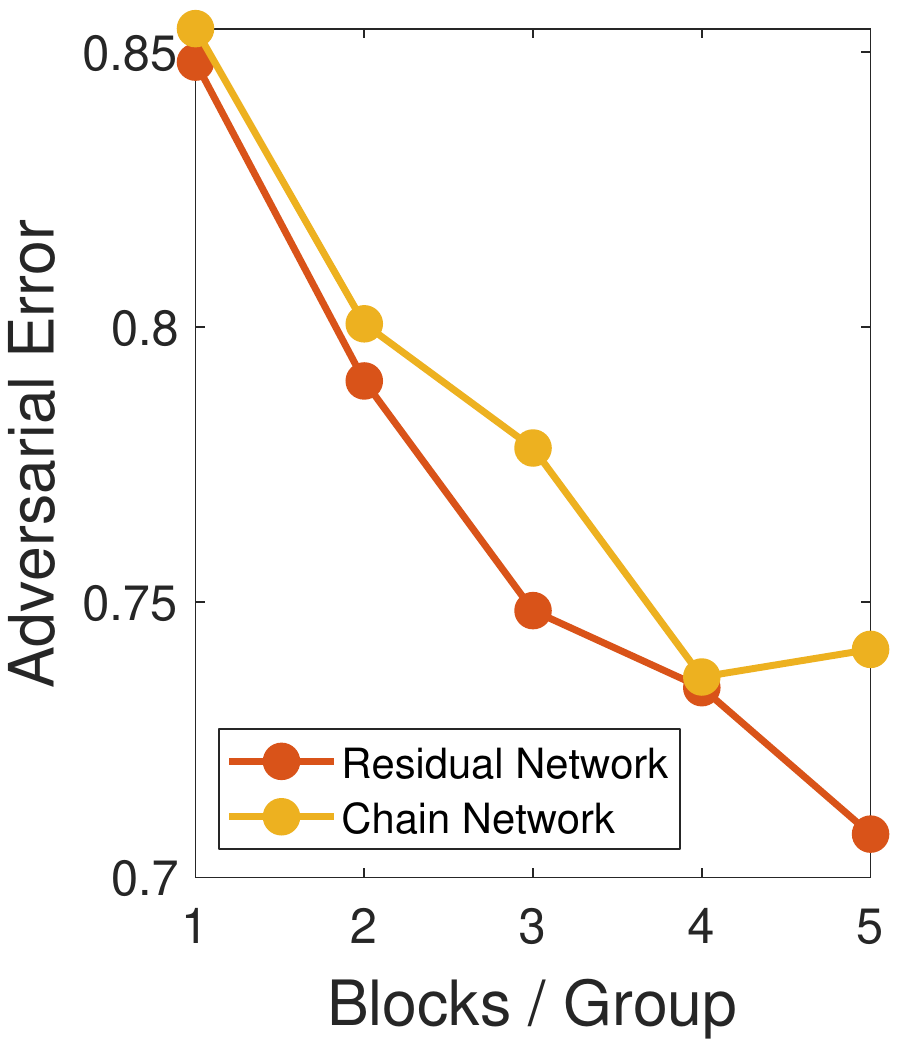}
}
\subfloat[Iterations (CIFAR-10)]{
\includegraphics[width=0.32\columnwidth]{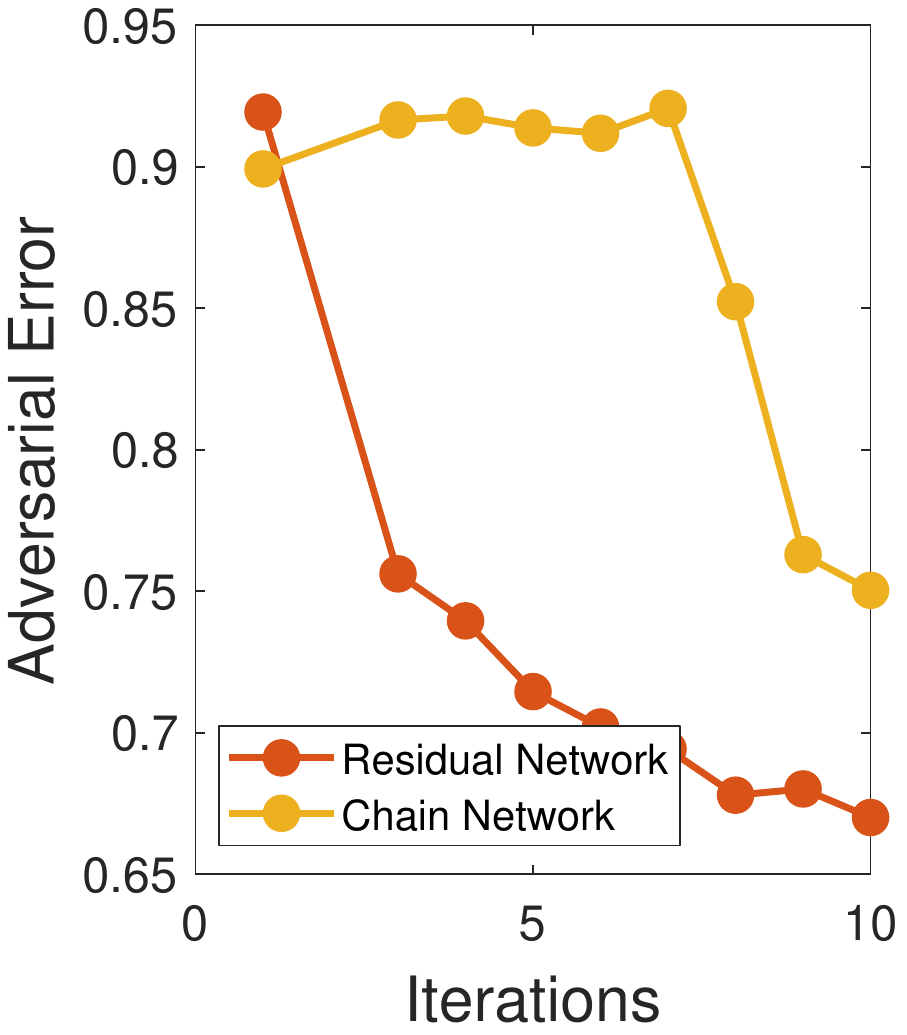}
}
\caption{
Adversarial robustness improves with decreasing minimum deep frame potential and increasing iterations. (a, b) For both CIFAR-10 and ImageNet, we see that increasing depth and the addition of residual connections improves adversarial robustness. (c) For chain networks with global iterative inference, a sufficient number of iterations is required before increased adversarial robustness is seen. However, the skip connections of residual networks allow global iterative inference to more quickly progagate information through the layers, resulting in significant performance gains with only a few iterations.
}
\label{fig:adversarial}
\end{figure}

Aside from validation error, lower deep frame potentials are also indicative of improved adversarial robustness. As we know from Fig. \ref{fig:resnet_layers}b, minimum deep frame potential decreases alongside increasing depth and dense layer connectivity. In Fig. \ref{fig:adversarial}a-b we see that added depth and the inclusion of residual connections in feed-worward networks also correlates with improved adversarial robustness.

\begin{figure*}
\centering
\subfloat[Chain CIFAR-10 Validation Error]{
\includegraphics[width=0.3\textwidth]{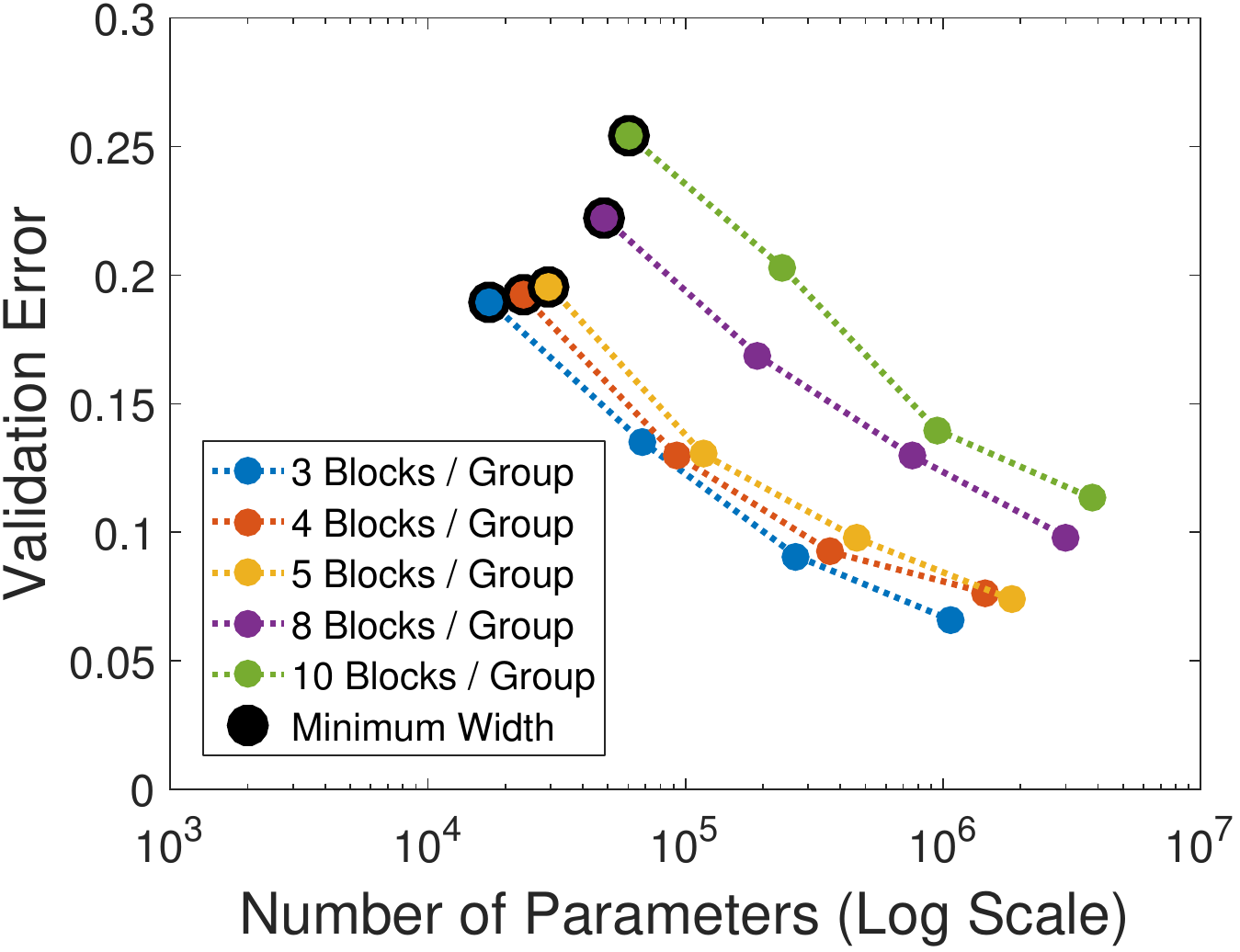}
}
\subfloat[ResNet CIFAR-10 Validation Error]{
\includegraphics[width=0.3\textwidth]{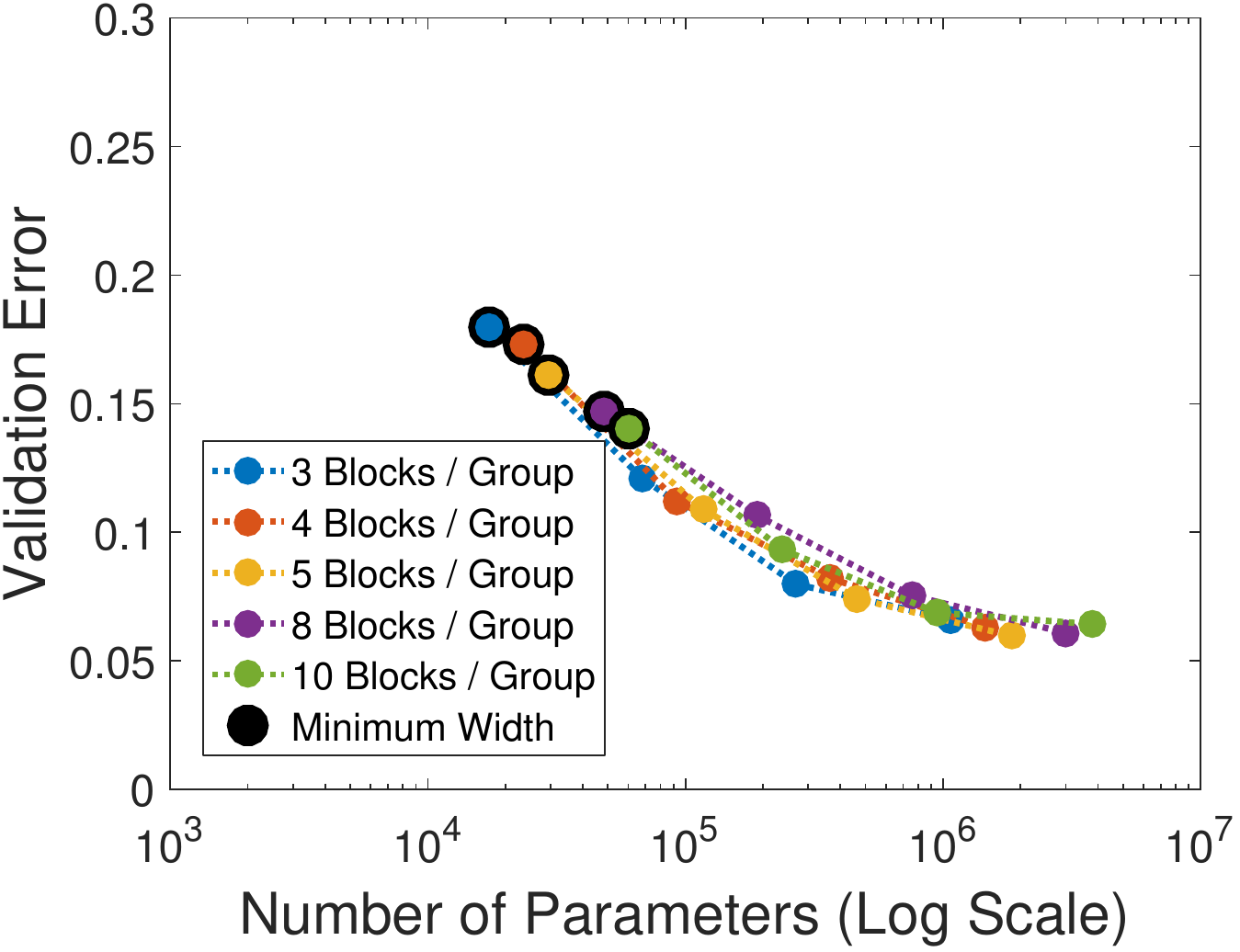}
}
\subfloat[DenseNet CIFAR-10 Validation Error]{
\includegraphics[width=0.3\textwidth]{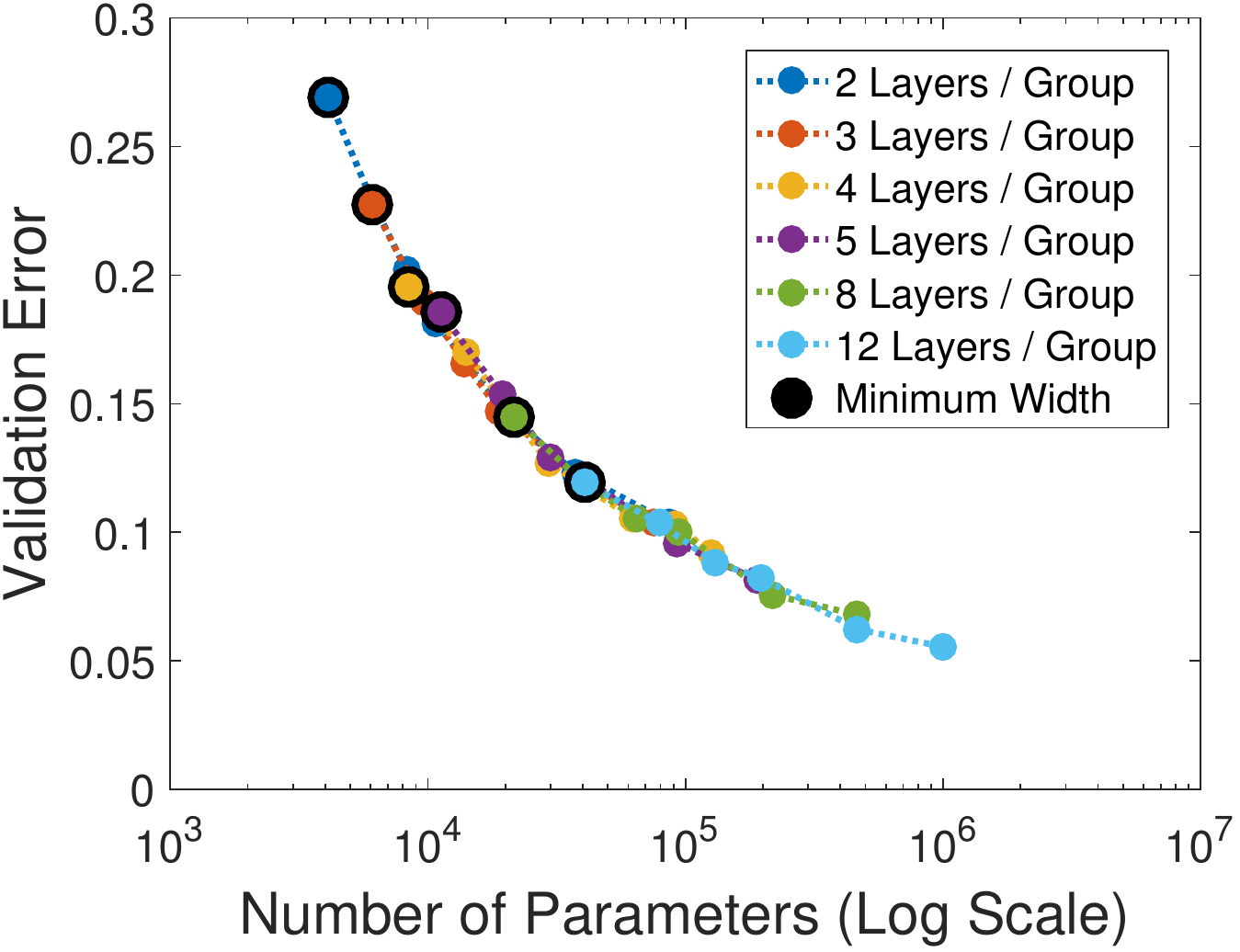}
}

\subfloat[Chain Frame Potential]{
\includegraphics[width=0.3\textwidth]{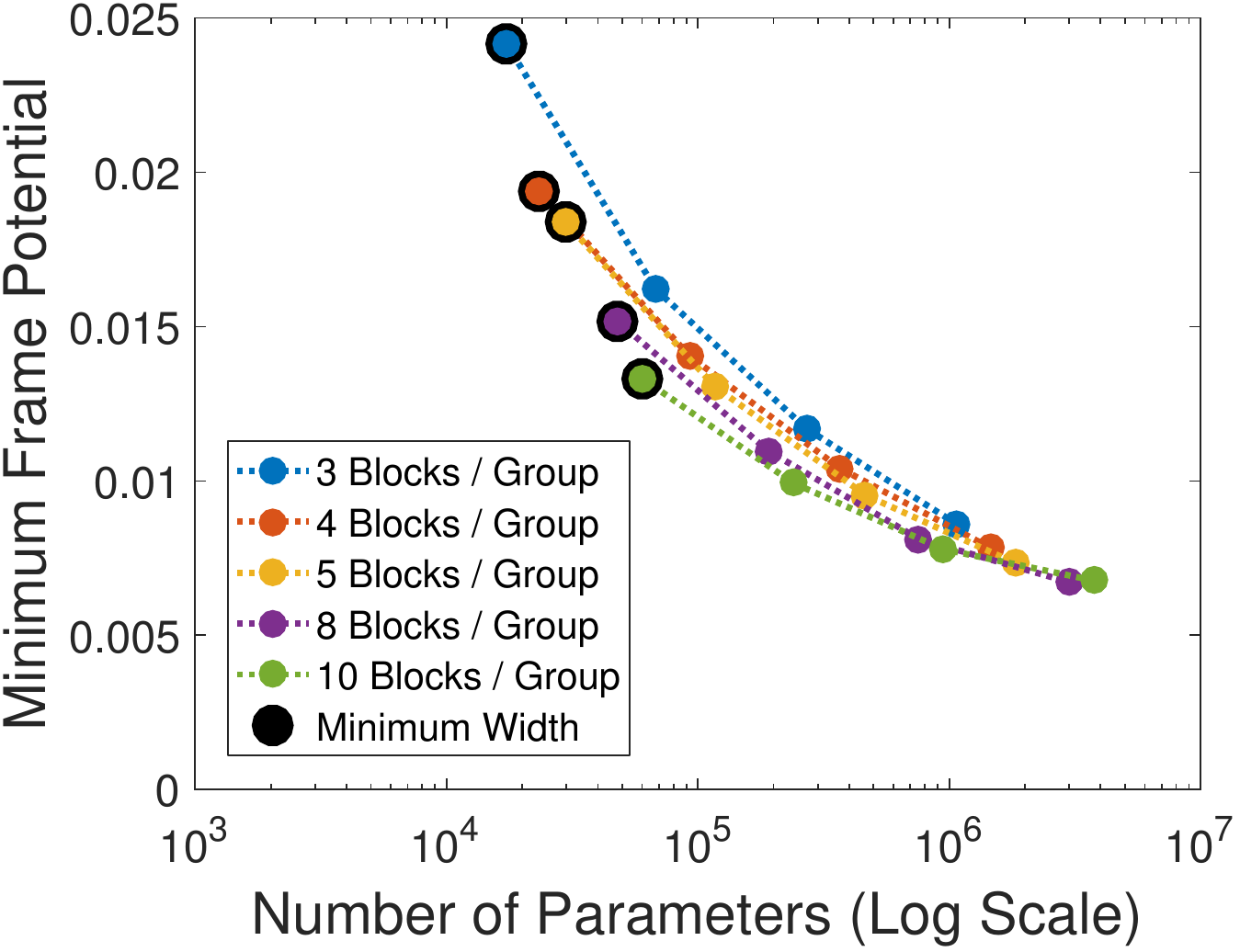}
}
\subfloat[ResNet Frame Potential]{
\includegraphics[width=0.3\textwidth]{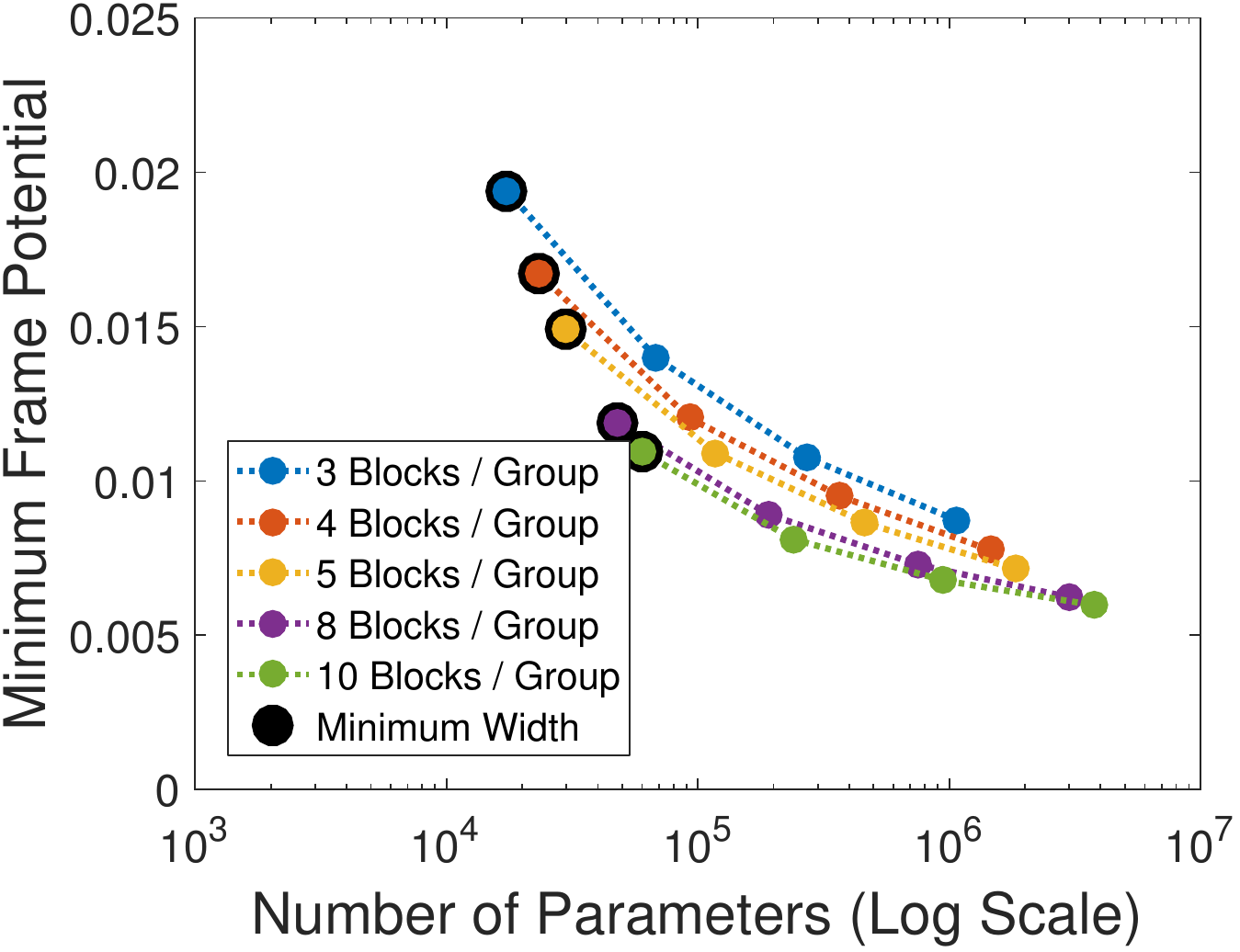}
}
\subfloat[DenseNet Frame Potential]{
\includegraphics[width=0.3\textwidth]{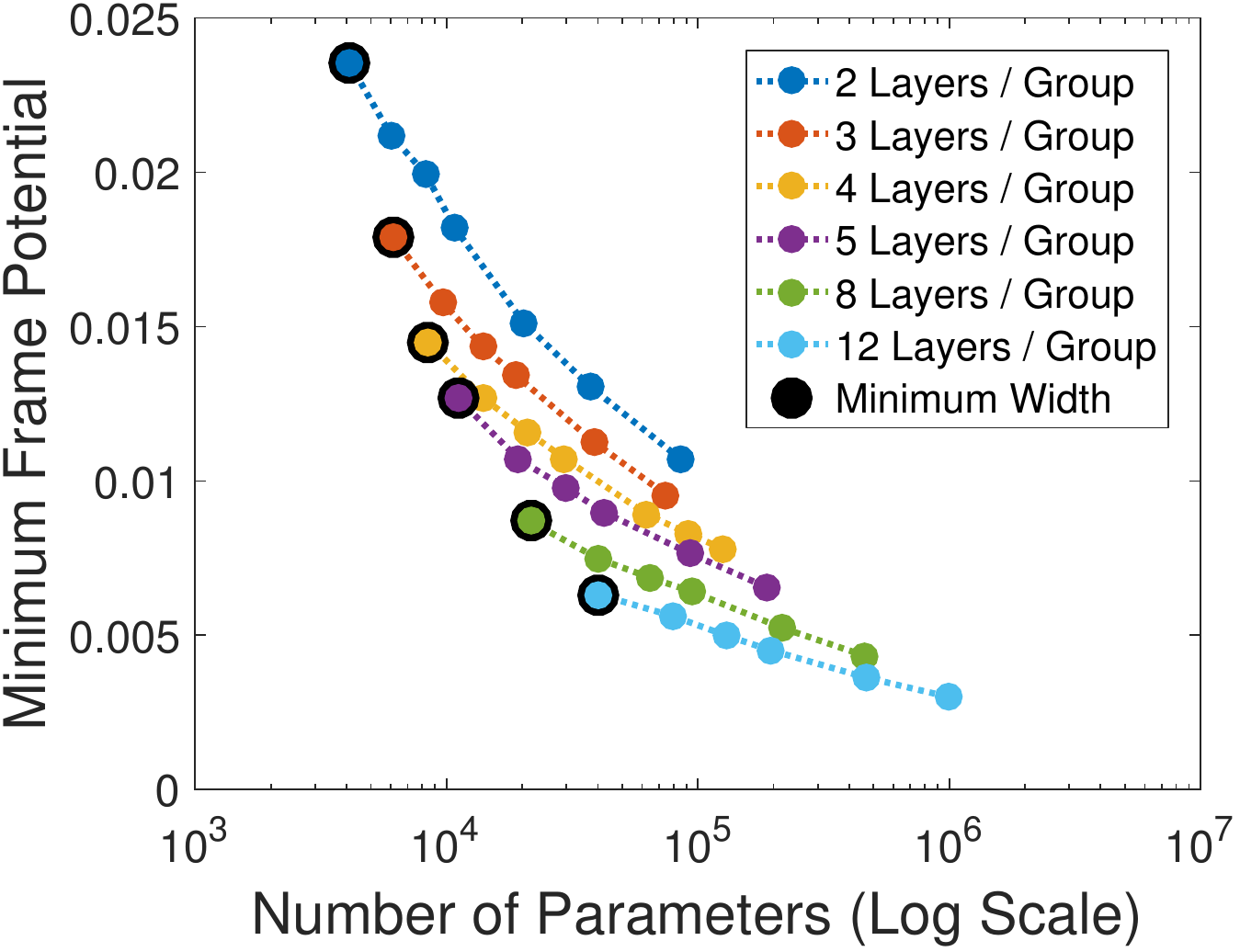}
}
\caption{
A demonstration of the improved scalability of networks with skip connections. (a) Chain networks with greater 
depths have
increasingly worse parameter efficiency in comparison to (b) the corresponding networks with residual
connections and (c) densely connected networks with similar size. The efficiency in reducing frame potential with fewer 
parameters saturates much faster with  (d) chain networks than (e) residual networks or 
(f) densely connected networks.
}
\label{fig:resnet_width}
\end{figure*}

In Fig.~\ref{fig:resnet_width}, we compare the parameter efficiency of chain networks, 
residual networks, and densely connected networks of different depths and widths. 
We visualize both validation error and 
minimum frame potential as functions of the number of parameters, 
demonstrating the improved scalability of 
networks with skip connections.
While chain networks demonstrate increasingly poor 
parameter efficiency with respect to increasing depth in 
Fig.~\ref{fig:resnet_width}a, the skip connections of ResNets and DenseNets 
allow for further reducing error with larger network sizes in 
Figs.~\ref{fig:resnet_width}b,c. Considering all network families together as in Fig.~\ref{fig:gram}d, 
we see that denser connections also allow for lower validation error with comparable numbers of parameters.
This trend is mirrored in the minimum frame potentials of Figs.~\ref{fig:resnet_width}d,e,f
shown together in Fig.~\ref{fig:gram}e. Despite fine variations in behavior across
different families of architectures, minimum deep frame potential is correlated with validation error across
network sizes and effectively predicts the increased generalization capacity provided by skip connections.

\section{Conclusion}

In this paper, we introduced deep frame approximation, a general framework for deep learning
that poses inference as global multilayer constrained approximation problems 
with structured overcomplete frames. Solutions may be approximated using
convex optimization techniques implemented by feed-forward or recurrent neural networks. 
From this perspective, the compositional nature of deep learning arises as a consequence of
structure-dependent efficient approximations to augmented shallow learning problems.

Based upon theoretical 
connections to sparse approximation and deep neural networks, we demonstrated how architectural
hyper-parameters such as 
depth, width, and skip connections induce different structural properties
of the frames in corresponding sparse coding problems. 
We compared these frame structures through their minimum frame potentials: lower
bounds on their mutual coherence, which is theoretically tied to their capacity for uniquely
and robustly representing data via sparse approximation. A theoretical lower bound was derived 
for chain networks and the deep frame potential was proposed as an empirical optimization objective
for constructing bounds for more complicated networks. While these techniques can be applied generally 
across many families of architectures, analysis of dynamic
connectivity mechanisms like self-attention is complicated due to the added data dependency and is 
left as a promising direction for future work.

Experimentally, we observed
correlations between minimum deep frame potential and validation error across different datasets and families of 
architectures, including residual networks and densely connected convolutional networks. 
This motivates future research towards the theoretical analysis and practical construction of deep network
architectures derived from connections between deep learning and overcomplete representations. 

\bibliographystyle{IEEEtran}
\bibliography{abrv,refs}

\begin{thebibliography}{10}
\providecommand{\url}[1]{#1}
\csname url@samestyle\endcsname
\providecommand{\newblock}{\relax}
\providecommand{\bibinfo}[2]{#2}
\providecommand{\BIBentrySTDinterwordspacing}{\spaceskip=0pt\relax}
\providecommand{\BIBentryALTinterwordstretchfactor}{4}
\providecommand{\BIBentryALTinterwordspacing}{\spaceskip=\fontdimen2\font plus
\BIBentryALTinterwordstretchfactor\fontdimen3\font minus
  \fontdimen4\font\relax}
\providecommand{\BIBforeignlanguage}[2]{{%
\expandafter\ifx\csname l@#1\endcsname\relax
\typeout{** WARNING: IEEEtran.bst: No hyphenation pattern has been}%
\typeout{** loaded for the language `#1'. Using the pattern for}%
\typeout{** the default language instead.}%
\else
\language=\csname l@#1\endcsname
\fi
#2}}
\providecommand{\BIBdecl}{\relax}
\BIBdecl

\bibitem{krizhevsky2012imagenet}
A.~Krizhevsky \emph{et~al.}, ``Imagenet classification with deep convolutional
  neural networks,'' in \emph{Advances in Neural Information Processing Systems
  (NeurIPS)}, 2012.

\bibitem{simonyan2014very}
K.~Simonyan and A.~Zisserman, ``Very deep convolutional networks for
  large-scale image recognition,'' in \emph{International Conference on
  Learning Representations (ICLR)}, 2015.

\bibitem{he2016deep}
K.~He, X.~Zhang, S.~Ren, and J.~Sun, ``Deep residual learning for image
  recognition,'' in \emph{Conference on Computer Vision and Pattern Recognition
  (CVPR)}, 2016.

\bibitem{huang2017densely}
G.~Huang, Z.~Liu, K.~Weinberger, and L.~van~der Maaten, ``Densely connected
  convolutional networks,'' in \emph{Conference on Computer Vision and Pattern
  Recognition (CVPR)}, 2017.

\bibitem{cortes1995support}
C.~Cortes and V.~Vapnik, ``Support-vector networks,'' \emph{Machine learning},
  vol.~20, no.~3, 1995.

\bibitem{vapnik1971uniform}
V.~N. Vapnik and A.~Y. Chervonenkis, ``On the uniform convergence of relative
  frequencies of events to their probabilities,'' \emph{Theory of Probability
  and Its Applications}, vol. XVI, no.~2, 1971.

\bibitem{papyan2017convolutional}
V.~Papyan, Y.~Romano, and M.~Elad, ``Convolutional neural networks analyzed via
  convolutional sparse coding,'' \emph{Journal of Machine Learning Research
  (JMLR)}, vol.~18, no.~83, 2017.

\bibitem{murdock2018deep}
C.~Murdock, M.~Chang, and S.~Lucey, ``Deep component analysis via alternating
  direction neural networks,'' in \emph{European Conference on Computer Vision
  (ECCV)}, 2018.

\bibitem{murdock2020frame}
C.~Murdock and S.~Lucey, ``Dataless model selection with the deep frame
  potential,'' in \emph{Conference on Computer Vision and Pattern Recognition
  (CVPR)}, 2020.

\bibitem{russakovsky2015imagenet}
O.~Russakovsky, J.~Deng, H.~Su, J.~Krause, S.~Satheesh, S.~Ma, Z.~Huang,
  A.~Karpathy, A.~Khosla, M.~Bernstein \emph{et~al.}, ``Imagenet large scale
  visual recognition challenge,'' \emph{International Journal of Computer
  Vision (IJCV)}, vol. 115, no.~3, 2015.

\bibitem{elad2010sparse}
M.~Elad, \emph{Sparse and redundant representations: from theory to
  applications in signal and image processing}.\hskip 1em plus 0.5em minus
  0.4em\relax Springer Science \& Business Media, 2010.

\bibitem{kovavcevic2008introduction}
J.~Kova{\v{c}}evi{\'c}, A.~Chebira \emph{et~al.}, ``An introduction to
  frames,'' \emph{Foundations and Trends in Signal Processing}, vol.~2, no.~1,
  2008.

\bibitem{donoho2003optimally}
D.~L. Donoho and M.~Elad, ``Optimally sparse representation in general
  (nonorthogonal) dictionaries via l1 minimization,'' \emph{Proceedings of the
  National Academy of Sciences}, vol. 100, no.~5, 2003.

\bibitem{zhang2017understanding}
C.~Zhang, S.~Bengio, M.~Hardt, B.~Recht, and O.~Vinyals, ``Understanding deep
  learning requires rethinking generalization,'' in \emph{International
  Conference on Learning Representations (ICLR)}, 2017.

\bibitem{dosovitskiy2020image}
A.~Dosovitskiy, L.~Beyer, A.~Kolesnikov, D.~Weissenborn, X.~Zhai,
  T.~Unterthiner, M.~Dehghani, M.~Minderer, G.~Heigold, S.~Gelly, J.~Uszkoreit,
  and N.~Houlsby, ``An image is worth 16x16 words: Transformers for image
  recognition at scale,'' in \emph{International Conference on Learning
  Representations (ICLR)}, 2020.

\bibitem{cazenavette2021adversarial}
G.~Cazenavette, C.~Murdock, and S.~Lucey, ``Architectural adversarial
  robustness: The case for deep pursuit,'' in \emph{Conference on Computer
  Vision and Pattern Recognition (CVPR)}, 2021.

\bibitem{wold1987principal}
S.~Wold, K.~Esbensen, and P.~Geladi, ``Principal component analysis,''
  \emph{Chemometrics and intelligent laboratory systems}, vol.~2, no. 1-3, pp.
  37--52, 1987.

\bibitem{bao2016dictionary}
C.~Bao, H.~Ji, Y.~Quan, and Z.~Shen, ``Dictionary learning for sparse coding:
  Algorithms and convergence analysis,'' \emph{Pattern Analysis and Machine
  Intelligence (PAMI)}, vol.~38, no.~7, pp. 1356--1369, 2016.

\bibitem{turk1991eigenfaces}
M.~Turk and A.~Pentland, ``Eigenfaces for recognition,'' \emph{Journal of
  cognitive neuroscience}, vol.~3, no.~1, pp. 71--86, 1991.

\bibitem{yang2009linear}
J.~Yang, K.~Yu, Y.~Gong, and T.~Huang, ``Linear spatial pyramid matching using
  sparse coding for image classification,'' in \emph{Conference on Computer
  Vision and Pattern Recognition (CVPR)}, 2009.

\bibitem{jutten1991blind}
C.~Jutten and J.~Herault, ``Blind separation of sources, part i: An adaptive
  algorithm based on neuromimetic architecture,'' \emph{Signal Processing},
  vol.~24, no.~1, pp. 1--10, 1991.

\bibitem{lee1999learning}
D.~D. Lee and H.~S. Seung, ``Learning the parts of objects by non-negative
  matrix factorization,'' \emph{Nature}, vol. 401, no. 6755, pp. 788--791,
  1999.

\bibitem{olshausen1996emergence}
B.~A. Olshausen \emph{et~al.}, ``Emergence of simple-cell receptive field
  properties by learning a sparse code for natural images,'' \emph{Nature},
  vol. 381, no. 6583, pp. 607--609, 1996.

\bibitem{liu2016dimensionality}
T.~Liu, D.~Tao, and D.~Xu, ``Dimensionality-dependent generalization bounds for
  k-dimensional coding schemes,'' \emph{Neural computation}, 2016.

\bibitem{gillis2012sparse}
N.~Gillis, ``Sparse and unique nonnegative matrix factorization through data
  preprocessing,'' \emph{Journal of Machine Learning Research (JMLR)}, vol.~13,
  no. November, pp. 3349--3386, 2012.

\bibitem{haeffele2014structured}
B.~Haeffele, E.~Young, and R.~Vidal, ``Structured low-rank matrix
  factorization: Optimality, algorithm, and applications to image processing,''
  in \emph{International Conference on Machine Learning (ICML)}, 2014.

\bibitem{baldi1989neural}
P.~Baldi and K.~Hornik, ``Neural networks and principal component analysis:
  Learning from examples without local minima,'' \emph{Neural networks},
  vol.~2, no.~1, 1989.

\bibitem{delatorre2012least}
F.~{De la Torre}, ``A least-squares framework for component analysis,''
  \emph{Pattern Analysis and Machine Intelligence (PAMI)}, vol.~34, no.~6,
  2012.

\bibitem{lee2009convolutional}
H.~Lee, R.~Grosse, R.~Ranganath, and A.~Y. Ng, ``Convolutional deep belief
  networks for scalable unsupervised learning of hierarchical
  representations,'' in \emph{International Conference on Machine Learning
  (ICML)}, 2009.

\bibitem{lecun1998gradient}
Y.~LeCun, L.~Bottou, Y.~Bengio, and P.~Haffner, ``Gradient-based learning
  applied to document recognition,'' \emph{Proceedings of the IEEE}, vol.~86,
  no.~11, 1998.

\bibitem{denil2013predicting}
M.~Denil \emph{et~al.}, ``Predicting parameters in deep learning,'' in
  \emph{Advances in Neural Information Processing Systems (NeurIPS)}, 2013.

\bibitem{alvarez2016learning}
J.~Alvarez and M.~Salzmann, ``Learning the number of neurons in deep
  networks,'' in \emph{Advances in Neural Information Processing Systems
  (NeurIPS)}, 2016.

\bibitem{he2017channel}
Y.~He, X.~Zhang, and J.~Sun, ``Channel pruning for accelerating very deep
  neural networks,'' in \emph{Conference on Computer Vision and Pattern
  Recognition (CVPR)}, 2017.

\bibitem{elsken2019neural}
T.~Elsken, J.~H. Metzen, and F.~Hutter, ``Neural architecture search: A
  survey.'' \emph{Journal of Machine Learning Research (JMLR)}, vol.~20,
  no.~55, 2019.

\bibitem{tan2019efficientnet}
M.~Tan and Q.~Le, ``Efficientnet: Rethinking model scaling for convolutional
  neural networks,'' in \emph{International Conference on Machine Learning
  (ICML)}, 2019.

\bibitem{arpit2017closer}
D.~Arpit \emph{et~al.}, ``A closer look at memorization in deep networks,'' in
  \emph{International Conference on Machine Learning (ICML)}, 2017.

\bibitem{hardt2016train}
M.~Hardt, B.~Recht, and Y.~Singer, ``Train faster, generalize better: Stability
  of stochastic gradient descent,'' in \emph{International Conference on
  Machine Learning (ICML)}, 2016.

\bibitem{cao2019generalization}
Y.~Cao and Q.~Gu, ``Generalization bounds of stochastic gradient descent for
  wide and deep neural networks,'' \emph{Advances in Neural Information
  Processing Systems (NeurIPS)}, 2019.

\bibitem{shwartz2017opening}
R.~Shwartz-Ziv and N.~Tishby, ``Opening the black box of deep neural networks
  via information,'' \emph{arXiv preprint arXiv:1703.00810}, 2017.

\bibitem{saxe2018information}
A.~M. Saxe, Y.~Bansal, J.~Dapello, M.~Advani, A.~Kolchinsky, B.~D. Tracey, and
  D.~D. Cox, ``On the information bottleneck theory of deep learning,'' in
  \emph{International Conference on Learning Representations (ICLR)}, 2018.

\bibitem{novak2018sensitivity}
R.~Novak \emph{et~al.}, ``Sensitivity and generalization in neural networks: an
  empirical study,'' in \emph{International Conference on Learning
  Representations (ICLR)}, 2018.

\bibitem{moustapha2017parseval}
C.~Moustapha, B.~Piotr, G.~Edouard, D.~Yann, and U.~Nicolas, ``Parseval
  networks: Improving robustness to adversarial examples,'' in
  \emph{International Conference on Machine Learning (ICML)}, 2017.

\bibitem{romano2018adversarial}
Y.~Romano, A.~Aberdam, J.~Sulam, and M.~Elad, ``Adversarial noise attacks of
  deep learning architectures-stability analysis via sparse modeled signals,''
  \emph{Journal of Mathematical Imaging and Vision}, 2018.

\bibitem{sulam2019multi}
J.~Sulam, A.~Aberdam, A.~Beck, and M.~Elad, ``On multi-layer basis pursuit,
  efficient algorithms and convolutional neural networks,'' \emph{Pattern
  Analysis and Machine Intelligence (PAMI)}, 2019.

\bibitem{waldron2018introduction}
S.~F. Waldron, \emph{An introduction to finite tight frames}.\hskip 1em plus
  0.5em minus 0.4em\relax Springer, 2018.

\bibitem{casazza2003equal}
P.~G. Casazza and J.~Kova{\v{c}}evi{\'c}, ``Equal-norm tight frames with
  erasures,'' \emph{Advances in Computational Mathematics}, vol.~18, no. 2-4,
  2003.

\bibitem{benedetto2003finite}
J.~Benedetto and M.~Fickus, ``Finite normalized tight frames,'' \emph{Advances
  in Computational Mathematics}, vol.~18, no. 2-4, 2003.

\bibitem{casazza2009gradient}
P.~Casazza and M.~Fickus, ``Gradient descent of the frame potential,'' in
  \emph{International Conference on Sampling Theory and Applications}, 2009.

\bibitem{parikh2014proximal}
N.~Parikh, S.~Boyd \emph{et~al.}, ``Proximal algorithms,'' \emph{Foundations
  and Trends{\textregistered} in Optimization}, vol.~1, no.~3, 2014.

\bibitem{combettes2020deep}
P.~L. Combettes and J.-C. Pesquet, ``Deep neural network structures solving
  variational inequalities,'' \emph{Set-Valued and Variational Analysis}, 2020.

\bibitem{daubechies2004iterative}
I.~Daubechies, M.~Defrise, and C.~De~Mol, ``An iterative thresholding algorithm
  for linear inverse problems with a sparsity constraint,''
  \emph{Communications on Pure and Applied Mathematics: A Journal Issued by the
  Courant Institute of Mathematical Sciences}, vol.~57, no.~11, 2004.

\bibitem{bengio2013representation}
Y.~Bengio, A.~Courville, and P.~Vincent, ``Representation learning: A review
  and new perspectives,'' \emph{Pattern Analysis and Machine Intelligence
  (PAMI)}, vol.~35, no.~8, pp. 1798--1828, 2013.

\bibitem{tropp2007signal}
J.~A. Tropp and A.~C. Gilbert, ``Signal recovery from random measurements via
  orthogonal matching pursuit,'' \emph{IEEE Transactions on information
  theory}, vol.~53, no.~12, 2007.

\bibitem{bruckstein2008uniqueness}
A.~M. Bruckstein, M.~Elad, and M.~Zibulevsky, ``On the uniqueness of
  nonnegative sparse solutions to underdetermined systems of equations,''
  \emph{IEEE Transactions on Information Theory}, vol.~54, no.~11, 2008.

\bibitem{donoho2006compressed}
D.~L. Donoho, ``Compressed sensing,'' \emph{IEEE Transactions on Information
  Theory}, vol.~52, no.~4, 2006.

\bibitem{donoho2005stable}
D.~L. Donoho, M.~Elad, and V.~N. Temlyakov, ``Stable recovery of sparse
  overcomplete representations in the presence of noise,'' \emph{IEEE
  Transactions on Information Theory}, vol.~52, no.~1, 2005.

\bibitem{livezey2019learning}
J.~A. Livezey, A.~F. Bujan, and F.~T. Sommer, ``Learning overcomplete, low
  coherence dictionaries with linear inference.'' \emph{Journal of Machine
  Learning Research (JMLR)}, vol.~20, no. 174, 2019.

\bibitem{welch1974lower}
L.~Welch, ``Lower bounds on the maximum cross correlation of signals,''
  \emph{IEEE Transactions on Information theory}, vol.~20, no.~3, 1974.

\bibitem{zeiler2010deconvolutional}
M.~D. Zeiler, D.~Krishnan, G.~W. Taylor, and R.~Fergus, ``Deconvolutional
  networks,'' in \emph{Conference on Computer Vision and Pattern Recognition
  (CVPR)}, 2010.

\bibitem{bristow2013fast}
H.~Bristow, A.~Eriksson, and S.~Lucey, ``Fast convolutional sparse coding,'' in
  \emph{Conference on Computer Vision and Pattern Recognition (CVPR)}, 2013.

\bibitem{papyan2017working}
V.~Papyan, J.~Sulam, and M.~Elad, ``Working locally thinking globally:
  Theoretical guarantees for convolutional sparse coding,'' \emph{IEEE
  Transactions on Signal Processing}, vol.~65, no.~21, 2017.

\bibitem{xu2013block}
Y.~Xu and W.~Yin, ``A block coordinate descent method for regularized
  multiconvex optimization with applications to nonnegative tensor
  factorization and completion,'' \emph{SIAM Journal on imaging sciences},
  vol.~6, no.~3, pp. 1758--1789, 2013.

\bibitem{chodosh2018deep}
N.~Chodosh, C.~Wang, and S.~Lucey, ``Deep convolutional compressed sensing for
  lidar depth completion,'' in \emph{Asian Conference on Computer Vision},
  2018.

\bibitem{krizhevsky2009learning}
A.~Krizhevsky and G.~Hinton, ``Learning multiple layers of features from tiny
  images,'' University of Toronto, Tech. Rep., 2009.

\bibitem{wu2016tensorpack}
Y.~Wu \emph{et~al.}, ``Tensorpack,''
  \url{https://github.com/tensorpack/tensorpack/blob/master/examples/ResNet/cifar10-resnet.py},
  2016.

\bibitem{li2018densepack}
Y.~Li, ``Tensorflow densenet,''
  \url{https://github.com/YixuanLi/densenet-tensorflow}, 2018.

\bibitem{ioffe2015batch}
S.~Ioffe and C.~Szegedy, ``Batch normalization: Accelerating deep network
  training by reducing internal covariate shift,'' in \emph{International
  Conference on Machine Learning (ICML)}, 2015.

\end{thebibliography}

\begin{IEEEbiography}[{\includegraphics[width=1in,height=1.25in,clip,keepaspectratio]{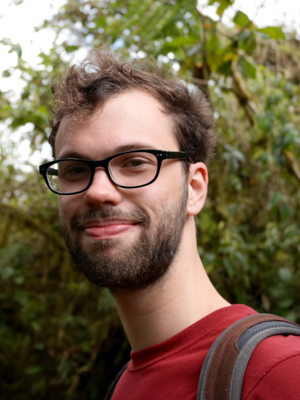}}]{Calvin Murdock}
received the PhD degree in machine learning from 
Carnegie Mellon University in 2020. His research concerns fundamental questions of 
representation learning towards the goal of effective computational perception, 
leveraging insights from geometry, optimization, and sparse approximation theory. He is a member
of the IEEE.
\end{IEEEbiography}

\begin{IEEEbiography}[{\includegraphics[width=1in,height=1.25in,clip,keepaspectratio]{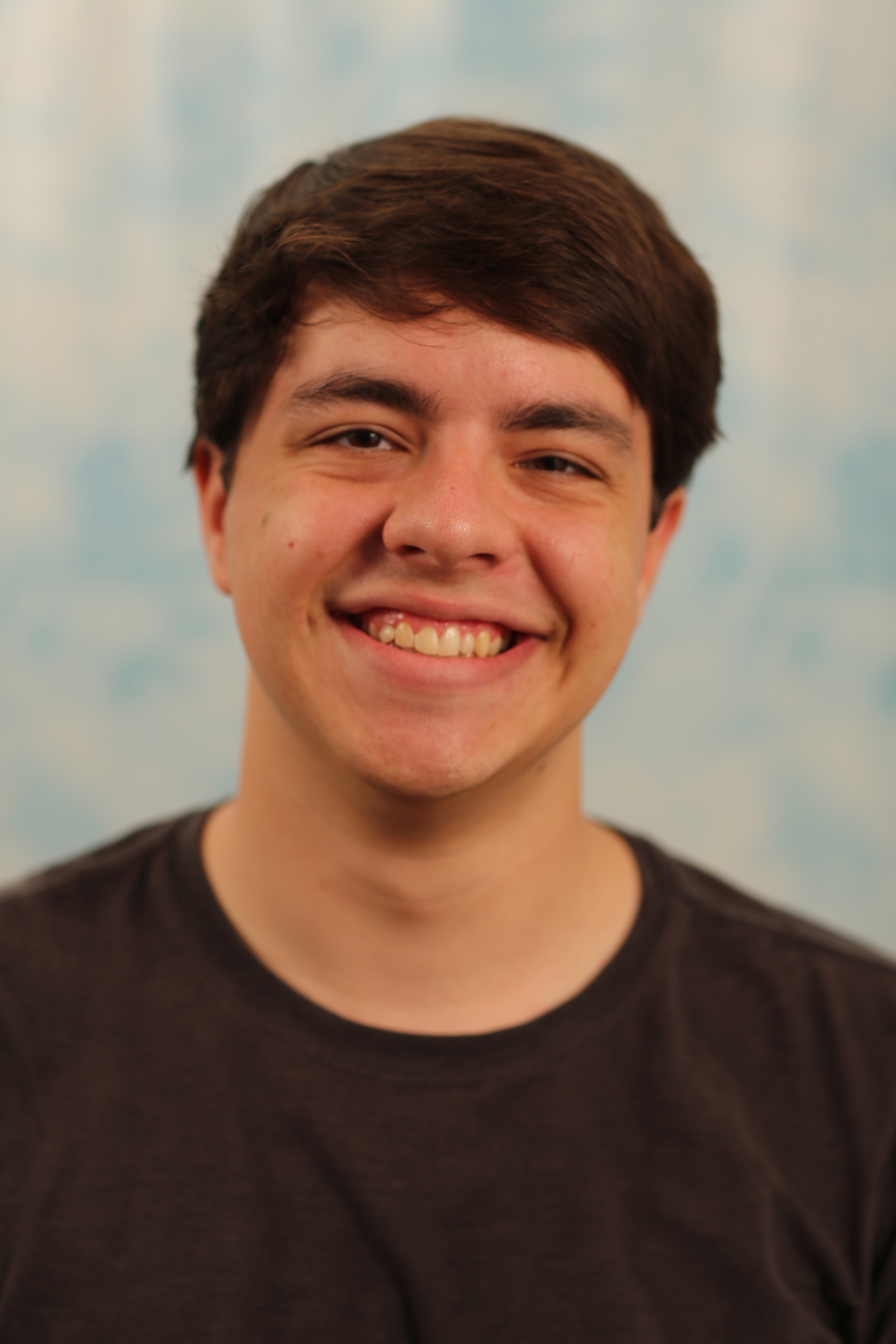}}]{George Cazenavette}
Is a graduate student in the Robotics Institute at Carnegie Mellon University. His research concerns the intrinsic properties of architecures and datasets and how they can be leveraged towards more-efficient learning paradigms. He is a student member
of the IEEE.
\end{IEEEbiography}

\begin{IEEEbiography}[{\includegraphics[width=1in,height=1.25in,clip,keepaspectratio]{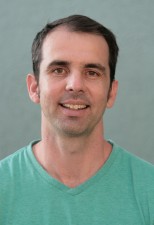}}]{Simon Lucey}
received the PhD degree from the
Queensland University of Technology, Brisbane,
Australia, in 2003. He is an Associate Research
Professor at the Robotics Institute at Carnegie
Mellon University. He also holds an adjunct professorial position at the Queensland University
of Technology. His research interests include
computer vision and machine learning, and their
application to human behavior. He is a member
of the IEEE.

\end{IEEEbiography}

\end{document}